\pdfoutput=1
\documentclass{article}

\usepackage[utf8]{inputenc} 
\usepackage{lmodern}
\usepackage{textcomp}
\usepackage{url}            
\usepackage{booktabs}       
\usepackage{amsfonts}       
\usepackage{nicefrac}       
\usepackage[activate={true,nocompatibility},final,kerning=true,spacing=true,factor=1100,stretch=10,shrink=10]{microtype} 
\usepackage{amsmath}
\usepackage{amsthm}
\usepackage{listings}
\usepackage{graphicx}
\usepackage{xspace}
\usepackage{multirow}
\usepackage{array}
\usepackage{bbm}
\usepackage{amsmath}
\usepackage{bm}
\usepackage{adjustbox}
\usepackage[round]{natbib}
\usepackage{rotating}
\usepackage{afterpage}
\usepackage{pdflscape}

\usepackage[paperVersion]{marcpaper}

\lstnewenvironment{python}[1][]
{
\pythonstyle
\lstset{#1}
}
{}

\DeclareFixedFont{\ttb}{T1}{txtt}{bx}{n}{8} 
\DeclareFixedFont{\ttm}{T1}{txtt}{m}{n}{8}  

\usepackage{color}
\definecolor{deepblue}{rgb}{0,0,0.5}
\definecolor{deepred}{rgb}{0.6,0,0}
\definecolor{deepgreen}{rgb}{0,0.5,0}

\newcommand{\name}[1]{#1 \vspace{-8pt}}
\newcommand{\affiliation}[1]{{\normalsize #1} \vspace{-8pt}}
\newcommand{\email}[1]{{\normalsize \texttt{#1}} \vspace{-4pt}}

\newcommand\pythonstyle{\lstset{
language=Python,
basicstyle=\ttm,
otherkeywords={self, with, as},             
keywordstyle=\ttb\color{deepblue},
numbers=left,
stepnumber=1,
firstnumber=1,
numberfirstline=false,
numberstyle=\tiny,
emph={@CompileMe, Param, Var},          
emphstyle=\ttb\color{deepred},    
commentstyle=\color{deepgreen},
frame=tb,                         
showstringspaces=false            %
}}

\usepackage{color}
\newcommand{\langname}{{\sc TerpreT}\xspace} 

\usepackage{commands}

\title{\langname: A Probabilistic Programming Language for Program Induction}

\author{
  \name{Alexander L. Gaunt} \\
  \affiliation{Microsoft Research} \\
  \email{t-algaun@microsoft.com}
  \and
  \name{Marc Brockschmidt} \\
  \affiliation{Microsoft Research} \\
  \email{mabrocks@microsoft.com}
  \and
  \name{Rishabh Singh} \\
  \affiliation{Microsoft Research} \\
  \email{risin@microsoft.com}
  \and
  \name{Nate Kushman} \\
  \affiliation{Microsoft Research} \\
  \email{nkushman@microsoft.com}
  \and
  \name{Pushmeet Kohli} \\
  \affiliation{Microsoft Research} \\
  \email{pkohli@microsoft.com}
  \and
  \name{Jonathan Taylor} \\
  \affiliation{perceptiveIO\footnote{Work done while author was at Microsoft Research.}} \\
  \email{jtaylor@perceptiveio.com}
  \and
  \name{Daniel Tarlow} \\
  \affiliation{Microsoft Research} \\
  \email{dtarlow@microsoft.com}
}
\date{}

\begin{document}

\maketitle

\begin{abstract}
  
  We study machine learning formulations of inductive program
  synthesis; that is, given input-output examples, we would like to
  synthesize source code that maps inputs to corresponding outputs.
  Our aims in this work are to develop new machine learning approaches
  to the problem based on neural networks and graphical models, and to
  understand the capabilities of machine learning techniques relative to
  traditional alternatives, such as those based on constraint solving
  from the programming languages community.

  Our key contribution is the proposal of \langname, a domain-specific
  language for expressing program synthesis problems. \langname is
  similar to a probabilistic programming language: a model is composed
  of a specification of a program representation (declarations of
  random variables) and an interpreter that describes how programs map
  inputs to outputs (a model connecting unknowns to observations).
  The inference task is to observe a set of input-output examples and
  infer the underlying program. \langname has two main benefits. First,
  it enables rapid exploration of a range of domains, program
  representations, and interpreter models.  Second, it separates the
  model specification from the inference algorithm, allowing proper
  like-to-like comparisons between different approaches to inference. From a single
  \langname~specification we can automatically perform inference using
  four different back-ends that include machine learning and program synthesis approaches. These are based on
  gradient descent (thus each specification can be seen as a differentiable
  interpreter), linear program (LP) relaxations for graphical
  models, discrete satisfiability solving, and the \sketch~program
  synthesis system.

  We illustrate the value of \langname~by developing several interpreter
  models and performing an extensive
  empirical comparison between alternative inference algorithms
  on a variety of program models. Our key, and perhaps surprising, empirical finding is that
  constraint solvers dominate the gradient descent and
  LP-based formulations.
   We conclude with some suggestions on how the machine learning community can make progress on program synthesis.
\end{abstract}
\section{Introduction}

Learning computer programs from input-output examples, or Inductive
Program Synthesis (IPS), is a fundamental problem in computer science,
dating back at least to \cite{summers1977methodology} and
\cite{biermann1978inference}.  The field has produced many successes,
with perhaps the most visible example being the FlashFill system in
Microsoft Excel~\citep{gulwani2011automating,cacm12}.

Learning from examples is also studied extensively in the statistics
and machine learning communities.  Trained decision trees
and neural networks could be considered to be synthesized computer programs, but it
would be a stretch to label them as such. Relative to traditional computer programs, these models typically lack several features: (a) key
functional properties are missing, like the ability to interact with external
storage, (b) there is no compact, interpretable source code representation of the learned model (in the case of neural networks), and (c) there is no explicit control flow (e.g. \texttt{while} loops and \texttt{if} statements). The absence of a precise control flow is a particular hindrance as it can lead to poor generalization. For example, whereas natural computer programs are often built with the inductive bias to use control statements ensuring correct execution on inputs of arbitrary size, models like Recurrent Neural Networks can struggle to generalize
from short training instances to instances of arbitrary length.

Several models have already been proposed which start to address the functional differences between neural networks and computer programs. These include Recurrent Neural Networks (RNNs) augmented with a stack or queue memory~\citep{Giles89,Joulin15,grefenstette2015learning}, Neural Turing Machines~\citep{Graves14}, Memory Networks~\citep{Weston14}, Neural GPUs~\citep{kaiser2015neural},
Neural Programmer-Interpreters~\citep{Reed15}, and Neural Random Access Machines~\citep{Kurach15}. These models
combine deep neural networks with external memory, external computational primitives, and/or built-in structure that reflects a desired algorithmic structure in their execution. Furthermore, they have been been shown to be trainable by gradient descent. However, they do not fix all of the absences noted above. First, none of these models produce programs as output. That is, the representation of the learned model is not interpretable source code. Instead, the program is hidden inside ``controllers'' composed of neural networks that decide which operations to perform, and the learned ``program'' can only be understood in terms of the executions that it produces on specific inputs. Second, there is still no concept of explicit control flow in these models. 

These works raise questions of (a) whether new models can be designed specifically to synthesize interpretable source code that may contain looping and branching structures, and (b) whether searching over program space using techniques developed for training deep neural networks is a useful alternative to the combinatorial search methods used in traditional IPS. In this work, we make several contributions in both of these directions. 

To address the first question we develop models inspired by intermediate
representations used in compilers like LLVM \citep{lattner2004llvm} that can be trained by gradient descent. These models address all of the deficiencies highlighted at the beginning of this section: they interact with external storage, handle non-trivial control flow with explicit \code{if} statements and loops, and, when appropriately discretized, a learned model can be expressed
as interpretable source code. 
We note two
concurrent works, Adaptive Neural Compilation~\citep{Bunel16} and
Differentiable Forth~\citep{Riedel16}, which implement similar ideas. Each design choice when creating differentiable representations of source code has an effect on the inductive bias of the model and the difficulty
of the resulting optimization problem. Therefore, we seek a way of
rapidly experimenting with different formulations to allow us to explore the full space of modelling variations.

To address the second question, concerning the efficacy of gradient descent, we need a way of specifying an IPS problem such that the gradient based approach can be compared to a variety of alternative approaches in
a like-for-like manner.  These alternative approaches originate from both a rich history of IPS in the programming
languages community and a rich literature of techniques for inference in discrete graphical models in the machine learning community. To our knowledge, no such comparison has previously been performed.

These questions demand that we explore both a range of model variants and a range of search techniques on top of these models. Our answer to both of these issues is the same: \langname, a
new probabilistic programming language for specifying IPS problems.
\langname provides a means for describing an \emph{execution model} (e.g., a
Turing Machine, an assembly language, etc.) by defining a
parameterization (a program representation) and an interpreter that
maps inputs to outputs using the parametrized program.  This \langname
description is independent of any particular inference algorithm.
The IPS task is to infer the execution model parameters
(the program) given an execution model and pairs of inputs and
outputs.  To perform inference, \langname is automatically ``compiled'' into an
intermediate representation which can be fed to a particular inference
algorithm.  Interpretable source code can be obtained directly from the
inferred model parameters.  The
driving design principle for \langname is to strike a subtle balance
between the breadth of expression needed to precisely capture a range
of execution models, and the restriction of expression needed to
ensure that automatic compilation to a range of different back-ends is
tractable. 

\langname currently has four back-end inference algorithms, which are listed in
\rTab{tbl:backends}: gradient-descent (thus any \langname~model can be
viewed as a differentiable interpreter), (integer) linear program (LP)
relaxations, SMT, and the $\sketch$ program synthesis system  \citep{sketch}. To allow all of these back-ends to be used regardless of the specified execution model requires some generalizations and extensions of previous work. For the
gradient descent case, we generalize the approach taken by \cite{Kurach15},
lifting discrete operations to operate on discrete distributions,
which then leads to a differentiable system. For the linear program case,
we need to extend the standard LP relaxation for discrete graphical models
to support \code{if} statements. In \Secref{sec:lp-back-end},
we show how to adapt the ideas of \textit{gates} \citep{minka2009gates} to the
linear program relaxations commonly used in graphical model inference
\citep{shlezinger1976syntactic,werner2007linear,wainwright2008graphical}.
This could serve as a starting point for further work on LP-based
message passing approaches to IPS (e.g., following
\cite{sontag2008tightening}).

\begin{table}
\begin{tabular}{p{1.2in}p{0.6in}cp{2.6in}}
\toprule
Technique name & Family & Optimizer/Solver & Description\\
\midrule
FMGD\newline(Forward marginals, gradient descent) & Machine learning & TensorFlow & A gradient descent based approach which generalizes the approach used by \cite{Kurach15}.\\
(I)LP\newline((Integer) linear programming) & Machine learning & Gurobi & A novel linear program relaxation approach based on adapting standard linear program relaxations to support Gates \citep{minka2009gates}. \\
SMT\newline(Satisfiability modulo theories) & Program synthesis & Z3 & Translation of the problem into a first-order logical formula with existential constraints.\\
$\sketch$ & Program synthesis & $\sketch$ & View the \langname model as a partial program (the interpreter) containing holes (the source code) to be inferred according to a specification (the input-output examples).\\
\bottomrule
\end{tabular}
\caption{\label{tbl:backends}Overview of considered \langname back-end inference algorithms.}
\end{table}

Finally, having built \langname, it becomes possible to develop
understanding of the strengths and weaknesses of the alternative
approaches to inference.  To understand the limitations of using gradient descent
for IPS problems, we first use \langname to define a simple example
where gradient descent fails, but which the alternative back-ends
solve easily.  By studying this example we can better understand the
possible failure modes of gradient descent. We prove that 
there are exponentially many local optima in the example and show
empirically that they arise often in practice (although they can be
mitigated significantly by using optimization heuristics like adding
noise to gradients during training \citep{neelakantan2015adding}).  We
then perform a comprehensive empirical study comparing different
inference back-ends and program representations. We show that some
domains are significantly more difficult for gradient descent than
others and show results suggesting that gradient descent performs best
when given redundant, overcomplete parameterizations. However, the
overwhelming trend in the experiments is that the techniques from the
programming languages community outperform the machine learning
approaches by a significant margin.

In summary, our main contributions are as follows:
\begin{itemize}
\item A novel `Basic Block' execution model that enables learning programs with
  complex control flow (branching and loops).
\item \langname, a probabilistic programming language tailored to IPS, with back-end
  inference algorithms including techniques based on gradient descent, linear programming, and
  highly-efficient systems from the programming languages community (SMT and
  $\sketch$).
  \langname also allows ``program sketching'', in which a partial solution is
  provided to the IPS system.
  For this, some parameters of an execution model can simply be fixed, e.g. to
  enforce control flow of a specified shape.
\item A novel linear program relaxation to handle the \texttt{if} statement structure that is common in execution models, and a generalization of the smoothing technique from \cite{Kurach15} to work on any execution model expressible in \langname.
\item Analytic and experimental comparisons of different inference techniques for IPS and experimental comparisons of different modelling assumptions.
\end{itemize}

This report is arranged as follows: We briefly introduce the `Basic Block' model
in \Secref{sec:motivating example} to discuss what features \langname needs to
support to allow modeling of rich execution models.
In \Secref{sec:frontend} we describe the core \langname language and illustrate
how to use it to explore different modeling assumptions using several example
execution models.
These include a Turing Machine, Boolean Circuits, a RISC-like assembly language,
and our Basic Block model.
In \Secref{sec:backend} we describe the compilation of \langname models to the
four back-end algorithms listed in \rTab{tbl:backends}.
Quantitative experimental results comparing these back-ends on the
aforementioned execution models is presented in \Secref{sec:experiments}.
Finally, related work is summarized in \Secref{sec:relatedWork} and we discuss
conclusions and future work in \Secref{sec:discussion}.


%
\section{Motivating Example: Differentiable Control Flow Graphs}
\label{sec:motivating example}

As an introductory example, we describe a new execution model that we
would like to use for IPS. In this section, we describe the model at
a high level. In later sections, we describe how to express the model
in \langname~and how to perform inference.

\begin{figure}
\begin{center}
\includegraphics[width=0.35\textwidth]{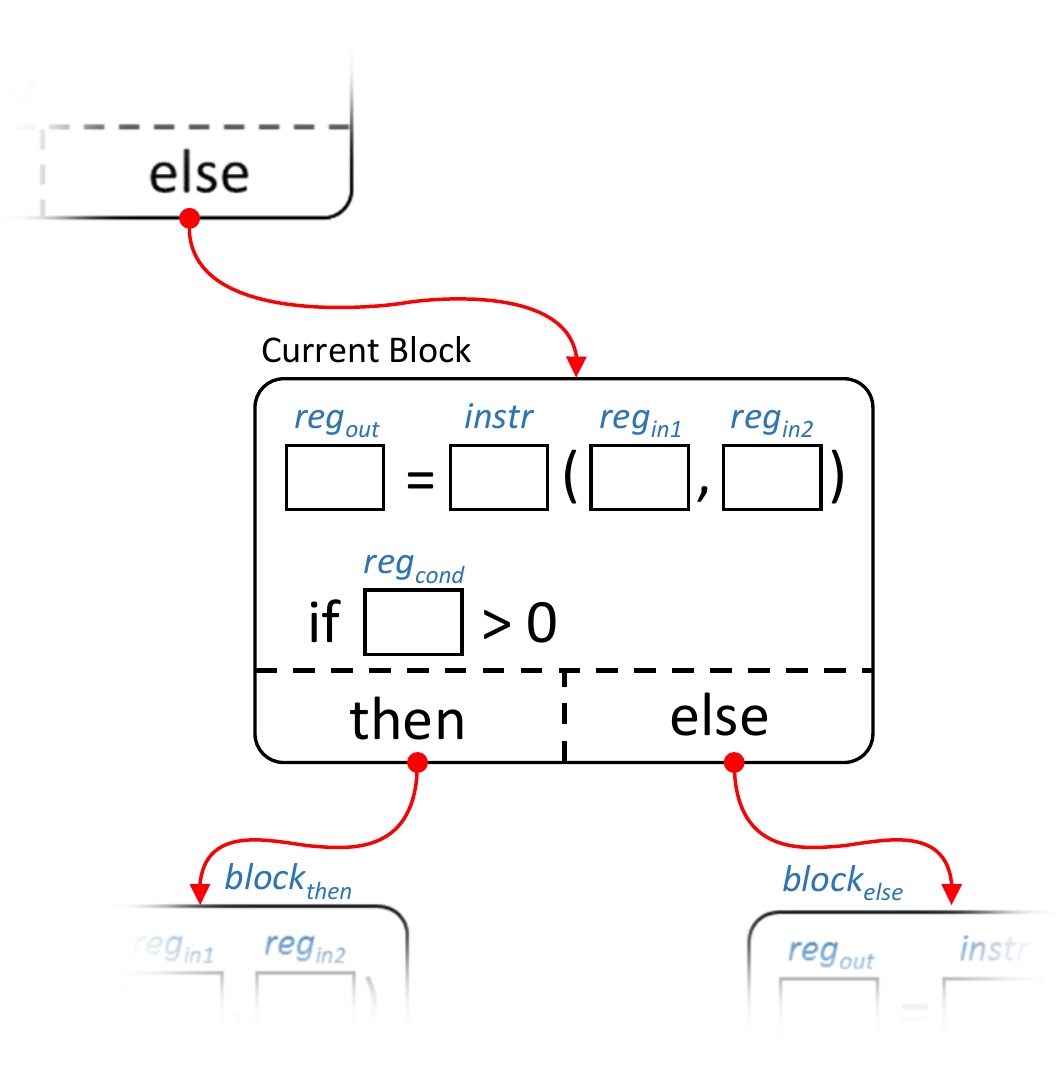}
\end{center}
\caption{\label{fig:basicBlock}
  Diagram of the Basic Block program representation. Empty boxes and connections denote per-block unknown
  parameters to be filled in by the inference algorithm. The choice of which block to go to in the \code{then}
  and \code{else} conditions are also unknown parameters.  The domain
  of unknown parameters is described in the small blue text.  An
  assignment to all unknown parameters yields a program.  }
\end{figure}

\emph{Control flow graphs} (CFGs) \citep{allen1970control} are a
representation of programs commonly used for static analysis and
compiler optimizations. They consist of a set of \emph{basic blocks},
which contain sequences of instructions with no jumps (i.e.,
straight-line code) followed by a jump or conditional jump instruction
to transfer control to another block. CFGs are expressive enough to
represent all of the constructs used in modern programming languages
like \texttt{C++}. Indeed, the intermediate representation of LLVM is based on
basic blocks.

Our first model is inspired by CFGs but is limited to use a restricted set
of instructions and does not support function calls. We refer to the model
as the \emph{Basic Block} model. An illustration of the model appears in 
\Figref{fig:basicBlock}.
In more detail, we specify a fixed number of blocks $B$, and we let
there be $R$ registers that can take on values $0, \ldots, M-1$.
We are given a fixed set of instructions that implement basic arithmetic
operations, like \code{ADD}, \code{INCREMENT}, and \code{LESS-THAN}.
An external memory can be written to and read from using special
instructions \code{READ} and \code{WRITE}.
There is an instruction pointer that keeps track of which block
is currently being executed. Each block has a single statement
parameterized by two argument registers, the instruction to be executed,
and the register in which to store the output. After the statement is
executed, a condition is checked, and a branch is taken. The condition is parameterized
by a choice of register to check for equality to 0 (C-style interpretation
of integers as booleans). Based upon the result, the instruction pointer
is updated to be equal to the \code{then} block or the \code{else} block.
The identities of these blocks are the parameterization of the branch
decision.

The model is set up to always start execution in block 0, and a special
\emph{end block} is used to denote termination. The program is executed
for a fixed maximum number of timesteps $T$.
To represent input-output examples, we can set an initial state of external memory, and assert that particular elements in the final memory should have the desired
value upon termination.

The job for \langname in this case is to precisely describe the execution model---how statements
are executed and the instruction pointer is updated---in a way which can be translated into a fully differentiable interpreter for the Basic Block language or into an intermediate representation for passing to other back-ends. In the next sections, we describe in more detail how \langname~execution models are specified and how the back-ends work.



\section{Front-end: Describing an IPS problem}
\label{sec:frontend}

\begin{figure}
\includegraphics[width=\textwidth]{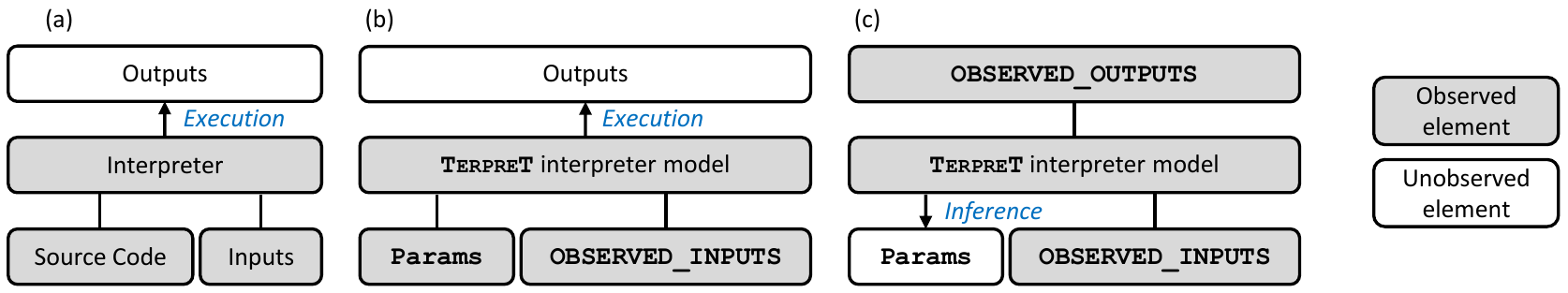}
\caption{A high level view of the program synthesis task. Forward execution of a traditional program interpreter is shown in (a) for analogy with the forward mode (b) and reverse mode (c) of a \langname IPS system.
\label{fig:highlevel}}
\end{figure}

One of our central aims is to disentangle the description of an execution model
from the inference task so that we can perform like-for-like comparisons between
different inference approaches to the same IPS task.
For reference, the key components for solving an IPS problem are illustrated in
\Figref{fig:highlevel}.
In the forward mode the system is analogous to a traditional interpreter, but in a
reverse mode, the system infers a representation of source code given only
observed outputs from a set of inputs.
Even before devising an inference method, we need both a means of parameterizing
the source code of the program, and also a precise description of the
interpreter layer's forward transformation.
This section describes how these modeling tasks are achieved in \langname.

  \subsection{The \langname~Probabilistic Programming Language}
\label{sec:language}

The full grammar for syntactically correct \langname programs is shown in
\Figref{fig:SyntaxDef}, and we describe the key semantic features of the language in the following sections.
For illustration, we use a running example of a simple automaton
shown in \Figref{fig:automaton1}.
In this example the `source code' is parameterised by a $2 \times 2$ boolean array, \texttt{ruleTable}, and we take as input the first two values on a binary tape of length $T$,
\texttt{\{tape[0], tape[1]\}}. The forward execution of the interpreter could be
described by the following simple \texttt{Python} snippet:

\vspace{5mm}
\begin{python}
for t in range(1, T - 1):
	tape[t + 1] = ruleTable[ tape[t - 1], tape [t] ]
\end{python}
\vspace{5mm}

\noindent Given an observed output, \texttt{tape[$T-1$]}, inference of a consistent \texttt{ruleTable}
is very easy in this toy problem, but it is instructive to analyse the \langname~implementation of
this automaton in the following sections. These sections describe variable declaration, control flow, user defined functions and handling of observations in \langname.

\newcommand{\nontermSym}[1]{\ensuremath{\mathsf{#1}}}
\newcommand{\termSym}[1]{\underline{\texttt{#1}}}

\newcommand{\bnfalt}{\;|\;}
\begin{figure}[t]
  \[
    \begin{array}{llcl}
        \text{Const Expr} & c & := & n \bnfalt v_c \bnfalt f(c_1,\cdots,c_k) \bnfalt c_0\;\mathit{op}_a\;c_1 \\
        \text{Arith Expr} & a & := & v \bnfalt c \bnfalt v[a_1,\cdots, a_k] \bnfalt a_0\;{op}_a \;a_1 \bnfalt f(a_0,\cdots, a_k) \\
        \text{Arith Op} & \mathit{op}_a & := & \code{+}\; \bnfalt \;\code{-}\; \bnfalt \;\code{*}\; \bnfalt \;\code{/}\; \bnfalt \;\code{\%} \\
        \text{Bool Expr} & b & := & a_0\; {op}_c\;a_1 \bnfalt \code{not}\; b \bnfalt b_0\;\code{and}\;b_1 \bnfalt b_0\;\code{or}\;b_1 \\
        \text{Comp Op} & \mathit{op}_c & := & \code{==} \bnfalt \code{<} \bnfalt \code{>} \bnfalt \code{<=} \bnfalt \code{>=} \\
        \text{Stmt} & s & := &
                    s_{0}\;\code{;}\; s_{1} \bnfalt \code{return} \; a \\
        &&\bnfalt & a_0\code{.set\_to(}a_1\code{)} \bnfalt a\code{.set\_to\_constant(}c\code{)} \bnfalt a_0\code{.observe\_value(}a_1\code{)}\\
        &&\bnfalt & \code{if}\; b_0\code{:} \;s_0\; \code{else:} \; s_1\\
        &&\bnfalt & \code{for}\; v \; \code{in} \; \code{range}(c_1)\: \code{:} \; s
          \bnfalt   \code{for}\; v \; \code{in} \; \code{range}(c_1,c_2)\: \code{:} \; s \\
        &&\bnfalt & \code{with} \; a \; \code{as} \; v \;\code{:}\; s \\
        \text{Decl Stmt} & s_d & := &
                    s_{d_0}\;\code{;}\; s_{d_1} \bnfalt v_c \;\code{=}\; c  \bnfalt v_c = \code{\#\_\_HYPERPARAM\_}v_c\code{\_\_} \\
        &&\bnfalt & v \;\code{=}\; \code{Var(}c\code{)} \bnfalt v \;\code{=}\; \code{Var(}c\code{)[}c_1\code{,}\cdots\code{,}c_k\code{]}\\
        &&\bnfalt & v \;\code{=}\; \code{Param(}c\code{)} \bnfalt v \;\code{=}\; \code{Param(}c\code{)[}c_1\code{,}\cdots\code{,}c_k\code{]}\\
        &&\bnfalt & \code{@CompileMe([}c_1\code{,}\cdots\code{,}c_k\code{],} c_r\code{);  def}\; f\code{(}v_0\code{,}\cdots\code{,}v_k\code{):} s\\
        \text{Input Decl} & s_i & := & \code{\#\_\_IMPORT\_OBSERVED\_INPUTS\_\_} \\
        \text{Output Decl} & s_o & := & \code{\#\_\_IMPORT\_OBSERVED\_OUTPUTS\_\_}     \\
        \text{Program} & p & := & s_d; s_i; s; s_o
    \end{array}
  \]
    \caption{\label{fig:SyntaxDef}The syntax of \langname, using natural numbers $n$, variable names
      $v$, constant names $v_c$ and function names $f$.}
    \label{langsyntax}
\end{figure}

\subsubsection{Declarations and Assignments}
\begin{figure}
\begin{tabular}{p{4in}p{1.7in}}
{\begin{python}
const_T = 5

#######################################################
#  Source code parametrisation                        #
#######################################################
ruleTable = Param(2)[2, 2]

#######################################################
#  Interpreter model                                  #
#######################################################
tape = Var(2)[const_T]

#__IMPORT_OBSERVED_INPUTS__
for t in range(1, const_T - 1):
    with tape[t] as x1:
        with tape[t - 1] as x0:
            tape[t + 1].set_to(ruleTable[x0, x1])
#__IMPORT_OBSERVED_OUTPUTS__
\end{python}} & \raisebox{-0.9\height}{\includegraphics[width=1.7in]{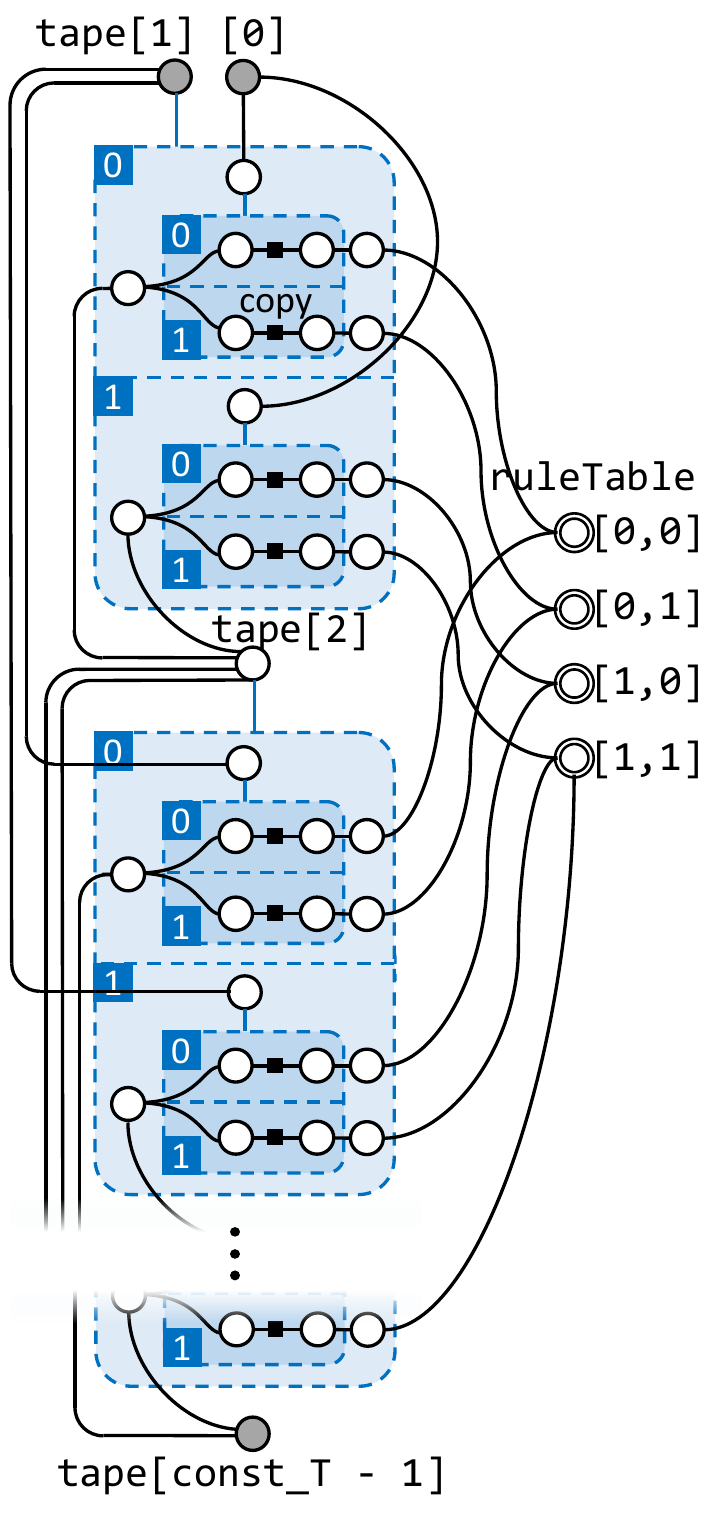}}
\end{tabular}
\caption{Illustrative example of a \langname~script and corresponding factor graph which describe a toy automaton that updates a binary \texttt{tape} according to the previous two entries and a \texttt{rule} (refer to \Figref{fig:factorGraphParts} for definition of graphical symbols).
\label{fig:automaton1}
}
\end{figure}

We allow declarations to give names to ``magic'' constants, as in
line 1 of \Figref{fig:automaton1}.
Additionally, we allow the declaration of \emph{parameters} and \emph{variables},
ranging over a finite domain $0, 1, \ldots N-1$ using \texttt{Param($N$)} and
\texttt{Var($N$)}, where $N$ has to be a compile-time constant (i.e., a natural
number or an expression over constants).
Parameters are used to model the source code to be inferred, whereas variables
are used to model the computation (i.e., intermediate values).
For convenience, (multi-dimensional) arrays of variables can be declared using
the syntax \texttt{foo = Var($N$)[$\mathit{dim}_1$, $\mathit{dim}_2$, \ldots]}, and accessed as \texttt{foo[$\mathit{idx}_1$, $\mathit{idx}_2$, \ldots]}. Similar syntax is available for \code{Param}s. 
These arrays can be unrolled during compilation such that unique symbols
representing each element are passed to an inference algorithm, i.e., they do
not require special support in the inference backend.
For this reason, dimensions $\mathit{dim}_i$ and indices $\mathit{idx}_i$ need to be compile-time constants.
Example variable declarations can be seen in lines 6 and 11 of
\Figref{fig:automaton1}.

Assignments to declared variables are not allowed via the usual assignment
operator (\texttt{$\mathit{var}$ = $\mathit{expr}$}) but are instead written as
\code{$\mathit{var}$.set\_to($\mathit{expr}$)}, to better distinguish them
from assignments to constant variables. Static single assignment (SSA) form is enforced, and it is only legal for a variable to appear multiple times as the target of \code{set\_to} statements if each assignment appears in different cases of a conditional block.
Because of the SSA restriction, a variable can only be written once. However, note that programs that perform multiple writes to a given variable can always be translated to their corresponding SSA forms. 

\subsubsection{Control flow}
\langname{} supports standard control-flow structures such as \texttt{if-else}
(where \code{elif} is the usual shorthand for \code{else if}) and \texttt{for}.
In addition, \langname uses a unique \texttt{with} structure.
The need for the latter is induced by our requirement to only use compile-time
constants for accessing arrays.
Thus, to set the \texttt{2}nd element of \texttt{tape} in our toy example (i.e.,
the first step of the computation), we need code like the following to access
the values of the first two values on the tape:

\vspace{5mm}
\begin{python}
if tape[1] == 0:
       if tape[0] == 0:
               tape[2].set_to(ruleTable[0,0])
       elif tape[0] == 1:
               tape[2].set_to(ruleTable[1,0])
elif tape[1] == 1:
       if tape[0] == 0:
               tape[2].set_to(ruleTable[0,1])
       elif tape[0] == 1:
               tape[2].set_to(ruleTable[1,1])
\end{python}
\vspace{5mm}

Intuitively, this snippet simply performs case analyses over all possible
values of \texttt{tape[1]} and \texttt{tape[0]}.
To simplify this pattern, we introduce the \texttt{with $\mathit{var}$ as $id$: $\mathit{stmt}$}
control-flow structure, which allows to automate this unrolling, or avoid it for
back-ends that do not require it (such as Sketch).
To this end, all possible possible values $0, \ldots, N-1$ of $\mathit{var}$ (known from
its declaration) are determined, and the \texttt{with}-statement is transformed
into
 \texttt{if $id$ == 0 then:
           $\mathit{stmt}[\mathit{var}/0]$;
         elif $id$ == 1 then:
           $\mathit{stmt}[\mathit{var}/1]$;
         \ldots
         elif $id$ == $n$ then:
           $\mathit{stmt}[\mathit{var}/(N-1)]$;},
where $\mathit{stmt}[\mathit{var}/n]$ denotes the statement $\mathit{stmt}$ in which all occurrences of the
variable $\mathit{var}$ have been replaced by $n$. Thus, the snippet from above can be
written as follows.

\vspace{5mm}
\begin{python}
with tape[1] as x1:
       with tape[0] as x0:
               tape[2].set_to(ruleTable[x0,x1])
\end{python}
\vspace{5mm}

In \langname{}, \texttt{for} loops may only take the shape
 \texttt{for $id$ in range($c_1$, $c_2$): $\mathit{stmt}$},
where $c_1$ and $c_2$ are compile-time constants.
Similar to the \texttt{with} statement, we can unroll such loops explicitly during
compilation, and thus if the values of $c_1$ and $c_2$ are $n_1$ and $n_2$, we
generate
\texttt{$\mathit{stmt}[id/n_1]$; $\mathit{stmt}[id/(n_1+1)]$; \ldots; $\mathit{stmt}[id/(n_2-1)]$}.
Using the \texttt{with} and \texttt{for} statements, we can thus describe the
evaluation of our example automaton for \texttt{const\_T} timesteps as shown in
lines 14-17 of \Figref{fig:automaton1}.

\subsubsection{Operations}
\langname{} supports user-defined
functions to facilitate modelling interpreters supporting non-trivial instruction sets.
For example,
  \texttt{bar($\mathit{arg}_1$,$\ldots$,$\mathit{arg}_M$)} will apply the
function \texttt{bar} to the arguments $\mathit{arg}_1$,$\ldots$,$\mathit{arg}_M$.
The function \texttt{bar}$:\mathbb{Z}^M\to \mathbb{Z}$ can be defined as a
standard \texttt{Python} function with the additional decoration
\texttt{@CompileMe($in\_domains$, $out\_domain$)}, specifying the domains of the
input and output variables.

\begin{figure}[t!]
\begin{tabular}{p{4in}p{2in}}
%
%
%
%
{\begin{python}
const_T = 5

@CompileMe([2, 2], 3)
def add(a, b):
    s = a + b
    return s

#######################################################
#  Source code parametrisation                        #
#######################################################
ruleTable = Param(2)[3]

#######################################################
#  Interpreter model                                  #
#######################################################
tape = Var(2)[const_T]
tmpSum = Var(3)[const_T - 1]

#__IMPORT_OBSERVED_INPUTS__
for t in range(1, const_T - 1):
    tmpSum[t].set_to(add(tape[t - 1], tape[t]))
    with tmpSum[t] as s:
        tape[t + 1].set_to(ruleTable[s])
#__IMPORT_OBSERVED_OUTPUTS__
\end{python}}& \raisebox{-1.1\height}{\includegraphics[width=1.7in]{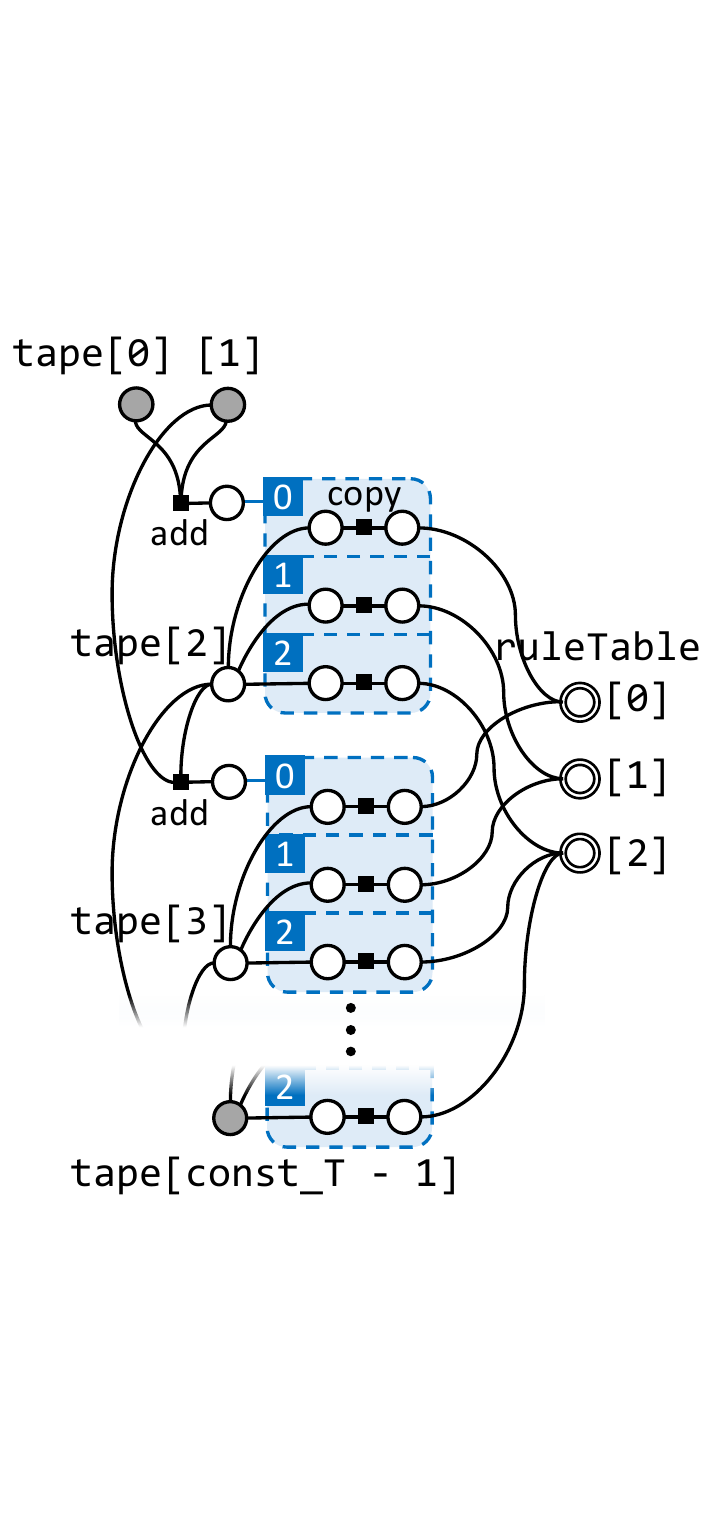}}
\end{tabular}
\caption{An example \langname~script and the corresponding factor graph which describe a toy automaton that updates a binary \texttt{tape} according to the previous two entries and a \texttt{rule}.
\label{fig:automaton2}
}
\end{figure}

To illustrate this feature, \Figref{fig:automaton2} shows variation of the running example where the automaton updates the \code{tape} according to a \code{ruleTable} which depends only on the \textit{sum} of the preceding two entries. This is implemented using the function \code{add} in lines 3-6. Note that we use standard \texttt{Python} to define this function and
leave it up to the compiler to present the function appropriately to the inference
algorithm.


\subsubsection{Modelling Inputs and Outputs}
\label{sec:nohtypObservations}
Using statements from the preceding sections, an execution model can be fully specified, and we now connect this model to input/output
observations to drive the program induction.
To this end, we use the statements \texttt{set\_to\_constant}
(resp. \texttt{observe\_value}) to model program input (resp. program output).
Thus, a single input-output observation for the running example could be written in \langname{} as
follows.

\vspace{5mm}
\begin{python}
# input
tape[0].set_to_constant(1)
tape[1].set_to_constant(0)

# output
tape[const_T - 1].observe_value(1)
\end{python}
\vspace{5mm}

To keep the execution model and the observations separate, we store the observation snippets in a separate file and use preprocessor directives \code{\#\_\_IMPORT\_OBSERVED\_*\_\_} to pull in the appropriate snippets before compilation (see lines 13 and 18 of \Figref{fig:automaton1}). We also allow any constant literals to be stored separately from the \langname execution model, and we import these values using preprocessor directives of the form \code{$v_c$ = \#\_\_HYPERPARAM\_\,$v_c$\,\_\_}.

In general, we want to infer programs from $n_{\rm obs}>1$ input-output
examples.
The simplest implementation achieves this by augmenting each \texttt{Var}
declaration with an additional array dimension of size $n_{\rm obs}$ and
wrapping the execution model in a \texttt{for} loop over the examples.
Examples of this are the outermost loops in the models in
\Appref{app:models}.


  \subsection{Example Execution Models}
\label{sec:models}

To illustrate the versatility of \langname, we use it to describe four
example execution models.
Broadly speaking, the examples progress from more abstract execution
models towards models which closely resemble assembly languages for RISC machines.

In each case, we present the basic model and fill in three representative synthesis tasks in \rTab{tab:benchmarkTasks} to
investigate. In addition, we provide the metrics for the ``difficulty" of each
task calculated from the minimal computational resources required in a
solution. Since the difficulty of a synthesis problem generally depends on the chosen inference
algorithm these metrics are primarily intended to give a sense of the scale
of the problem. The first difficulty metric, $D$, is the number of structurally distinct (but not necessarily functionally distinct) programs which would have to be
enumerated in a worst-case brute-force search, and the second metric, $T$, is the
unrolled length of all steps in the synthesized program. 

\subsubsection{Automaton: Turing Machine}
\label{sec:turing}
A Turing machine consists of an infinite tape of memory cells which each contain
one of $S$ symbols, and a head which moves over the tape in one of $H+1$ states (one state is the special \code{halt} case).
At each execution step, while the head is in an unhalted state $h_t$, it reads
the symbol $s_t$ at its current position, $x_t$, on the tape, then it writes the
symbol \code{newValue[$s_t$,$h_t$]} to position $x_t$, moves in the direction
specified by \code{direction[$s_t$,$h_t$]} (one cell left or right or no move) and
adopts a new state \code{newState[$s_t$,$h_t$]}.
The source code for the Turing machine is the entries of the control tables
\code{newValue}, \code{direction} and \code{newState}, which can be in any of
$D=\left[3 S (H+1)\right]^{S H}$ configurations.

We modify the canonical Turing machine to have a circular tape of finite length,
$L$, as described in the \langname model in \Appref{app:turingModel}. For each of our examples, we represent the symbols on the tape as $\{\code{0}, \code{1}, \code{blank}\}$.


\begin{table}
\vspace{-1cm}
\begin{center}
\begin{tabular}{p{1.3in}ccccp{3.1in}}
\toprule
TURING MACHINE & $H$ & $L$ & $\log_{10}D$ & $T$ & Description\\
\midrule
Invert & 1 & 5 & 4 & 6 & Move from left to right along the tape and invert all the binary symbols, halting at the first \code{blank} cell.\\
Prepend zero & 2 & 5 & 9 & 6 & Insert a ``\code{0}" symbol at the start of the tape and shift all other symbols rightwards one cell. Halt at the first \code{blank} cell.\\
Binary decrement & 2 & 5 & 9 & 9 &  Given a tape containing a binary encoded
number $b_{\rm in}>0$ and all other cells \code{blank}, return a tape
containing a binary encoding of $b_{\rm in}-1$ and all other cells
\code{blank}.\\
\bottomrule
\end{tabular}
\begin{tabular}{p{1.57in}cccp{3.1in}}
BOOLEAN CIRCUITS& $R$ & $\log_{10}D$ & $T$ & Description\\
\midrule
2-bit controlled shift register & 4 & 10 & 4 & Given input registers $(r_1,r_2,r_3)$, output $(r_1,r_2,r_3)$ if $r_1 == 0$ otherwise output $(r_1,r_3,r_2)$ (i.e. $r_1$ is a control bit stating whether $r_2$ and $r_3$ should be swapped).\\
full adder & 4 & 13 & 5 &  Given input registers $(c_{\rm in},a_1,b_1)$ representing a carry bit and two argument bits, output a sum bit and carry bit $(s,c_{\rm out})$, where $s+2c_{\rm out}=c_{\rm in}+a_1+b_1$.\\
2-bit adder & 5 & 22 & 8 &   Perform binary addition  on two-bit numbers: given
registers $(a_1,a_2,b_1,b_2)$, output $(s_1,s_2,c_{\rm out})$ where
$s_1+2s_2+4c_{\rm out}=a_1+b_1+2(a_2+b_2)$.\\
\bottomrule
\end{tabular}
\begin{tabular}{p{1in}cccccp{3.1in}}
BASIC BLOCK & $M$ & $R$ & $B$ & $\log_{10}D$ & $T$ & Description\\
\midrule
Access & 5 & 2 & 5 & 14 & 5 & Access the $k^{\rm th}$ element of a contiguous array. Given an initial heap 
$\code{heap}_0\code{[0]} = k$, 
$\code{heap}_0\code{[$1:J+1$]}=\code{A[:]} $
and $\code{heap}_0\code{[$J+1$]} = 0$, 
where $\code{A[$j$]}\in\{1,...,M-1\}$ for $0 \leq j < J$, $J+1 < M$ and $0\leq k < J$, terminate with $\code{heap[0]} = \code{A[$k$]}$.\\
Decrement & 5 & 2 & 5 & 19 & 18 & Decrement all elements in a contiguous array. Given an initial heap $\code{heap}_0\code{[k]} \in \{2,...,M-1\}$ for $0\leq k<K<M$ and $\code{heap}_0\code{[$K$]} = 0$, terminate with $\code{heap[k]} = \code{heap}_0\code{[k]}-1$.\\
List-K & 8 & 2 & 8 & 33 & 11 & Access the $k^{\rm th}$ element of a linked list. The initial heap is $\code{heap}_0\code{[0]} = k$, $\code{heap}_0\code{[1]} = p$, and $\code{heap}_0\code{[2:M]} = \code{linkList}$, where \code{linkList} is a linked list represented in the heap as adjacent [next pointer, value] pairs in random order, and $p$ is a pointer to the head element of \code{linkList}. Terminate with $\code{heap[0]}=\code{linkList[$k$].value}.$
\\
\bottomrule
\end{tabular}
\begin{tabular}{p{1in}cccccp{3.1in}}
ASSEMBLY & $M$ & $R$ & $B$ & $\log_{10}D$ & $T$ & Description\\
\midrule
Access & 5 & 2 & 5 & 13 & 5 &\\
Decrement & 5 & 2 & 7 & 20 & 27 & \hspace{1cm}As above.\\
List-K & 8 & 2 & 10 & 29 & 16 &\\
\bottomrule
\end{tabular}
\end{center}
\caption{\label{tab:benchmarkTasks} Overview of benchmark problems, grouped by execution model. For each benchmark we manually find the minimal feasible resources (e.g. minimum number of registers, Basic Blocks, timesteps etc.). These are noted in this table and we try to automatically solve the synthesis task with these minimal settings. }
\end{table}

\subsubsection{Straight-line programs: Boolean Circuits}
\label{sec:digital}
As a more complex model, we now consider a simple machine capable of performing
a sequence of logic operations (\code{AND}, \code{OR}, \code{XOR},
\code{NOT}, \code{COPY}) on a set of registers holding boolean values.
Each operation takes two registers as input (the second register is ignored in
the \code{NOT} and \code{COPY} operation), and outputs to one register, reminiscent of
standard three-address code assembly languages.
To embed this example in a real-world application, analogies linking the
instruction set to electronic logic gates and linking the registers to
electronic wires can be drawn.
This analogy highlights one benefit of interpretability in our model: the
synthesized program describes a digital circuit which could easily be translated
to real hardware (see e.g. \Figref{fig:overcompleteCircuit}). The \langname implementation of this execution model is shown in \Appref{app:booleanModel}.

There are $D=H^TR^{3T}$ possible programs (circuits) for a model
consisting of $T$ sequential instructions (logic gates) each chosen from the set of $H=5$ possible operations acting on $R$ registers (wires).


\subsubsection{Loopy programs 1: Basic block model}
\label{sec:bblock}

To build loopy execution models, we take inspiration from compiler intermediate languages (e.g., LLVM
Intermediate Representation), modeling full programs as graphs of ``basic
blocks''.
Such programs operate on a fixed number of registers, and a byte-addressable
heap store accessible through special instructions, \code{READ} and \code{WRITE}.
Each block has an instructions of the form
 \code{$\mathit{reg}_{\mathrm{out}}$ = $\mathit{instr}$ $\mathit{reg}_{\mathrm{in} 1}$ $\mathit{reg}_{\mathrm{in} 2}$},
followed by a branch decision 
 \code{if $\mathit{reg}_{\mathrm{cond}} > 0$
              goto $\mathit{block}_{\mathrm{then}}$
         else goto $\mathit{block}_{\mathrm{else}}$} (see \Figref{fig:basicBlock}, and the \langname model in \Appref{app:bbModel}).
This representation can easily be transformed back and forth to higher-level program source
code (by standard compilation/decompilation techniques) as well as into executable machine
code.

We use an instruction set containing $H=9$ instructions:
\code{ZERO}, \code{INC}, \code{DEC}, \code{ADD}, \code{SUB}, \code{LESSTHAN}, \code{READ}, \code{WRITE} and \code{NOOP}. This gives $D=[HR^4(B+1)^2]^B$ possible programs for a system with $R$ registers and $(B+1)$ basic blocks (including a special stop block which executes \code{NOOP} and redirects to itself). We consider the case where registers and heap memory cells all store a single data type - integers in the range $0,..,M-1$, where $M$ is the number of memory cells on the heap. This single data type allows both intermediate values and pointers into the heap to be represented in the registers and heap cells.

While this model focuses on interpretability, it also builds on an observation
from the results of \cite{Kurach15}.
In NRAMs, a RNN-based controller chooses a short sequence of instructions to
execute next based on observations of the current program state.
However, the empirical evaluation reports that correctly trained models usually
picked one sequence of instructions in the first step, and then repeated another
sequence over and over until the program terminates.
Intuitively, this corresponds to a loop initialization followed by repeated
execution of a loop body, something which can naturally be expressed in
the Basic Block model.

\subsubsection{Loopy programs 2: Assembly model}
\label{sec:assembly}
\begin{figure}
\begin{center}
\includegraphics[width=0.4\textwidth]{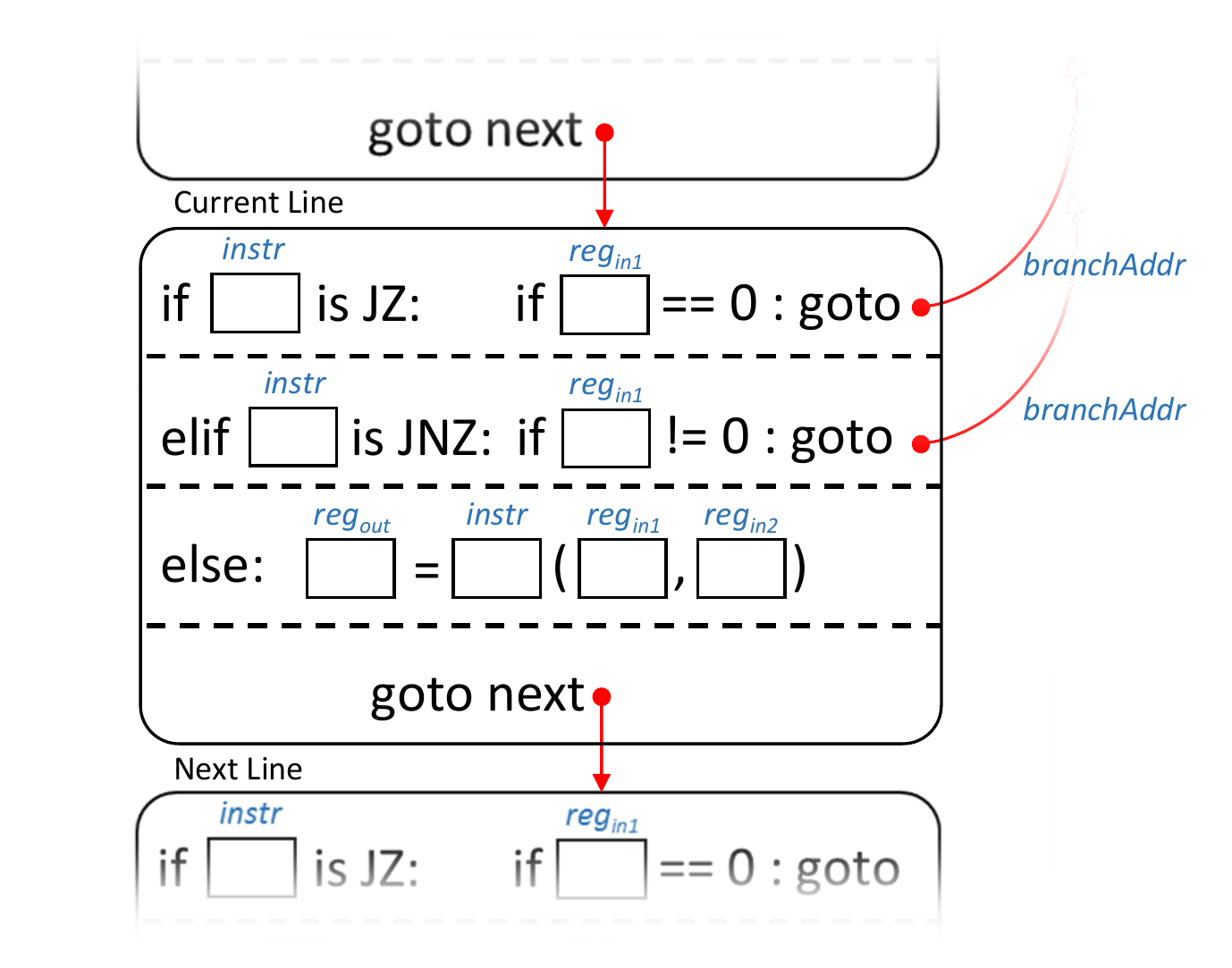}
\end{center}
\caption{\label{fig:assembly}
  Diagram of the assembly program representation. We present the model using the same graphical style as the Basic Block model in \Figref{fig:basicBlock}. }
\end{figure}

In the basic block model every expression is followed by a conditional branch, giving the model great freedom to represent rich control flow graphs. However, useful programs often execute a sequence of \textit{several} expressions between each branch. Therefore, it may be beneficial to bias the model to create chains of sequentially ordered basic blocks with only occasional branching where necessary. This is achieved by replacing the basic blocks with objects which more closely resemble lines of assembly code. The instruction set is augmented with the jump statements jump-if-zero~(\code{JZ}$(reg_{\rm in1}):branchAddr$), and jump-if-not-zero~(\code{JNZ}$(reg_{\rm in1}):branchAddr$), the operation of which are shown in \Figref{fig:assembly} (and in the \langname code in \Appref{app:assemblyModel}). Each line of code acts like a conditional branch only if the assigned $instr\in\{\code{JZ},\code{JNZ}\}$ otherwise it acts like a single expression which executes and passes control to the next line of code. This assembly model can express the same set of programs as the basic block model, and serves as an example of how the design of the model affects the success of program inference. 

In addition, we remove \code{NOOP} from the instruction set (which can be achieved by a jump operation pointing to the next line) leaving $H=10$ instructions, and we always include a special stop line as the $(B+1)^{\rm th}$ line of the program. The total size of the search space is then $D=[HR^3(B+1)]^B$.

%
%
%
%
%



\section{Back-ends: Solving the IPS problem}
\label{sec:backend}

\langname is designed to be compiled to a variety of intermediate representations for handing to different inference algorithms. This section outlines the compilation steps for each of the back-end algorithms listed in \rTab{tbl:backends}.

\begin{figure}[t]
\begin{tabular}{m{2.1in}m{2in}c}
\toprule
Graph element & \langname representation & Graphical representation \\
\midrule
Random variable (intermediate) & \texttt{Xi = Var($N$)} & \includegraphics[height=5mm]{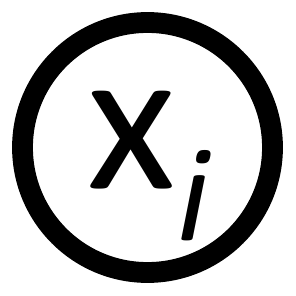} \\
Random variable (inference target) & \texttt{Xi = Param($N$)} & \includegraphics[height=5mm]{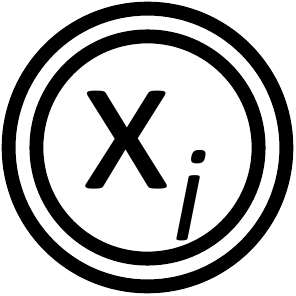} \\[20pt]

Observed variable (input) & \texttt{Xi.set\_to\_constant(x$^*$)} & \multirow{2}{*}{\includegraphics[height=5.8mm]{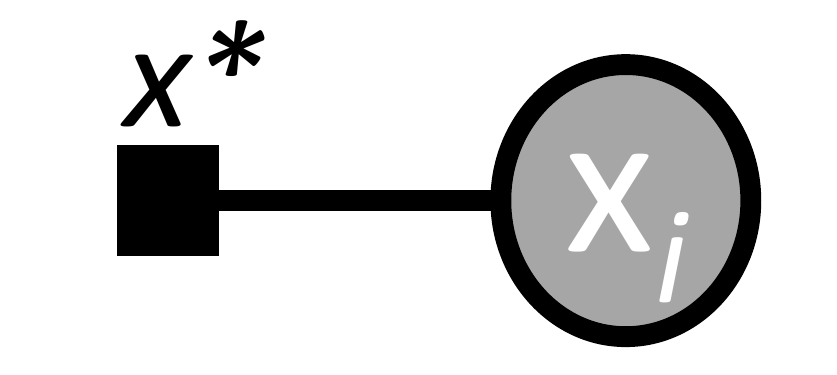}} \\
Observed variable (output) & \texttt{Xi.observe\_value(x$^*$)} & \\[20pt]

Factor (copy) & \texttt{X0.set\_to(X1)} & \raisebox{-.3\height}{\includegraphics[height=6.7mm]{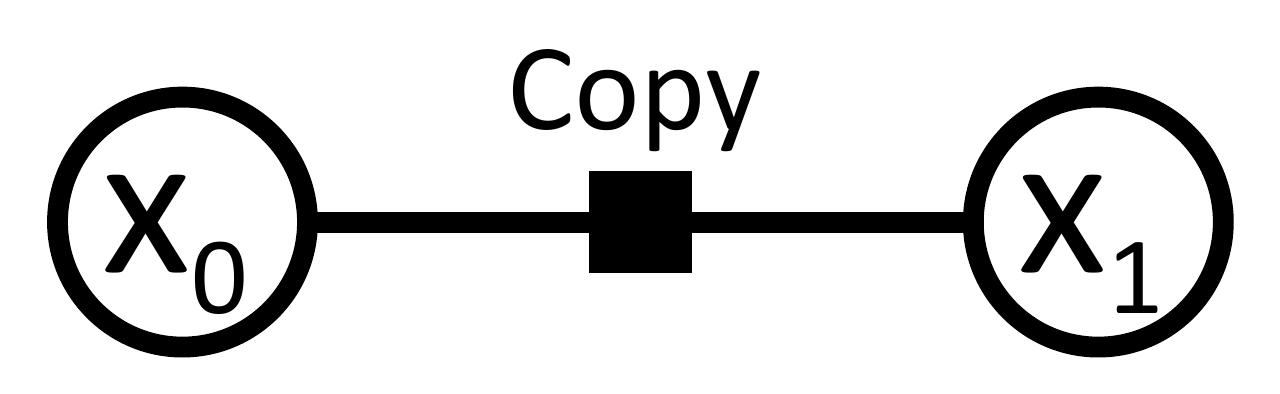}} \\[5pt]
Factor (general) & \texttt{X0.set\_to($f$(X1,X2,$\ldots$))} & \raisebox{-.5\height}{\includegraphics[height=19.2mm]{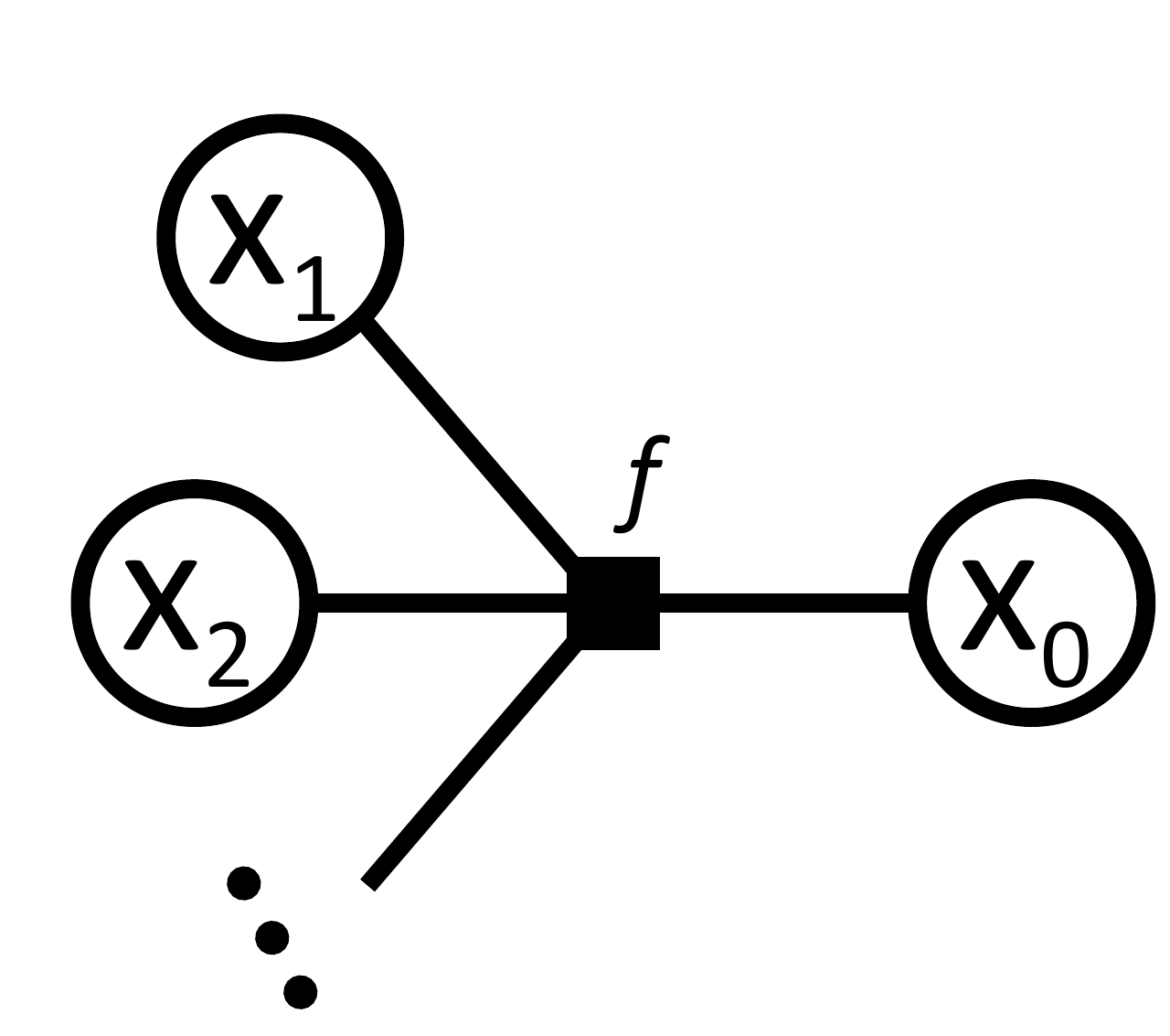}}\\[40pt]

Gates & \texttt{if C == 0:} \newline 
$\phantom{.}\quad stmt_0$\newline 
\texttt{elif C == 1:}\newline
$\phantom{.}\quad stmt_1$\newline
\texttt{elif C == 2:}\newline
$\phantom{.}\quad stmt_2$\newline
$\ldots$\newline
\texttt{elif C == n:}\newline
$\phantom{.}\quad stmt_n$\newline  & \raisebox{-.5\height}{\includegraphics[height=35mm]{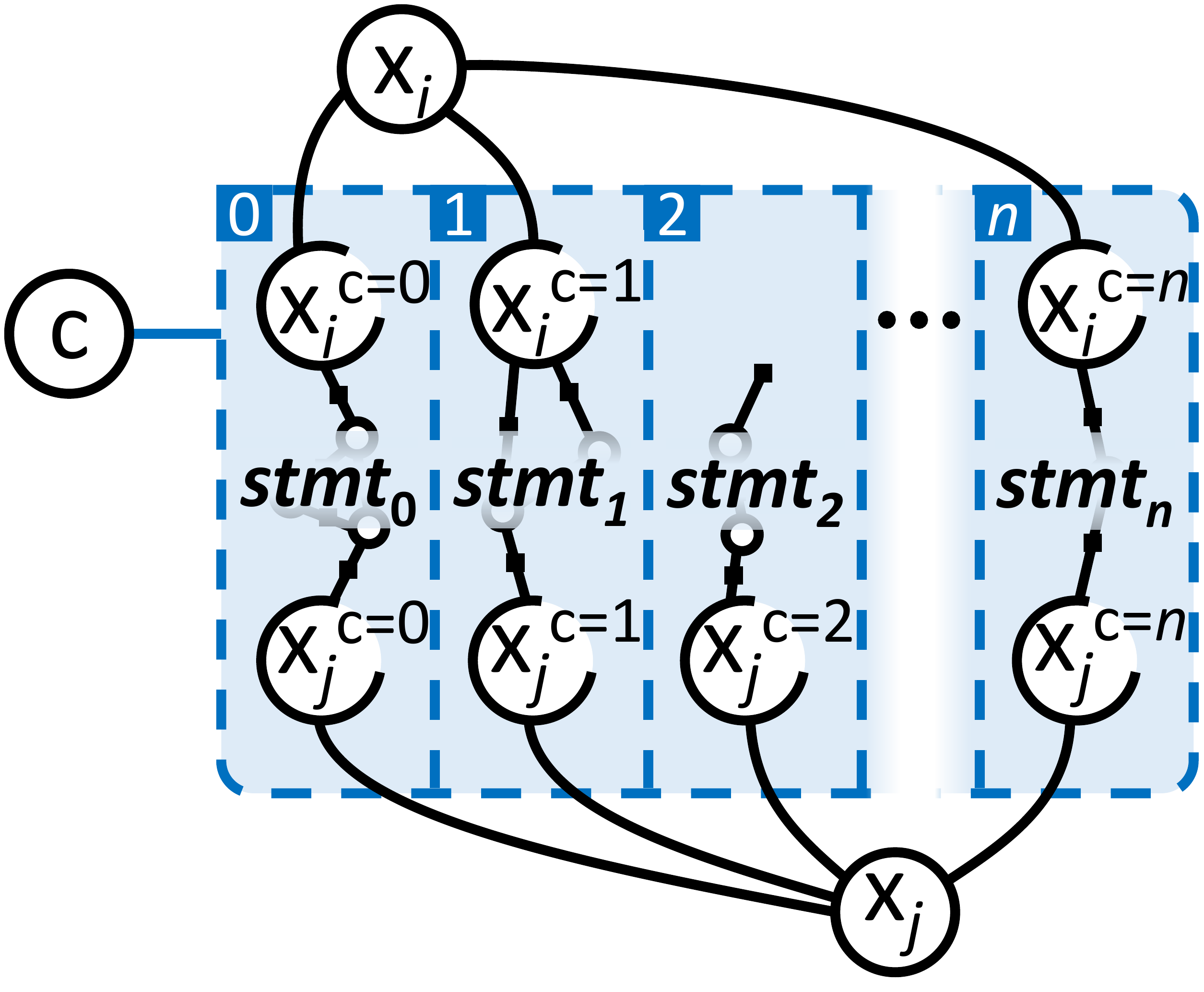}}\\
\bottomrule
\end{tabular}
\caption{The main \langname primitives and their corresponding graphical representation.
\label{fig:factorGraphParts}}
\end{figure}

For each back-end we present the compiler transfomation of the \langname
primitives listed in \Figref{fig:factorGraphParts}. For some back-ends, we find it useful to present
these transformations via an intermediate \emph{graphical} representation
resembling a factor graph, or more specifically, a gated factor graph \citep{minka2009gates},
which visualises the \langname program. Below we describe gated factor graphs
and provide the mapping from \langname syntax to primitives in these models. Then in \Secref{sec:fmgd} - \ref{sec:sketch} we show how to compile \langname for each
back-end solver.

\subsection{\langname for Gated Factor Graph Description}
\label{sec:factorGraphs}
A factor graph is a means of representing the factorization of a complex function or probability distribution into a composition of simpler functions or distributions. In these graphs, inputs, outputs and intermediate results are stored in \textit{variable} nodes linked by \textit{factor} nodes describing the functional relationships between variables. A \langname model defines the structure of a factor graph, and an inference algorithm is used to populate the variable nodes with values consistent with observations. 

\begin{figure}[t!]
\begin{tabular}{p{3in}p{3in}}
{\begin{python}

  # X4 =  0        if X0 == 0 and X1 == 0
  #     | X2 + 1   if X0 == 0 and X1 == 1
  #     | 2*X2     if X0 == 1
  #
  # Observe X4 = 5; infer X0, X1, X2

  @CompileMe([2, 10], 10)
  def Plus(a, b): return (a + b) % 10
  @CompileMe([10], 10)
  def MultiplyByTwo(a): return (2 * a) % 10

  X0 = Param(2); X1 = Param(2); X2 = Param(10)
  X3 = Var(10);  X4 = Var(10)

  if X0 == 0:
    if X1 == 0:
      X3.set_to(0); X4.set_to(X3)
    elif X1 == 1:
      X3.set_to(Plus(X1, X2)); X4.set_to(X3)
  elif X0 == 1:
    X4.set_to(MultiplyByTwo(X2))

  X4.observe_value(5)
\end{python}} & 
\begin{center}
\raisebox{-2.7in}{\includegraphics[width=2.3in]{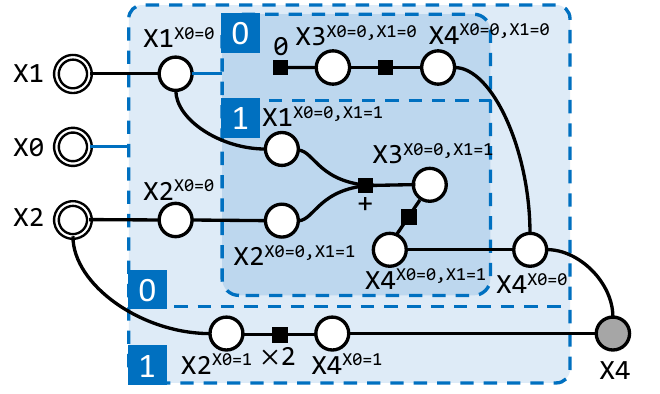}}
\end{center}
\end{tabular}
\caption{Interpreting \langname as a gated factor graph description. We model the inference task shown in lines 1-6 using \langname and provide the corresponding gated factor graph using symbols from \Figref{fig:factorGraphParts}. The solution to this inference task is \code{X0 = 0}, \code{X1 = 1}, and \code{X2 = 4}).
\label{fig:gatesExample}}
\end{figure}

Particular care is needed to describe factor graphs containing conditional branches since the value of a variable $X_i$ in conditions of the form \code{$X_i$ == $c$} is not known until inference is complete. This means that we must explore all branches during inference. \textit{Gated} factor graphs can be used to handle these \code{if} statements, and we introduce additional terminology to describe these gated models below. Throughout the next sections we refer to the \langname snippet shown in \Figref{fig:gatesExample} for illustration.

\paragraph{Local unary marginal. } We restrict attention to the case where each variable $X_i$ is discrete, with finite domain $\mcX_i = \{0, \ldots, N_i-1\}$. For each variable we instantiate a \emph{local unary marginal} $\mu_{i}(x)$ defined on the support $x \in \mcX_i$. In an \emph{integral} configuration, we demand that $\mu_{i}(x)$ is only non-zero at a particular value $x^*_i$, allowing us to interpret $X_i = x^*_i$. Some inference techniques relax this constraint and consider a continuous model $\mu_{i}(x) \in \reals \; \forall x\in \mcX_i$. In these relaxed models, we apply continuous optimization schemes which, if successful, will converge on an interpretable integral solution.

\paragraph{Gates. } Following \cite{minka2009gates}, we refer to \code{if} statements
as \emph{gates}. More precisely, an \code{if} statement consists of a condition (an expression
that evaluates to a boolean) and a body (a set of assignments or factors).
We will refer to the condition as the \emph{gate condition} and the body as the \emph{gate body}. In this work, we restrict attention to cases where all gate conditions are of the form $X_i == ConstExpr$. In future work we could relax this restriction.

In the example in \Figref{fig:gatesExample}, there is a nested gate structure. At the outer-most level, there are two gates with gate conditions \code{(X0 == 0)} (lines 16-20) and \code{(X0 == 1)} (lines 21-22).
Inside the \code{(X0 == 0)} gate, there are two nested gates (corresponding to \code{(X1 == 0)} and \code{(X1 == 1)}).

\paragraph{Path conditions. } Each gate $A$ has a \emph{path condition} $\psi_A$, which is a list of variables and values they need to take on in order for the gate body to be executed. For example, in \Figref{fig:gatesExample}, the path condition for the innermost gate body on lines 19-20 is $(X_0=0, X_1=1)$, where commas denote conjunction. We will use the convention that the condition in the deepest gate's \code{if} statement is the last entry of the path condition. Gates belong to a tree structure, and if gate $B$ with gate condition $\phi_B$ is nested inside gate $A$ with path condition $\psi_A$, then we say that $A$ is a parent of $B$, and the path condition for $B$ is $\psi_B=(\psi_A, \phi_B)$. We can equally speak of the path condition $\psi_j$ of a factor $j$, which is the path condition of the most deeply nested gate that the factor is contained in.

\paragraph{Active variables. } Define a variable $X$ to be \emph{active} in a gate $A$ if both of the following hold:
\begin{itemize}
\item $X$ is used in $A$ or one of its descendants, and
\item $X$ is declared in $A$ or one of its ancestors.
\end{itemize}
That is, $X$ is active in $A$ iff $A$ is on the path between $X$'s declaration and one of its uses.

For each gate $A$ in which a variable is active, we instantiate a separate local marginal annotated with the path condition of $A$ ($\psi_A$).
For example, inside the gate corresponding to \code{(X0 == 0)} in \Figref{fig:gatesExample}, the local marginal for $X_i$ is $\mu_i^{X_0=0}(x)$.\footnote{Strictly speaking, this notation
  does not handle the case where there are multiple gates with identical path
  conditions; for clearness of notation, assume that all gate path conditions
  are unique. However, the implementation handles repeated path conditions
  (by identifying local marginals according to a unique gate id).}
In the global scope we drop the superscript annotation and just use $\mu_i(x)$. We can refer to parent-child relationships between different local marginals of the same variable; the parent (child) of a local marginal $\mu_i^{\psi_A}(\cdot)$ is the local marginal for $X_i$ in the parent (child) gate of $A$.

 \paragraph{Gate marginals. } Let the \emph{gate marginal} of a gate $A$ be the marginal of the gate's condition in the parent gate of $A$. In \Figref{fig:gatesExample}, the first outer gate's gate marginal is $\mu_{0}(0)$, and the second outer gate's is $\mu_{0}(1)$.
 In the inner gate, the gate marginal for the (\code{X1 == 0}) gate is
 $\mu_{1}^{X_0=0}(0)$.


  \subsection{Forward Marginals Gradient Descent (FMGD) Back-end}
\label{sec:fmgd}

The factor graphs discussed above are easily converted into computation graphs representing the execution of an interpreter by the following operations.

\begin{itemize}
\item Annotate the factor graph edges with the direction of traversal during forwards execution of the \langname program.
\item Associate executable functions $f_i$ with factor $i$ operating on scope $S_i = \{\boldsymbol{X},Y\}$. The function transforms the incoming variables $\boldsymbol{X}$ to the outgoing variable, $Y=f_i(\boldsymbol{X})$.
\end{itemize}

In the FMGD approach, we initialize the source nodes of this directed graph by instantiating independent random variables $X_p \sim \mu_p$ at each \code{Param} node, and variables $X_i \sim {\rm onehot}(x_i^*)$ at nodes associated with input observations of the form \code{Xi.set\_to\_constant($x_i^*$)}. Here ${\rm onehot}(x_i^*)$ is a distribution over $\mcX_i$ with unit mass at $x_i^*$. We then propagate these \textit{distributions} through the computation graph using the FMGD approximation, described below, to obtain distributions $\mu_o$ at the output nodes associated with an \code{observe\_value($x_o^*$)} statement. This fuzzy system of distributions is fully differentiable. Therefore inference becomes an optimization task to maximize the weight $\mu_o(x^*_o)$ assigned to the observations by updating the parameter distributions $\{\mu_p\}$ by gradient descent.

The key FMGD approximation arises whenever a derived variable, $Y$ depends on several immediate input variables, $\boldsymbol{X}$. In an ungated graph, this occurs at factor nodes where $Y=f_i(\boldsymbol{X})$. FMGD operates under the under the approximation that \textit{all $\boldsymbol{X}$ are independent}. In this case, we imagine a local joint distribution $\mu_{Y\boldsymbol{X}}$ constructed according to the definition of $f_i$ and the independent unary marginal distributions for $\boldsymbol{X}$. From this distribution we marginalize out all of the input variables to obtain the unary marginal $\mu_Y$ (see \Secref{sec:fm}). Only $\mu_Y$ is propagated forward out of the factor node and correlations between $Y$ and $\boldsymbol{X}$ (only captured by the full local joint distribution) are lost. In the next section we explicitly define these operations and extend the technique to allow for gates in the factor graph.

It is worth noting that there is a spectrum of approximations in which we form joint distributions for subgraphs of size ranging from single nodes (FMGD) to the full computation graph (enumerative search) with only independent marginal distributions propagated between subgraphs. Moving on this spectrum trades computational efficiency for accuracy as more correlations can be captured in larger subgraphs. An exploration of this spectrum could be a basis for future work.

\subsubsection{Forward Marginals...}
\label{sec:fm}
\begin{figure}
\begin{center}
\begin{tabular}{|l|}
\cline{1-1}\\
\begin{tabular}{P{2in}P{1.5in}P{1.5in}}
\raisebox{-.5\height}{\includegraphics[height=5mm]{resources/var.pdf}}\hspace{3mm}\raisebox{-.5\height}{\includegraphics[height=5mm]{resources/param.pdf}} &
\raisebox{-.5\height}{\includegraphics[height=5.8mm]{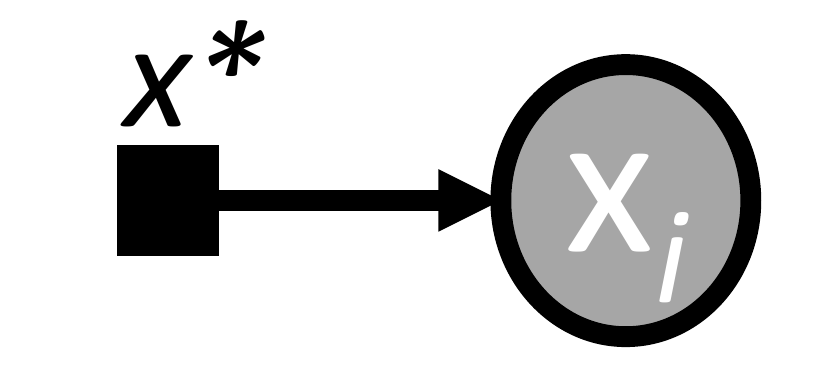}} &
\raisebox{-.5\height}{\includegraphics[height=6.7mm]{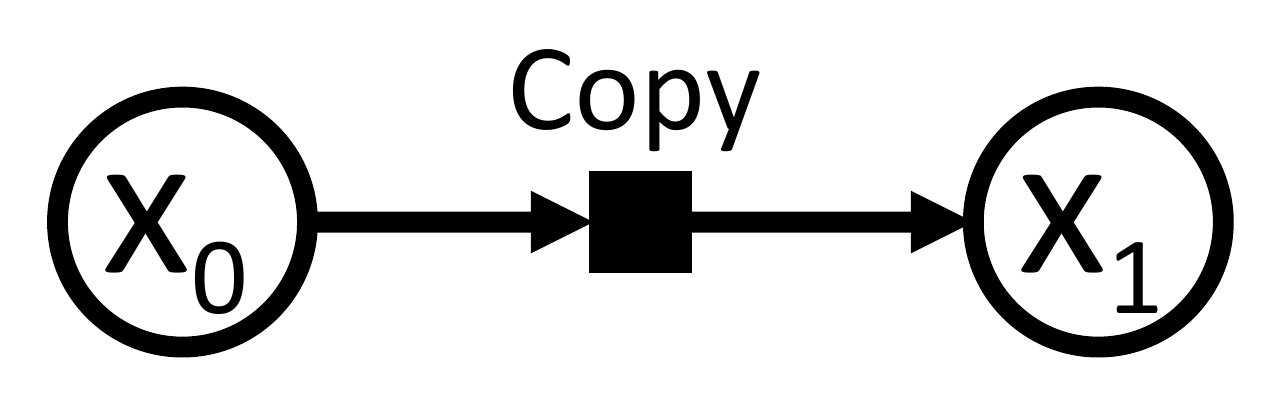}}\\
\begin{tabular}{c}
$\mu_i(x) \;;\;x\in\mcX_i$\\
\code{Param}:\; $\mu_i(x)={\rm softmax}\left[m_i(x)\right]$
\end{tabular}&
$\mu_i(x)=\mathbbm{1}\{x=x^*\}$ &
$\mu_1(x)=\mu_0(x)$\\ \\
\raisebox{-.5\height}{\includegraphics[height=19.2mm]{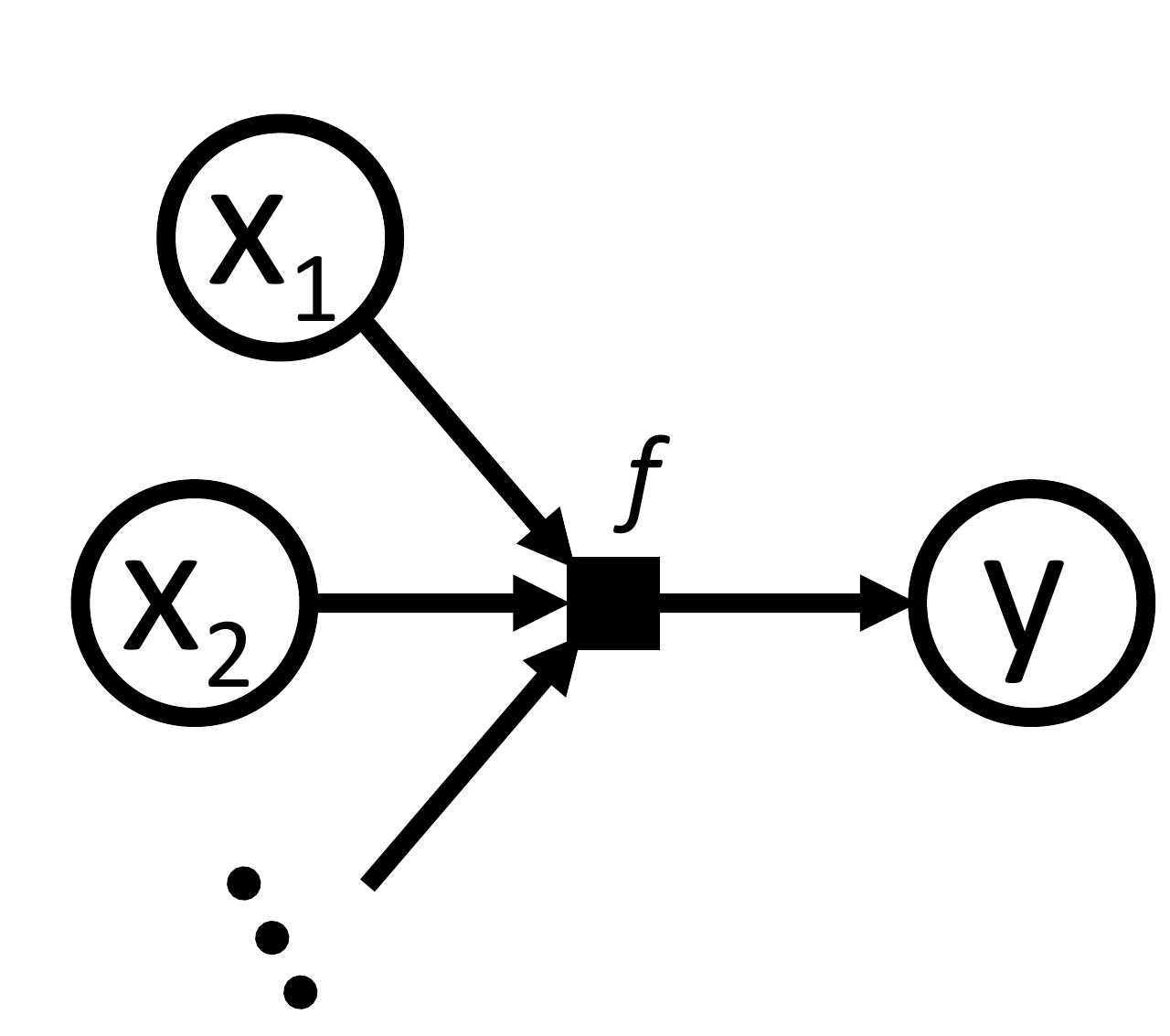}} &
\multicolumn{2}{P{3in}}{
\raisebox{-.5\height}{\includegraphics[height=30mm]{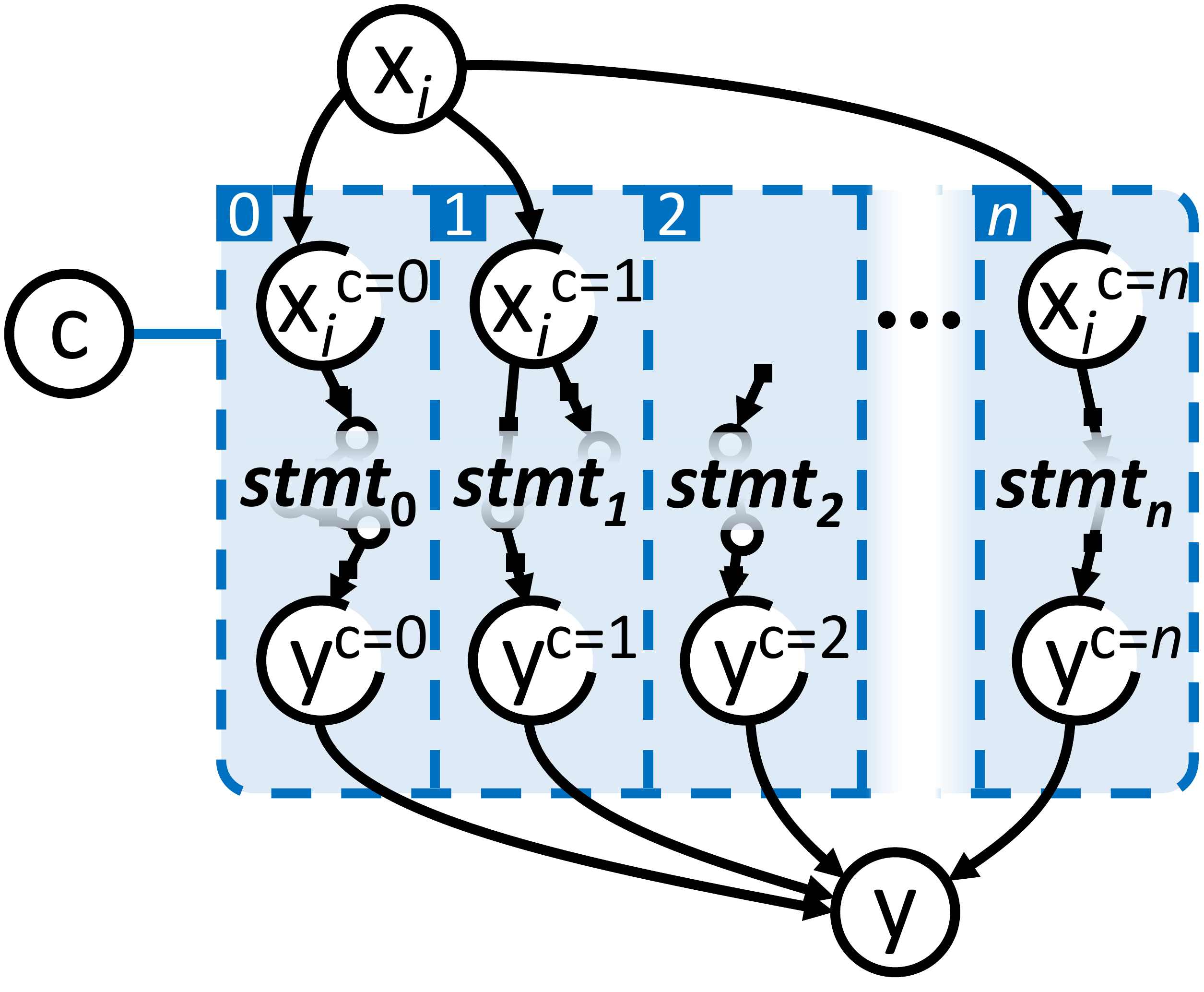}}
}\\
\begin{tabular}{l}
$\mu_Y(y)=\sum_{k}\mathbbm{1}\{y=f_i(\bx_k)\}w(\bx_k)$ \\
where $w(\mathbf{x}_k)=\prod_{i=1}^{M}\mu_i\left([\bx_k]_i\right)$
\end{tabular} &
\multicolumn{2}{P{3in}}{
\begin{tabular}{ll}
Enter gate & $\mu_i^{\psi,c=k}(x)=\mu_i^{\psi}(x)$ \\
Leave gate & $\mu_Y(y)=\sum_k\mu_Y^{c=k}(y)\mu_c$ \\
\end{tabular}}
\end{tabular}\\ \\
\cline{1-1}
\end{tabular}
\end{center}
\caption{\label{fig:fmgdFactBox} Summary of the forward execution of graphical primitives in the FMGD algorithm. See text for definition of symbols}
\end{figure}

\Figref{fig:fmgdFactBox} illustrates the transformation of each graphical primitive to allow a differentiable forward propagation of marginals through a factor graph. Below we describe more details of factor and gate primitives in this algorithm.

\paragraph{Factors. } The scope $S$ of a factor function $f$ contains the $M$ immediate input variables $X_{i}$ and the immediate output, $Y$. In this restricted environment, we enumerate the possible outputs $Y$ from all $\prod_{i=1}^{M}|\mcX_{i}|$ possible input configurations $\bx_k$ of the form $[\bx_k]_i \in \mcX_{i}$ for $i\in\{1,...,M\}$. We then marginalise over the configuration index, $k$, using weightings $\mu_i([\bx_k]_i)$ to produce $\mu_{Y}$ as follows:

\begin{equation}
\mu_{Y}(y)=\sum_{k}\mathbbm{1}\{y=f(\bx_k)\}w(\bx_k),
\label{eqn:fmgd}
\end{equation}

\noindent where $\mathbbm{1}$ is an indicator function and the weighting function $w$ is:

\begin{equation}
w(\bm{x}_k)=\prod_{i=1}^{M}\mu_i\left([\bx_k]_i\right).
\label{eqn:fmgdWeight}
\end{equation}

\noindent Note that \rEq{eqn:fmgd} and \rEq{eqn:fmgdWeight} can be implemented efficiently as a series of tensor contractions of the $M_i+1$ dimensional binary tensor $I_{y\bx}=\mathbbm{1}\{y=f(\bx)\}$ with the $M_i$ vectors $[\bm{\mu}_{i}]_x=\mu_i(x)$.



\paragraph{Gates. } We can include gates in the FMGD formulation as follows. Let $B_1,\ldots,B_k$ be the set of child gates of $A$ which are controlled by gate marginal $\mu_B^{\psi_A}(i)\;;\;i\in \{1,...,k\}$. Inside gate $B_i$, there is a subgraph $G_{i}$ described by \langname code $T_{i}$ which references a set of active variables $\mathcal{B}_i$. We divide $\mathcal{B}_i$ into $\mathcal{L}_i$ containing variables which are written-to during execution of $G_{i}$ (i.e. appear on the left hand side of expressions in $T_{i}$), and $\mathcal{R}_i$ containing variables which are not written-to (i.e. appear only on the right hand side of expressions in $T_{i}$). In addition, we use $\mathcal{A}$ to refer to the active variables in $A$, and $\mathcal{B^+}\subset\mathcal{A}$ to be variables used in the graph downstream of gates $B_1,\ldots,B_k$ on paths which terminate at observed variables.


On entering gate $B_i$, we import references to variables in the parent scope, $A$, for all $X \in \mathcal{R}_i\cap\mathcal{A}$:

\begin{equation}
\mu_{X}^{\psi_{B_i}}=\mu_{X}^{\psi_{A}}.
\label{eqn:enterContext}
\end{equation}

We then run $G_{i}$, to produce variables $\mathcal{L}_i$. Finally, when leaving a gate, we marginalise using the gate marginal to set variables $Y\in\mathcal{B^+}$:

\begin{equation}
\mu_{Y}^{\psi_{A}}(y)=\sum_{i=1}^k \mu_Y^{\psi_{B_i}}(y) \mu_B^{\psi_A}(i).
\label{eqn:fmgdGateAverage}
\end{equation}

\paragraph{Restrictions on factor functions. }
The description above is valid for any $f:\times_{i=1}^{M}\mcX_{i}\to \mathcal{Y}$, subject to the condition that $\mathcal{Y}\subseteq\mcX_{\rm out}$, where $\mcX_{\rm out}$ is the domain of the variable which is used to store the output of $f$. One scenario where this condition could be violated is illustrated below:
\vspace{5mm}
\begin{python}
@CompileMe([4],2)
def largeTest(x): return 1 if x >= 2 else 0
@CompileMe([4],4)
def makeSmall(x): return x - 2

X = Param(4) ; out = Var(4) ; isLarge = Var(2)

isLarge.set_to(largeTest(X))
if isLarge == 0:
    out.set_to(X)
elif isLarge == 1:
    out.set_to(makeSmall(X))
\end{python}
\vspace{5mm}

\noindent The function \code{makeSmall} has a range $\mathcal{Y}=\{-2,...,1\}$ which contains elements outside $\mcX_{\rm out}=\{0,...,3\}$. However, deterministic execution of this program does not encounter any error because the path condition \code{isLarge == 1} guarantees that the invalid cases $\mathcal{Y}\setminus\mcX_{\rm out}$ would never be reached. In general, it only makes sense to violate $\mathcal{Y}\subseteq\mcX_{\rm out}$ if we are inside a gate where the path condition ensures that the input values lie in a restricted domain $\tilde{\mcX}_{i}\subseteq\mcX_{i}$ such that $f:\times_{i=1}^{M}\tilde{\mcX}_{i}\to \mcX_{\rm out}$. In this case a we can simply enforce the normalisation of $\mu_{\rm out}$ to account for any leaked weight on values $\mathcal{Y}\setminus\mcX_{\rm out}$.

\begin{equation}
\mu_{Y}(y)=\frac{1}{Z}\sum_{k}\mathbbm{1}\{y=f(\mathbf{x}_k)\}w(\mathbf{x}_k),\qquad \mbox{where} \qquad Z=\sum_{k,y\in\mcX_{\rm out}}\mathbbm{1}\{y=f(\mathbf{x}_k)\}w(\mathbf{x}_k).
\label{eqn:fmgdAsTensorNorm}
\end{equation}

With this additional caveat, there are no further constraints on factor functions $f:\mathbb{Z}^M\to \mathbb{Z}$.

\subsubsection{... Gradient Descent}
\label{sec:gd}
Given a random initialization of marginals for the the \code{Param} variables $X_p\in\mathcal{P}$, we use the techniques above to propagate marginals forwards through the \langname model to reach all variables, $X_o\in\mathcal{O}$, associated with an \code{observe\_value}$(x^*_o)$ statement. Then we use a cross entropy loss, $L$, to compare the computed marginal to the observed value.

\begin{equation}
L=-\sum_{X_o\in\mathcal{O}}\log\left[\mu_{o}(x^*_o)\right].
\end{equation}

\noindent $L$ reaches its lower bound $L=0$ if each of the marginals $\mu_p(x)$ representing the \code{Params} put unit weight on a single value $\mu_p(x_p^*) = 1$ such that the assignments $\{X_p = x_p^*\}$ describe a valid program which explains the observations. The synthesis task is therefore an optimisation problem to minimise $L$, which we try to solve using backpropagation and gradient descent to reach a zero loss solution.

To preserve the normalisation of the marginals during this optimisation, rather than updating $\mu_p(x)$ directly, we update the log parameters $m_p(x)=[\boldsymbol{m}_p]_x$ defined by $\mu_{p}(x) = {\rm softmax}\left[m_p(x)\right]$. These are initialized according to 

\begin{equation}
\exp(\boldsymbol{m}_p) \sim \mbox{Dirichlet}(\boldsymbol{\alpha}),
\end{equation}

\noindent where $\boldsymbol{\alpha}$ are hyperparameters.

\subsubsection{Optimization Heuristics}
\label{sec:fmgd-tricks}
Using gradient information to search over program space is only guaranteed to succeed if all points with zero gradient correspond to valid programs which explain the observations. Since many different programs can be consistent with the observations, there can be many global optima ($L=0$) points in the FMGD loss landscape. However, the FMGD approximation can also lead to \textit{local} optima which, if encountered, stall the optimization at an uninterpretable point where $\mu_p(x_p^*)$ assigns weight to several distinct parameter settings. For this reason, we try several different random initializations of $m_p(x)$ and record the fraction of initializations which converge at a global optimum. Specifically, we try two approaches for learning using this model:

\begin{itemize}
\item \textbf{Vanilla FMGD. } Run the algorithm as presented above, with $\alpha_i = 1$ and using the \tool{RMSProp} \citep{Tieleman2012} gradient descent optimization algorithm.
\item \textbf{Optimized FMGD. } Add the heuristics below, which are inspired by \cite{Kurach15} and designed to avoid getting stuck in local minima, and optimize the hyperparameters for these heuristics by random search. We also include the initialization scale $\alpha_i = \alpha$ and the gradient descent optimization algorithm in the random search (see \Secref{sec:parity-chain-experiments} for more details). By setting $\alpha=1$, parameters are initialized uniformly on the simplex. By setting $\alpha$ smaller, we get peakier distributions, and by setting $\alpha$ larger we get more uniform distributions.
\end{itemize}

\paragraph{Gradient clipping. }  The FMGD neural network depth grows linearly with the number of time steps.  We mitigate the ``exploding gradient” problem~\citep{bengio1994learning} by 
globally rescaling the whole gradient vector so that its $L2$ norm is not bigger than some hyperparameter value $C$.

\paragraph{Noise. } We added random Gaussian noise to the computed gradients after the backpropagation step. Following~\cite{neelakantan2015adding}, we decay the variance of this noise during the training according to the following schedule:
\begin{equation}
\sigma^2_t=\frac{\eta}{(1+t)^{\gamma}}
\end{equation}
where the values of $\eta$ and $\gamma$ are hyperparameters and $t$ is the epoch counter.

\paragraph{Entropy. } Ideally, the algorithm would explore the loss surface to find a global minimum rather than fixing on some particular configuration early in the training process, causing the network to get stuck in a local minimum from which it's unlikely to leave.  To bias the network away from committing to any particular solution during early iterations, we add an {\em entropy bonus} to the loss function.  Specifically, for each softmax distribution in the network, we subtract
the entropy scaled by a coefficient $\rho$, which is a hyperparameter. The coefficient is exponentially decayed with rate
$r$, which is another hyperparameter.

\paragraph{Limiting the values of logarithms. } FMGD uses logarithms in computing both the cost function as well as the entropy.  Since the inputs to these logarithms can be very small, this can lead to very big values for the cost function and floating-point arithmetic overflows.  We avoid this problem by replacing $\log(x)$ with $\log(\max[x,\epsilon])$ wherever a logarithm is computed, for some small value of $\epsilon$.

%
%

\vspace{\baselineskip}
\noindent \cite{Kurach15} considered two additional tricks which we did not implement
generally. 

\paragraph{Enforcing Distribution Constraints. }  Because of the depth of the networks, propagation of numerical errors can result in $\sum_x\mu_i(x)\neq1$.  \cite{Kurach15} solve this by adding rescaling operations to ensure normalization. We find that we can avoid this problem by using 64-bit floating-point precision.

\paragraph{Curriculum learning. }  \cite{Kurach15} used a curriculum learning scheme which involved first training on small instances of a given problem, and only moving to train on larger instances once the error rate had reduced below a certain value.  Our benchmarks contain a small number of short examples (e.g., 5-10 examples acting on memory arrays of up to 8 elements), so there is less room for curriculum learning to be helpful. We manually experimented with hand-crafted
curricula for two hard problems (shift and adder), but it did not lead to improvements.

\vspace{\baselineskip}
To explore the hyperparameters for these optimization heuristics we ran preliminary
experiments to manually chose a distribution over hyperparameter
space for use in random search over hyperparameters. The aim was to
find a distribution that is broad enough to not disallow reasonable
settings of hyperparameters while also being narrow enough so that
runs of random search were not wasted on parameter settings that would
never lead to convergence. This distribution over hyperparameters was then
fixed for all random search experiments.

  \subsection{(Integer) Linear Program Back-end}
\label{sec:lp-back-end}
We now turn attention to the first alternative back-end to be compared with the FMGD. Casting the \langname program as a factor graph allows us to build upon standard practice in constructing LP relaxations for solving maximum a posteriori (MAP) inference problems in discrete graphical
models \citep{shlezinger1976syntactic,wainwright2008graphical}. In the following sections we describe how to apply these techniques to the \langname models, and in particular, how to extend the methods to handle gates.

\subsubsection{LP Relaxation}

The inference problem can be phrased as the task of finding the
highest scoring configuration of a set of discrete variables $X_0, \ldots, X_{D-1}$. The score is defined as the sum of local factor scores, $\theta_j$, where ${\theta_j: \bmcX_j\to \reals}$, and $\bmcX_j = \times_{i \in S_j} \mcX_i$ is the joint configuration space of the variables $\bx$ with indices $S_j = ( i_0, \ldots, i_{M_j} )$ spanning the scope of factor $j$. In the simplest case (when we are searching for \textit{any} valid solution) the factor score at a node representing a function $f_j$ will simply measure the consistency of the inputs ($\bx_{\setminus0}$) and output ($x_0$) at that factor:
\begin{equation}
\theta_j(\bx) = \mathbbm{1}\{x_0=f_j(\bx_{\setminus0})\}.
\end{equation}

Alongside these scoring functions, we can build a set of linear constraints and an overall linear objective function which represent the graphical model as an LP. The variables of this LP are the local unary marginals $\mu_{i}(x) \in \reals$ as before, and new \emph{local factor marginals} $\mu_{S_j}(\bx) \in \reals$ for $\boldsymbol{x} \in \bmcX_j$ associated with each factor, $j$. 

In the absence of gates, we can write the LP as:

\begin{align}
\max_{\boldsymbol{\mu}}\quad&\sum_{j} \sum_{\bx \in \boldsymbol{\mcX_j}} \mu_{S_j}(\bx) \theta_j(\bx)\nonumber\\
{\rm s.t.}\quad & \mu_i(x)\geq 0\;;\;\mu_{S_j}(\bx)\geq 0\nonumber\\
& \sum_{x \in \mcX_i} \mu_i(x) = 1\nonumber\\
& \sum_{\substack{\bx \in \bmcX_j \\ X_i = x}} \mu_{S_j}(\bx) = \mu_i(x),
\end{align}

\noindent where the final set of constraints say that when $X_i$ is fixed to value $x$ and all other variables are marginalized out from the local factor marginal, the result is equal to the value that the local marginal for $X_i$ assigns to value $x$. This ensures that factors and their neighboring variables have consistent local marginals.


If all local marginals $\mu(\cdot)$ are \emph{integral}, i.e., restricted to be 0 or 1, then the LP above becomes an integer linear program corresponding exactly to the
original discrete optimization problem. When the local marginals are real-valued
(as above), the resulting LP is not guaranteed to have equivalent solution to
the original problem, and \textit{fractional} solutions can appear. More formally, the LP constraints define what is known as the \emph{local polytope} $\local$.
which is an outer
approximation to the convex hull of all valid integral configurations
of the local marginals (known as the marginal polytope $\marginalpolytope$). 
In the case of program synthesis, fractional solutions are problematic, because
they do not correspond to discrete programs and thus cannot be represented as source code or executed on new instances.
When a fractional solution is found, heuristics such
as rounding, cuts, or branch \& bound search must be used in order to find an integral
solution.

 \subsubsection{Linear Constraints in Gated Models}

We now extend the LP relaxation above to cater for models with gates. In each gate we instantiate local unary marginals $\mu_i^\psi$ for each active variable and local factor marginals $\mu_{S_j}^\psi$ for each factor, where $\psi$ is the path condition of the parent gate.

The constraints in the LP are then updated to handle these gate specific marginals as follows:

\begin{figure}[t!]
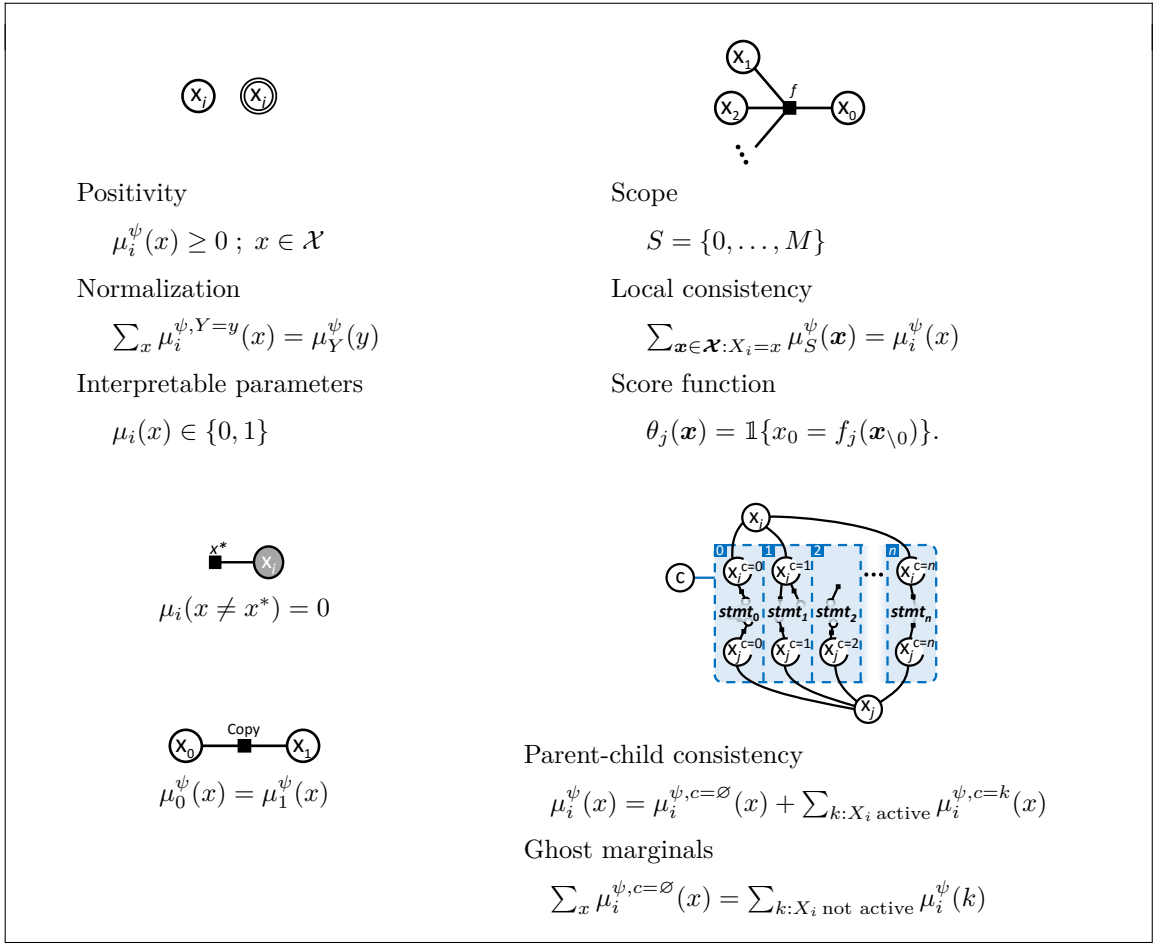

\begin{tabular}{|c|}
\cline{1-1}\\[-10pt]
\begin{tabular}{P{2in}P{3.5in}}
\raisebox{-.5\height}{\includegraphics[height=5mm]{resources/var.pdf}}\hspace{3mm}\raisebox{-.5\height}{\includegraphics[height=5mm]{resources/param.pdf}} &
\raisebox{-.5\height}{\includegraphics[height=19.2mm]{resources/factor.pdf}} \\
\begin{tabular}{l}
Positivity  \\ \hphantom{$\quad$} $\mu_i^\psi(x)\geq 0\;;\; x\in\mcX$\\
Normalization  \\ \hphantom{$\quad$} $\sum_{x}\mu_i^{\psi,Y=y}(x) = \mu_Y^\psi(y)$ \\
Interpretable parameters  \\ \hphantom{$\quad$} $\mu_i(x)\in\{0,1\}$
\end{tabular}
&
\begin{tabular}{l}
Scope \\ \hphantom{$\quad$} $S=\{0,\ldots,M\}$\\
Local consistency \\ \hphantom{$\quad$} $\sum_{\bx \in \bmcX : X_i = x} \mu_{S}^\psi(\bx) = \mu_i^\psi(x)$\\
Score function \\ \hphantom{$\quad$} $\theta_j(\bx) = \mathbbm{1}\{x_0=f_j(\bx_{\setminus0})\}.$
\end{tabular}\\[-15pt]
\begin{tabular}{P{2in}} \\[25pt]
\raisebox{-.5\height}{\includegraphics[height=5.8mm]{resources/observed.pdf}}\\
$\mu_i(x\neq x^*)=0$\\[30pt]
\raisebox{-.5\height}{\includegraphics[height=6.7mm]{resources/copy.pdf}}  \\
$\mu_{0}^\psi(x)=\mu_{1}^\psi(x)$ 
\end{tabular}
&
\begin{tabular}{P{3.5in}}
\raisebox{-.5\height}{\includegraphics[height=30mm]{resources/entercontext.pdf}}
\end{tabular}
\begin{tabular}{l}
Parent-child consistency\\ \hphantom{$\quad$}$ \mu^{\psi}_i(x) = \mu^{\psi,c=\emptyset}_i(x)+\sum_{k : X_i\,\hbox{\scriptsize active}} \mu^{\psi,c=k}_i(x)$\\ 
Ghost marginals \\ \hphantom{$\quad$}$\sum_x \mu^{\psi,c=\emptyset}_i(x) = \sum_{k : X_i\,\hbox{\scriptsize not active}} \mu^{\psi}_i(k)$\\
\end{tabular}
\end{tabular}\\[170pt]
\cline{1-1}
\end{tabular}
\caption{\label{fig:lp} Summary of the construction of a mixed integer linear program from a gated factor graph. Removing the binary constraint $\mu_i(x)\in\{0,1\}$ on the parameters produces a continuous LP relaxation. See main text for definition of symbols.}
\end{figure}

\paragraph{Normalization constraints. }
The main difference in the Gate LP from the standard LP is how normalization constraints are handled.
The key idea is that \emph{each local marginal in gate $A$ is normalized to sum to $A$'s gate marginal}. Thus the local marginal for $X_i$ in the gate with path condition $(\psi, Y=y)$ with gate marginal $\mu_Y$ is:
\begin{align}
\sum_{x\in \mcX_i}\mu_i^{\psi,Y=y}(x) = \mu_Y^\psi(y).
\end{align}

\noindent For local marginals in the global scope (not in any gate), the marginals are constrained to sum to 1, as in the standard LP.

\paragraph{Factor local marginals. } The constraint enforcing local consistency between the factor local marginals and the unary local marginals is augmented with path condition superscripts:
\begin{align}
  \sum_{\bx \in \bmcX_j : X_i = x} \mu^{\psi_A}_{S_j}(\bx) = \mu^{\psi_A}_i(x).
\end{align}

\paragraph{Parent-child consistency. }
There needs to be a relationship between different local marginals for the same variable.
We do this by enforcing consistency between parent-child local marginals.
Let $A$ be a parent gate of $B$, and let $X_i$ be active in both $A$ and $B$.
Then we need to enforce consistency between $\mu^{\psi_A}_i(x)$ and $\mu^{\psi_B}_i(x)$.
It is not quite as simple as setting these quantities equal; in general there are multiple
children gates of $A$, and $X$ may be active in many of them.
Let $B_1, \ldots, B_K$ be the set of children gates of $A$, and suppose that $X$ is active
in all of the children. Then the constraint is
\begin{align}
\sum_{k=1}^K \mu^{\psi_{B_k}}_i(x) & = \mu^{\psi_{A}}_i(x) \quad \forall x \in \mcX_i.
\label{eq:parent-child-consistency}
\end{align}
This can be thought of as setting a parent local marginal to be a weighted average of children local marginals, where the ``weights'' come from children marginals being capped at their corresponding gate marginal's value.

\paragraph{Ghost marginals. }
A problem arises if a variable is used in some but not all children gates. It may be tempting in this case to replace the above constraint with one that leaves out the children where the variable is inactive:
\begin{align}
\sum_{k : X_i \hbox{ is active}} \mu^{\psi_{B_k}}_i(x) & = \mu^{\psi_{A}}_i(x).
\end{align}
This turns out to lead to a contradiction. To see this, consider $X_3$ in \Figref{fig:gatesExample}. $X_3$ is inactive in the {\code{(X0 == 1)}} gate, and thus the parent-child consistency constraints would be
\begin{align}
\mu^{X_0=0}_3(x) & = \mu_3(x) \quad \forall x.
\end{align}
However, the normalization constraints for these local marginals are
\begin{align}
  \sum_x \mu^{X_0=0}_3(x) & = \mu_0(0) \\
  \sum_x \mu_3(x) & = 1.
\end{align}
This implies that $\mu_0(0) = 1$, which means we must assign zero probability to the case when $X_3$ is not active. This removes the possibility of $X_0=1$ from consideration which is clearly undesirable, and if there are disjoint sets of variables active in the different children cases, then the result is an infeasible LP.

The solution is to instantiate \emph{ghost marginals}, which are local marginals for a variable in the case where it is undefined (hence the term ``ghost''). We denote a ghost marginal with a path condition entry where the value is set to $\emptyset$, as in  $\mu^{X_0=\emptyset}_3(x)$.
Ghost marginals represent the distribution over values in all cases where a variable is not defined, so the normalization constraints are defined as follows:
\begin{align}
  \sum_x \mu^{X_0=\emptyset}_i(x) & = \sum_{k : X_i \hbox{ \scriptsize is not active}} \mu^{}_0(k).
\end{align}

Finally, we can fix the parent-child consistency constraints in the case where a variable is active in some children.
The solution is to consider the ghost marginal as one of the child cases.
In the example of $X_3$, the constraint would be the following:
\begin{align}
\mu^{}_3(x) & =
\mu^{X_0=\emptyset}_3(x) +
\sum_{k : X_3 \hbox{ \scriptsize is active} } \mu^{X_0=k}_3(x) \hbox{ for all $x \in \mcX_3$}.
\end{align}

\vspace{\baselineskip}
\noindent The full set of constraints for solving \langname IPS problems using gated (integer) LPs is summarized in \Figref{fig:lp}.

  \subsection{SMT Back-end}
\label{sec:SMT}

At its core, an IPS problem in \langname induces a simple linear integer
constraint system.
To exploit mature constraint-solving systems such as \tool{Z3}~\citep{Moura08}, we
have implemented a satisfiability modulo theories (SMT) back-end.
For this, a \langname instance is translated into a set of constraints in the
\tool{SMT-LIB} standard~\citep{smt-lib}, after which any standard SMT solver can be
called.

\newcommand{\smtTranslate}[1]{\ensuremath{\llbracket #1 \rrbracket}_{\mathsf{SMT}}}
\newcommand{\smtExprTranslate}[1]{\ensuremath{\llbracket #1 \rrbracket}_{\mathsf{SMT}}^{\mathsf{E}}}

\begin{figure}
  \[
  \begin{array}{lcl}
    \smtExprTranslate{n} & = & n \\
    \smtExprTranslate{v_c} & = & v_c \\
    \smtExprTranslate{f(c_1,\cdots,c_k)} & = &
       \smtExprTranslate{s[v_0/c_0, \cdots, v_n/c_k]}
       \quad\text{for } \code{def}\; f\code{(}v_0\code{,}\cdots\code{,}v_k\code{):} s\\
    \smtExprTranslate{c_0\;\mathit{op}_a\;c_1} & = & \code{(}\mathit{op}_a\; \smtExprTranslate{c_0}\; \smtExprTranslate{c_1}\code{)}\\
    \smtExprTranslate{v} & = & v \\
    \smtExprTranslate{a_0\;\mathit{op}_a\;a_1} & = & \code{(}\mathit{op}_a\; \smtExprTranslate{a_0}\; \smtExprTranslate{a_1}\code{)}\\ 
    \smtExprTranslate{f(a_0,\cdots,a_k)} & = &
       \smtExprTranslate{s[v_0/a_0, \cdots, v_k/a_k]}
       \quad\text{for } \code{def}\; f\code{(}v_0\code{,}\cdots\code{,}v_k\code{):} s\\
    \smtExprTranslate{\code{not} \; b} & = & \code{(not}\;\smtExprTranslate{b}\code{)}\\
    \smtExprTranslate{b_0\;\code{and}\;b_1} & = & \code{(and}\;\smtExprTranslate{b_0}\;\smtExprTranslate{b_1}\code{)}\\
    \smtExprTranslate{b_0\;\code{or}\;b_1} & = & \code{(or}\;\smtExprTranslate{b_0}\;\smtExprTranslate{b_1}\code{)}\\
    \smtExprTranslate{a_0\;{op}_c \;a_1} & = & \code{(}\mathit{op}_c\;\smtExprTranslate{a_0}\;\smtExprTranslate{a_1}\code{)}\\
    \smtExprTranslate{v \; \code{=} \; a \code{;} s} & = & \smtExprTranslate{s[v/a]}\\
    \smtExprTranslate{\code{if}\; b_0\code{:} \;s_0\; \code{else:} \; s_1} & = &
      \code{(ite } \; \smtExprTranslate{b_0} \; \smtExprTranslate{s_0} \; \smtExprTranslate{s_1} \code{)}\\
    \smtExprTranslate{\code{return} \; a} & = & \smtExprTranslate{a}\\
  \end{array}
  \]
  \caption{A syntax-directed translation $\smtExprTranslate{\cdot}$ of \langname
    expressions to \tool{SMT-LIB 2}.
    Here, $s[\mathit{var}/\mathit{expr}]$ replaces all occurrences of
    $\mathit{var}$ by $\mathit{expr}$.}
  \label{fig:smtExprTranslation}
\end{figure}

To this end, we have defined a syntax-guided transformation function
$\smtExprTranslate{\cdot}$ that translates \langname expressions into \tool{SMT-LIB}
expressions over integer variables, shown in \Figref{fig:smtExprTranslation}.
We make use of the unrolling techniques discussed earlier to eliminate arrays,
\code{for} loops and \code{with} statements.
When encountering a function call as part of an expression, we use
\emph{inlining}, i.e., replace the call by the function definition in which
formal parameters have been replaced by actual arguments.
This means that some \langname statements have to be expressed as \tool{SMT-LIB}
expressions, and also means that the SMT back-end only supports a small subset
of functions, namely those that are using only \langname (but not arbitrary
Python) constructs.

\begin{figure}
  \[
  \begin{array}{lcl}
    \smtTranslate{s_{0}\;\code{;}\;s_{1}} & = & 
      \smtTranslate{s_0} \; @ \; \smtTranslate{s_1} \\
    \smtTranslate{a_0\code{.set\_to(}a_1\code{)}} & = & 
      [\code{(=} \; \smtExprTranslate{a_0} \; \smtExprTranslate{a_1} \code{)}]\\
    \smtTranslate{a\code{.set\_to\_constant(}c\code{)}} & = & 
      [\code{(=} \; \smtExprTranslate{a} \; \smtExprTranslate{c} \code{)}]\\
    \smtTranslate{a_0\code{.observe\_value(}a_1\code{)}} & = & 
      [\code{(=} \; \smtExprTranslate{a_0} \; \smtExprTranslate{a_1} \code{)}]\\
    \smtTranslate{\code{if}\; b_0\code{:} \;s_0\; \code{else:} \; s_1} & = &
      [\code{(=>}\;                \smtExprTranslate{b_0}         \; \code{(and} \; \smtTranslate{s_0}\code{))},\\
      &&\phantom{[}
       \code{(=>}\; \code{(not} \; \smtExprTranslate{b_0}\code{)} \; \code{(and} \; \smtTranslate{s_1}\code{))}]\\
    \smtTranslate{s_{d_0}\;\code{;}\;s_{d_1}} & = &
       \smtTranslate{s_{d_0}} \; @ \; \smtTranslate{s_{d_1}}\\
    \smtTranslate{v \;\code{=}\; c} & = & 
       [\code{(=} \; v \; \smtExprTranslate{c}\code{)}]\\
    \smtTranslate{v \;\code{=}\; \code{Var(}c\code{)}} & = & 
       [\code{(>=} \; v \; \code{0)}, \code{(<} \; v \; \smtExprTranslate{c}\code{)}]\\
    \smtTranslate{v \;\code{=}\; \code{Param(}c\code{)}} & = & 
       [\code{(>=} \; v \; \code{0)}, \code{(<} \; v \; \smtExprTranslate{c}\code{)}]\\
  \end{array}
  \]
  \caption{A syntax-directed translation $\smtTranslate{\cdot}$ of \langname
    statements to \tool{SMT-LIB 2}.
    Here, $[\cdots]$ are lists of constraints, and $@$ concatenates such lists.}
  \label{fig:smtTranslation}
\end{figure}

Building on $\smtExprTranslate{\cdot}$, we then define the statement translation
function $\smtTranslate{\cdot}$ shown in \Figref{fig:smtTranslation}.
Every statement is translated into a list of constraints, and a solution to the
IPS problem encoded by a \langname program $p$ is a solution to the conjunction
of all constraints generated by $\smtExprTranslate{p}$.
Together with the unrolling of loops for a fixed length, this approach is
reminiscent of bounded model checking techniques (e.g. \citep{Clarke01}) which
for a given program, search for an input that shows some behavior.
Instead, we take the input as given, and search for a program with the desired
behavior.


  \subsection{Sketch Back-end}
\label{sec:sketch}

\renewcommand{\t}[1]{\texttt{#1}}
\newcommand{\stranslate}[1]{\ensuremath{\llbracket #1 \rrbracket}_{\mathsf{Sk}}}

\lstnewenvironment{sketchp}[1][]
{
\sketchstyle
\lstset{#1}
}
{}

\newcommand\sketchstyle{\lstset{
language=C,
  morekeywords=[1]{harness, assert, implements, generator},
  morekeywords=[2]{Int, Bool, and, or, not},
  alsoletter=?!-,
  alsodigit=\$\%&*+./:<=>@^_~,
  sensitive=false,
  morecomment=[s]{/*}{*/},
  morestring=[b]",
  basicstyle=\small\ttfamily,
  keywordstyle=[1]\bf\ttfamily\color[rgb]{0,.3,.7},
  keywordstyle=[2]\bf\ttfamily\color[rgb]{0.57,0.12,0},
  commentstyle=\color[rgb]{0.133,0.545,0.133},
  stringstyle={\color[rgb]{0.75,0.49,0.07}},
  upquote=true,
  breaklines=true,
  breakatwhitespace=true,
  literate=*{`}{{`}}{1},
frame=tb,                         
showstringspaces=false            %
}}

The final back-end which we consider is based on the $\sketch$~\citep{sketch} program synthesis system, which allows programmers to write partial programs called \emph{sketches} while leaving fragments unspecified as \emph{holes}. The goal of the synthesizer is to automatically fill in these holes such that the completed program conforms to a desired specification. The $\sketch$ system supports multiple forms of specifications such as input-output examples, assertions, reference implementation, etc.

\paragraph{Background. } The syntax for the $\sketch$ language is similar to the C language with only one additional feature -- a symbol \t{??} that represents an unknown constant integer value.
A simple example sketch is shown in \Figref{sketchexample}, which represents a partial program with an unknown integer \t{h} and a simple assertion.
The \t{harness} keyword indicates to the synthesizer that it should compute a value for \t{h} such that in the complete function, all assertions are satisfied for all input values \t{x}.
For this example, the $\sketch$ synthesizer computes the value $\t{h}=3$ as expected.

\begin{figure}
\begin{sketchp}
harness void tripleSketch(int x){
  int h = ??;  // hole for unknown constant
  assert h * x == x + x + x;
}
\end{sketchp}
\caption{A simple sketch example.}
\label{sketchexample}
\end{figure}

The unknown integer values can be used to encode a richer hypothesis space of program fragments. For example, the sketch in \Figref{sketchricherex} uses an integer hole to describe a space of binary arithmetic operations. The $\sketch$ language also provides a succinct language construct to specify such expression choices : \t{ lhs \{| + | - | * | / | \% |\} rhs}.

\begin{figure}[!htpb]
\begin{sketchp}
int chooseArithBinOp(int lhs, int rhs){
  int c = ??; // unknown constant integer value
  assert c < 5;
  if(c == 0) return lhs + rhs;
  if(c == 1) return lhs - rhs;
  if(c == 2) return lhs * rhs;
  if(c == 3) return lhs / rhs;
  if(c == 4) return lhs % rhs;
}
\end{sketchp}
\caption{Using unknown integer values to encode a richer set of unknown expressions.}
\label{sketchricherex}
\end{figure}

The $\sketch$ synthesizer uses a counter-example guided inductive synthesis algorithm (CEGIS)~\citep{asplos2006} to efficiently solve the second order exists-forall synthesis constraint. The key idea of the algorithm is to divide the process into phases : i) a synthesis phase that computes the value of unknowns over a finite set of input-output examples, and ii) a verification phase that checks if the current solution conforms to the desired specification. If the current completed program satisfies the specification, it returns the program as the desired solution. Otherwise, it computes a counter-example input that violates the specification and adds it to the set of input-output examples and continues the synthesis phase. More details about the CEGIS algorithm in $\sketch$ can be found in~\cite{sketch}.

\paragraph{Compiling \langname to {\sc \sketch}. } In \Figref{sketchtranslation}, we present a syntax-directed translation of the \langname language to $\sketch$. The key idea of the translation is to model \t{Param} variables as unknown integer constants (and integer arrays with constant values) such that the synthesizer computes the values of parameters to satisfy the observation constraints. For a $\t{Param}(N)$ integer value, the translation creates corresponding unknown integer value \t{??} with an additional constraint that the unknown value should be less than $N$. Similarly, for the $\t{Param}(N)$ array values, the translation creates a $\sketch$ array with unknown integer values, where each value is constrained to be less than $N$.  The \t{set\_to} statements are translated to assignment statements whereas the \t{observe} statements are translated to \t{assert} statements in $\sketch$. The user-defined functions are translated directly to corresponding functions in sketch, whereas the \t{with} statements are translated to corresponding assignment statements. The sketch translation of the \langname model in \Figref{fig:automaton1}(a) is shown in \Figref{sketchtranslationex}.

\begin{figure}
    \[
    \begin{array}{lcl}
    \stranslate{n} & = & n \\
    \stranslate{v_c} & = & v_c \\
    \stranslate{f(c_1,\cdots,c_k)} & = & f\code{(}\stranslate{c_1}\code{,}\cdots\code{,}\stranslate{c_k}\code{)} \\
    \stranslate{c_0\;\mathit{op}_a\;c_1} & = & \stranslate{c_0}\;\mathit{op}_a\;\stranslate{c_1} \\
    \stranslate{v} & = & v \\
    \stranslate{v[a_1,\cdots,a_k]} & = & v\code{[}a_1\code{][}\cdots\code{][}a_k\code{]} \\
    \stranslate{a_0\;\mathit{op}_a\;a_1} & = & \stranslate{a_0}\;\mathit{op}_a\;\stranslate{a_1}\\
    \stranslate{f(a_0,\cdots,a_k)} & = & f\code{(}\stranslate{a_0}\code{,}\cdots\code{,}\stranslate{a_k}\code{)} \\
    \stranslate{\code{not} \; b} & = & \code{!(}\stranslate{b}\code{)} \\
    \stranslate{b_0\;\code{and}\;b_1} & = & \stranslate{b_0}\;\code{\&\&}\;\stranslate{b_1}\\
    \stranslate{b_0\;\code{or}\;b_1} & = & \code{(} \stranslate{b_0}\;\code{||}\;\stranslate{b_1} \code{)}\\
    \stranslate{a_0\;{op}_c \;a_1} & = & \stranslate{a_0}\;{op}_c \;\stranslate{a_1}\\
    \stranslate{s_{0}\;\code{;}\;s_{1}} & = & \stranslate{s_0}\code{;}\; \stranslate{s_1} \\
    \stranslate{a_0\code{.set\_to(}a_1\code{)}} & = & \stranslate{a_0}\;\code{=}\;\stranslate{a_1}\code{;}  \\
    \stranslate{a\code{.set\_to\_constant(}c\code{)}} & = & \stranslate{a}\;\code{=}\;\stranslate{c}\code{;}  \\
    \stranslate{a_0\code{.observe\_value(}a_1\code{)}} & = & \code{assert}\;\stranslate{a_0}\; \code{==}\; \stranslate{a_1};  \\
    \stranslate{\code{return} \; a} & = & \code{return} \; \stranslate{a}\code{;} \\
    \stranslate{\code{if}\; b_0\code{:} \;s_0\; \code{else:} \; s_1} &=&
      \code{if (}    \stranslate{b_0}\code{) \{} \stranslate{s_0} \code{\}}\;
      \code{else \{}                             \stranslate{s_1} \code{\}}\\
    \stranslate{\code{for}\; v \; \code{in} \; \code{range}(c_1)\: \code{:} \;  s} & = &
      \code{for(int } v \;\code{= 0}\code{;}\; v \;\code{<}\; \stranslate{c_1}\code{;}\; v\code{++) \{} \stranslate{s} \code{\}} \\
    \stranslate{\code{for}\; v \; \code{in} \; \code{range}(c_1,c_2)\: \code{:} \; s} & = &
      \code{for(int } v \;\code{=}\; \stranslate{c_1}\code{;}\; v \;\code{<}\; \stranslate{c_2}\code{;}\; v\code{++) \{} \stranslate{s} \code{\}} \\
    \stranslate{\code{with} \; a \; \code{as} \; v \;\code{:}\; s} & = & \code{\{int}\; v \;\code{=}\; \stranslate{a} \code{;}\; \stranslate{s} \code{\}}\\
    \stranslate{s_{d_0}\;\code{;}\;s_{d_1}} & = & \stranslate{s_{d_0}}\code{;}\;\;\stranslate{s_{d_1}}\\
    \stranslate{v \;\code{=}\; c} & = & \code{int} \; v \;\code{=}\; \stranslate{c}\code{;}\\
    \stranslate{v \;\code{=}\; \code{Var(}c\code{)}} & = &
      \code{int} \; v\code{;}\\
    \stranslate{v \;\code{=}\; \code{Var(}c\code{)[}c_1\code{,}\cdots\code{,}c_k\code{]}} & = &
      \code{int[}\stranslate{c_1}\code{][}\cdots\code{][}\stranslate{c_k}\code{]} \; v\code{;}\\
    \stranslate{v \;\code{=}\; \code{Param(}c\code{)}} & = &
      \code{int} \; v \;\code{=}\; \code{??};\;\code{assert}\; v \;\code{<}\; \stranslate{c}\code{;}\\
    \stranslate{v \;\code{=}\; \code{Param(}c\code{)[}c_1\code{,}\cdots\code{,}c_k\code{]}} & = &
      \code{int[}\stranslate{c_1}\code{][}\cdots\code{][}\stranslate{c_k}\code{]} \; v\code{;}\\
    && \forall i_1 \in \stranslate{c_1}, \cdots, i_k \in \stranslate{c_k}: \\
    && \hspace*{5mm} v\code{[}i_1\code{][}\cdots\code{][}i_k\code{] = ??;}\; \code{assert} \; v\code{[}i_1\code{][}\cdots\code{][}i_k\code{]}\;\code{<}\; \stranslate{c}\code{;}\\
    \stranslate{\code{@CompileMe([}c_1\code{,}\cdots\code{,}c_k\code{],} c_r\code{)}} &=& \code{;}\\
    \stranslate{\code{def}\; f\code{(}v_0\code{,}\cdots\code{,}v_k\code{):} s} &=&
      \code{int}\; f\code{(}v_0\code{,}\;\cdots\code{,}v_k\code{) \{} \stranslate{s} \code{\}}
    \end{array}
    \]
    \caption{A syntax-directed translation $\stranslate{\cdot}$ of \langname programs to $\sketch$.}
    \label{sketchtranslation}
    \end{figure}

    \begin{figure}[!htpb]
\begin{sketchp}
int const_n = 5;
int[2][2] ruleTable = (int[2][2]) ??;
for(int i=0; i<2; i++){
  for(int j=0; j<2; j++){
    assert ruleTable[i][j] < 2;
  }
}
int[const_n] tape;

// assignment statements for input initialisations

for(int t=1; t<const_n-1; t++){
  int x1 = tape[t];
  int x0 = tape[t-1];
  tape[t+1] = ruleTable[x0,x1];
}

// assert statements for corresponding outputs

\end{sketchp}
\caption{The sketch translation for the \langname model shown in \Figref{fig:automaton1}(a) .}
\label{sketchtranslationex}
\end{figure}


%
\section{Analysis}
\label{sec:analysis}
One motivation of this work was to compare the performance of the gradient based FMGD technique for IPS with other back-ends. Below we present a task which all other back-ends solve easily, but FMGD is found to fail due to the prevalence of local optima.

\subsection{Failure of FMGD}

\cite{Kurach15} and \cite{neelakantan2015adding} mention that many random
restarts and a careful hyperparameter search are needed in order to converge to a correct
deterministic solution. Here we develop an understanding of the loss
surface that arises using FMGD in a simpler setting, which we believe
sheds some light on the local optima structure that arises when using
FMGD more generally.

Let $x_0, \ldots, x_{K-1}$ be binary variables with $x_0=0$ and all
others unobserved. For each $k=0, \ldots, K-1$, let $y_k = (x_k +
x_{(k+1) \mod K}) \mod 2$ be the parity of neighboring $x$ variables
connected in a ring shape. Suppose all $y_k$ are observed to be 0 and
the goal is to infer the values of each $x_k$. The \langname~program
is as follows, which we refer to as the \emph{Parity Chain} model:
\vspace{0.5cm}
\begin{python}
const_K = 5
x = Param(2)[const_K]
y = Var(2)[const_K]

@CompileMe([2,2], 2)
def Parity(a,b): return (a + b) % 2

x[0].set_to_constant(0)

for k in range(K):
   y[k].set_to(Parity(x[k], x[(k+1) % K]))
   y[k].observe_value(0)
\end{python}
\vspace{0.5cm}
Clearly, the optimal configuration is to set all $x_k=0$. Here we show analytically that there are exponentially many suboptimal local optima that FMGD can fall into, and experimentally that the probability of falling into a suboptimal local optimum grows quickly in $K$.

To show that there are exponentially many local optima, we give a
technique for enumerating them and show that the gradient is equal to
$\boldsymbol{0}$ at each. Letting $m_i(a)$ for $i \in \{0, \ldots K-1\}, a \in \{0, 1\}$
be the model parameters and 
$\mu_i = \frac{\exp m_i(1)}{\exp m_i(0) + \exp m_i(1)}$, the main
observation is that locally, a configuration of $[\mu_{i-1}, \mu_i,
  \mu_{i+1}] = [0, .5, 1]$ or $[\mu_{i-1}, \mu_i, \mu_{i+1}] = [1, .5,
  0]$ gives rise to zero gradient on $m_{i-1}(\cdot), m_i(\cdot), m_{i+1}(\cdot)$, as
does any configuration of $[\mu_{i-1}, \mu_i, \mu_{i+1}] = [0, 0, 0]$
or $[\mu_{i-1}, \mu_i, \mu_{i+1}] = [1, 1, 1]$.  This implies that any
configuration of a sequence of $\mu$'s of the form $[0, .5, 1, 1,
  \ldots, 1, .5, 0]$ also gives rise to zero gradients on all the
involved $m$'s.  We then can choose any configuration of alternating
$\mu$ (so e.g., $\mu_2, \mu_4, \mu_6, \ldots, \mu_K \in
\{0,1\}^{K/2}$), and then fill in the remaining values of $\mu_1,
\mu_3, \ldots$ so as to create a local optimum. The rule is to set
$\mu_i = .5$ if $\mu_{i-1} + \mu_{i+1} = 1$ and $\mu_{i} =
\mu_{i-1}=\mu_{i+1}$ otherwise.  This will create ``islands'' of 1's,
with the boundaries of the islands set to be .5. Each configuration of
islands is a local optimum, and there are at least $2^{(K-1)/2}$ such
configurations. A formal proof appears in Appendix
\ref{sec:fmgd-proof}.

One might wonder if these local optima arise in practice.  That is, if
we initialize $m_i(a)$ randomly, will we encounter these suboptimal
local optima? Experiments in \Secref{sec:parity-chain-experiments} show that the answer is
yes. The local optima can be avoided in small models by using
optimization heuristics such as gradient noise, but as the models grow
larger (length 128), we were not able to find any configuration of
hyperparameters that could solve the problem from a random
initialization. The other inference algorithms will solve these problems
easily. For example, the LP relaxation from \Secref{sec:lp-back-end} will lead
to integral solutions for tree-structured graphical models, which is
the case here.

\subsection{Parity Chain Experiments}
\label{sec:parity-chain-experiments}

Here we provide an empirical counterpart to the theoretical analysis in
the previous section. Specifically, we showed that there are
exponentially many local optima for FMGD to fall into in the Parity
Chain model, but this does not necessarily mean that these local
optima are encountered in practice. It is conceivable that there is a
large basin of attraction around the global optimum, and smaller,
negligible basins of attraction around the suboptimal local optima.

To answer this question, we run Vanilla FMGD (no optimization
heuristics) with random initialization parameters chosen so that
initial parameters are drawn uniformly from the simplex. Measuring the
fraction of runs (from 100 random initializations) that converge to
the global optimum then gives an estimate of the volume of parameter
space that falls within the basin of attraction for the global
optimum. Results for chain lengths of $K=4, 8, 16, 32, 64, 128$ appear
in the Vanilla FMGD row of \Tableref{tbl:parity-chain}. FMGD is able to
solve very small instances reliably, but performance quickly falls off
as $K$ grows. This shows that the basins of attraction for the
suboptimal local optima are large.

Next, we try the optimization
heuristics discussed in \Secref{sec:fmgd-tricks}.  For each chain length $K$, we draw
100 random hyperparameter settings from the manually chosen hyperparameter distribution.
At each hyperparameter setting, we run 10 runs with different random seeds and measure
the fraction of runs that converge to the global optimum. In the ``Best Hypers''
row of \rTab{tbl:parity-chain}, we report the percentage of successes
from the hyperparameter setting that yielded the best results. In the
``Average Hypers'' row, we report the percentage of success across all
1000 runs.

\begin{figure}
\begin{center}
\includegraphics[width=0.5\textwidth]{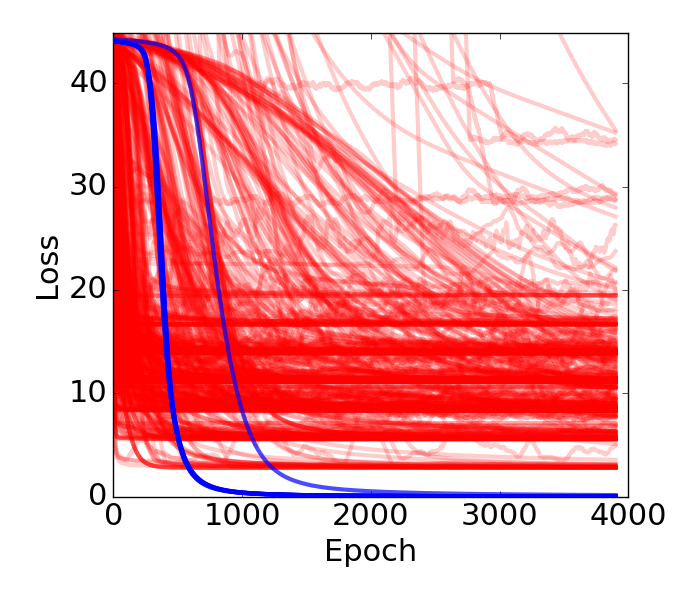}
\end{center}
  \caption{\label{fig:parity-chain64-traces} Loss versus epoch for all
    runs of the random search on Parity Chain ($K=64$).}
\end{figure}

Note that the 80\% success rate for Best Hypers on $K=64$ is an
anomaly, as it was able to find a setting of hyperparameters for which
the random initialization had very little effect, and the successful runs
followed nearly identical learning trajectories. See \Figref{fig:parity-chain64-traces} for a
plot of optimization objective versus epoch. Successful runs are colored
blue while unsuccessful ones are in red. The large cluster of successful runs
were all from the same hyperparameter settings.

\begin{table}
  \begin{center}
  \begin{tabular}{rcccccc}
    \toprule
     & $K=4$  & $K=8$  & $K=16$  & $K=32$  & $K=64$   & $K=128$ \\
    \midrule
    {\bf Vanilla FMGD}   &  100\% & 53\%   & 14\%   & 0\%   & 0\%   & 0\% \\
    {\bf Best Hypers}    &  100\% & 100\%  & 100\%  & 70\%  & 80\%  & 0\% \\
    {\bf Average Hypers}    &  84\% & 42\%  & 21\%  & 4\%  & 1\%  & 0\% \\
    \bottomrule
  \end{tabular}
  \end{center}
  \caption{\label{tbl:parity-chain}
    Percentage of runs that converge to the global optimum for FMGD on
    the Parity Chain example.
  }
\end{table}

\section{Experiments}
\label{sec:experiments}

We now turn attention to experimental results. Our primary aim is to better
understand the capabilities of the different back-ends on a range of problems,
and to establish some trends regarding the performance of the back-ends as
problem properties are varied.

\subsection{Benchmarks Results}

We now present the main results of this investigation: a benchmarking of all four inference techniques listed in \rTab{tbl:backends} on all twelve synthesis tasks listed in \rTab{tab:benchmarkTasks}. As described in \Secref{sec:models}, the tasks are split across four execution models of increasing practicality, with each model set three tasks of increasing difficulty. For any given task we ensure a fair test by presenting all four back-end compilers with the same \langname program (as listed in \Appref{app:models}) and the same set of input-output examples.
Since the different back-ends use very different approaches to solve the tasks, the only comparable metric to record is the wall time to reach a solution.

With the exception of the FMGD algorithm, we set a timeout of 4 hours for each back-end on each task (excluding any compilation time), give each algorithm a single run 
to find a solution, and do not tune the algorithms to each specific task. For the FMGD algorithm we run both the Vanilla and the Optimized form. In the Vanilla case we report the fraction of 20 different random initializations which lead to a globally optimal solution and also the wall clock time for 1000 epochs of the gradient descent algorithm (which is the typical number of iterations required to reach convergence on a successful run). In the Optimized FMGD case, we follow a similar protocol as in the
previous section but allow training to run for 2000 epochs. We use the same manually chosen distribution over hyperparameters to perform a
random search, drawing 120 random settings of hyperparameters. For each setting we run the learning with 20 different random initializations. For the ListK tasks,
the runtime was very long, so we ran for fewer settings of hyperparameters (28 for Assembly and 10 for Basic Block).
As before, in the Optimized case we report the success rate for the best hyperparameters found and
also for the average across all runs in the random search.
\afterpage{
\begin{landscape}
\begin{table}
\begin{center}
\begin{tabular}{lccc|cccc|ccc}
&&&&\multicolumn{4}{c|}{FMGD}& ILP & SMT & $\sketch$\\
& $\log_{10}(D)$ & $T$ & $N$ & Time & Vanilla & Best Hypers & Average Hypers & Time & Time & Time \\
\toprule
TURING MACHINE\\
\midrule
Invert & 4 & 6 & 5 & 76.5 & 100\% & 100\% & 51\% & 0.6 & 0.7 & 3.1 \\
Prepend zero & 9 & 6 & 5 & 98 & 60\% & 100\% & 37\% &  17.0 & 0.9 & 2.6\\
Binary decrement & 9 & 9 & 5 &  163 & 5\% & 25\% & 2\% & 191.9 & 1.6 & 3.3\\
\bottomrule
BOOLEAN CIRCUITS\\
\midrule
2-bit controlled shift register & 10 & 4 & 8 & - & - & - & - & 2.5 & 0.7 & 2.7\\
Full adder & 13 & 5 & 8 & - & - & - & - &  38 & 1.9 & 3.5\\
2-bit adder & 22 & 8 & 16 & - & - & - & - &  13076.5 & 174.4 & 355.4\\
\bottomrule
BASIC BLOCK\\
\midrule
Access & 14 & 5 & 5 & 173.8 & 15\% & 50\% & 1.1\%& 98.0 & 14.4 & 4.3\\
Decrement & 19 & 18 & 5 &  811.7  & - & 5\% & 0.04\% & - & - & 559.1\\
List-K & 33 & 11 & 5 & - & - & - & - & - & - & 5493.0\\
\bottomrule
ASSEMBLY\\
\midrule
Access & 13 & 5 & 5 & 134.6 & 20\% & 90\% & 16\% & 3.6 &  10.5 & 3.8\\
Decrement & 20 & 27 &  5 & - & - & - & - & - & - & 69.4\\
List-K & 29 & 16 & 5 & - & - & - & - & - & - & 16.8\\
\bottomrule
\end{tabular}
\end{center}
\caption{\label{tbl:benchmarkResults} Benchmark results. For FMGD we present the time in seconds for 1000 epochs and the success rate out of \{20, 20, 2400\} random restarts in the \{Vanilla, Best Hypers and Average Hypers\} columns respectively. For other back-ends we present the time in seconds to produce a synthesized program. The symbol - indicates timeout ($>4$h) or failure of any random restart to converge.
$N$ is the number of provided input-output examples used to specify the task in each case.
}
\end{table}
\end{landscape}
}

Our results are compiled in \rTab{tbl:benchmarkResults}, from which we can draw two high level conclusions:

\paragraph{Back end algorithm. } There is a clear tendency for traditional techniques employing constraint solvers (SMT and $\sketch$) to outperform the machine learning methods, with $\sketch$ being the only system able to solve all of these benchmarks before timeout (see \Secref{sec:sketchFailure}).

Nevertheless, the machine learning methods have qualitatively appealing properties. Firstly, they are primarily \textit{optimizers} rather than solvers\footnote{Both $\sketch$ and SMT (in the form of max-SMT) can also be configured to be optimizers}, and additional terms could be added to the cost function of the optimization to find programs with desired properties (e.g. minimal length \citep{Bunel16} or resource usage). Secondly, FMGD makes the synthesis task fully differentiable, allowing its incorporation into a larger end-to-end differentiable system \citep{Kurach15}. This encourages us to persevere with analysis of the FMGD technique, and in particular to study the surprising failure of this method on the simple boolean circuit benchmarks in \Secref{sec:fmgdBoolean}.

\paragraph{Interpreter models. } \rTab{tbl:benchmarkResults} highlights that the precise formulation of the interpreter model can affect the speed of synthesis. Both the Basic Block and Assembly models are equally expressive, but the Assembly model is biased towards producing straight line code with minimal branching. In all cases where synthesis was successful the Assembly representation is seen to outperform the Basic Block model in terms of synthesis time. The only anomaly is that the Optimized FMGD algorithm is able to find a solution in the Decrement task using the Basic Block model, but not the Assembly model. This could be because the \textit{minimal} solution to this program is shorter in the Basic Block architecture than in the Assembly model ($T=18$ vs. $27$ respectively). We observe in \Secref{sec:overcompleteness} that increasing the size of a model by adding \textit{superfluous} resources can help the FMGD algorithm to converge on a global optimum. However, we generally find that synthesis is difficult if the minimal solution is already large.

\subsection{Zooming in on FMGD Boolean Circuits}
\label{sec:fmgdBoolean}
There is a stark contrast between the performance of FMGD and the alternatives
on the Boolean Circuit problems. On the Controlled Shift and Full Adder benchmarks,
each run of FMGD took $35-80 \times$ as long as the SMT back-end. On top of this, we ran
$120 \times 20 = 2400$ runs during the random search. However, there were no
successful runs.

\subsubsection{Slow convergence}
 While most runs converged to a local optimum
during the 2000 epochs they were allocated, some cases had not. Thus,
we decided to allocate the algorithm $5 \times$ as many epochs (10,000) and run the random
search over. This did produce some successes, although very few. For the
Controlled Shift problem, 1 of 2400 runs converged, and for the
Full Adder, 3 of 2400 runs converged. Thus it does appear that results
could be improved somewhat by running FMGD for longer. However, given the
long runtimes of FMGD relative to the SMT and $\sketch$ back-ends, this would not
change the qualitative conclusions from the previous section.

\subsubsection{Varying the problem dimension}
\label{sec:overcompleteness}
We take inspiration from neural network literature which approaches the issue of stagnation in local minima by increasing the dimension of the problem. It has been argued that local optima become increasingly rare in neural network loss surfaces as the dimension of the hidden layers increase, and instead saddle points become increasingly common~\citep{dauphin2014identifying}. Exchanging local minima for saddle points is beneficial because dynamic learning rate schedules such as \tool{RMSProp} are very effective at handling saddle points and plateaus in the loss function.

\begin{figure}
\begin{center}
\begin{tabular}{ccc}
\multicolumn {1}{l}{(a)}&\multicolumn{1}{l}{\hspace*{0.5cm}(b)}\\
\includegraphics[width=2in]{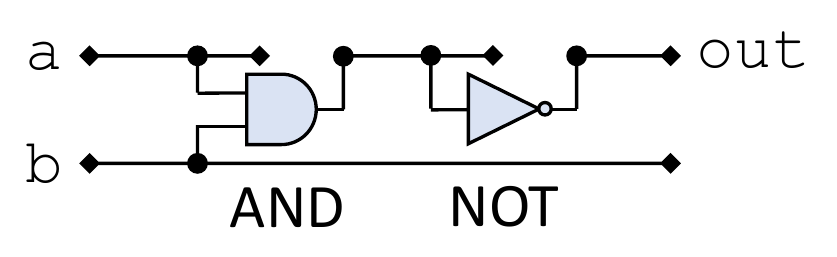} &
\multicolumn{2}{c}{\multirow{2}{*}{
\includegraphics[width=2.4in]{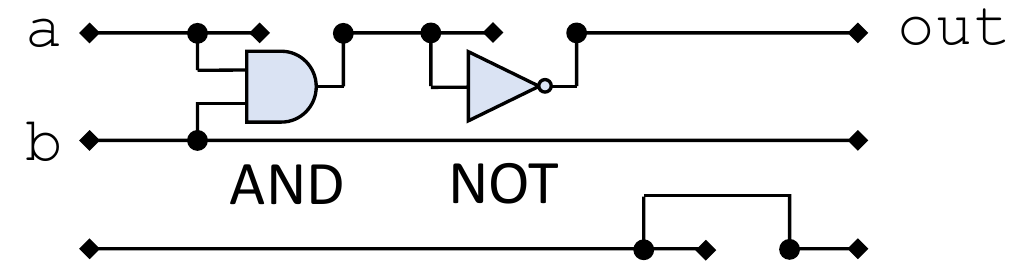}}}\\
\includegraphics[width=2in]{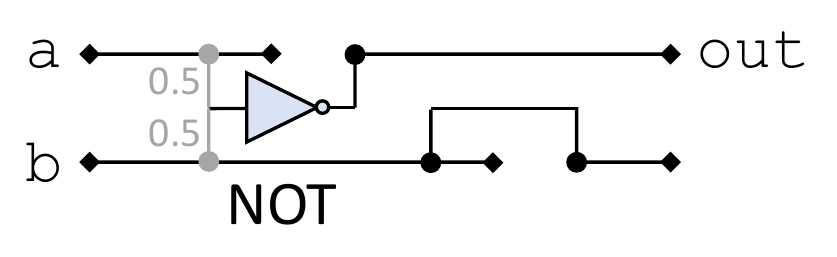}\\
\includegraphics[width=2in]{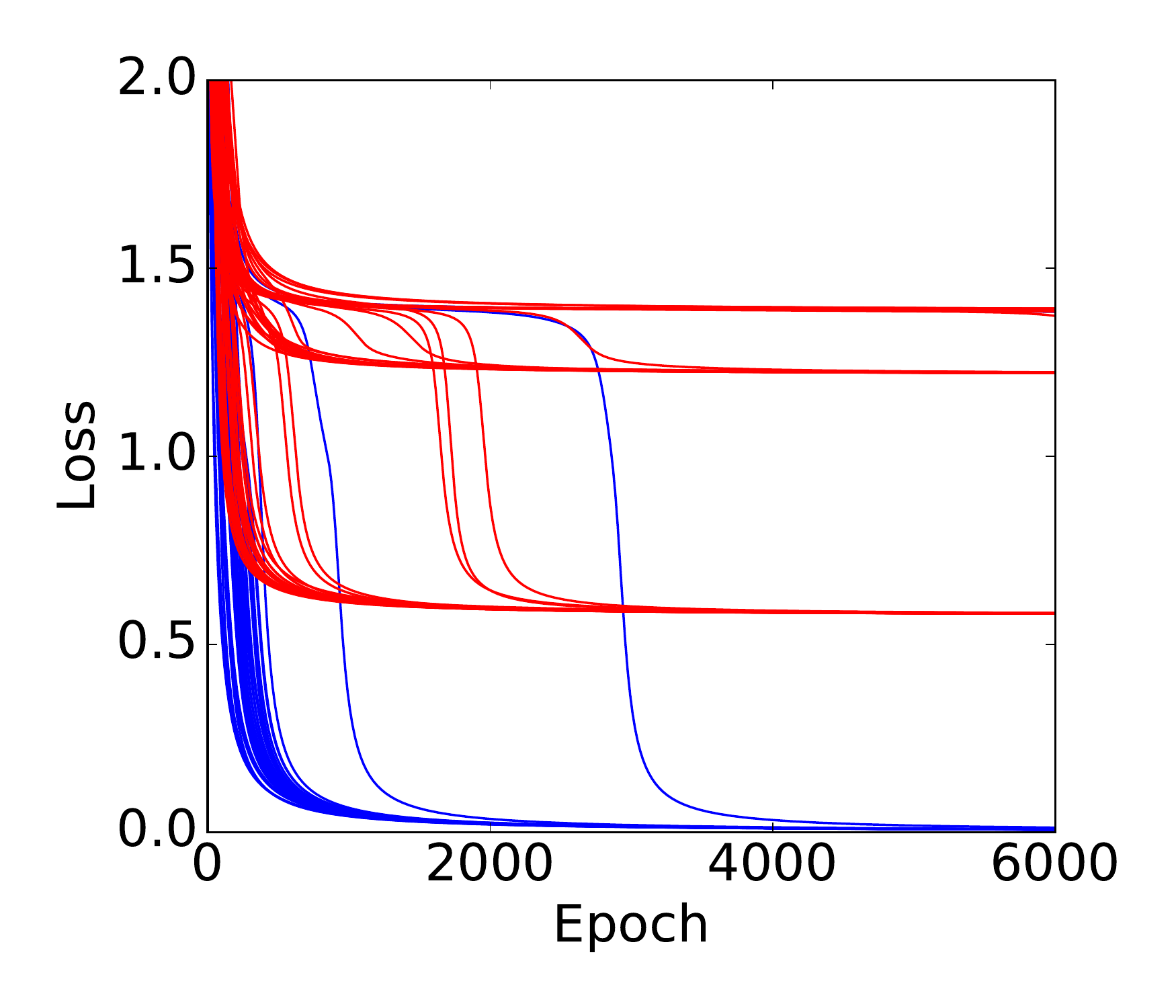} \raisebox{3.5cm}{\hspace*{-1.5cm}{\tool{SGD}}}&\includegraphics[width=2in]{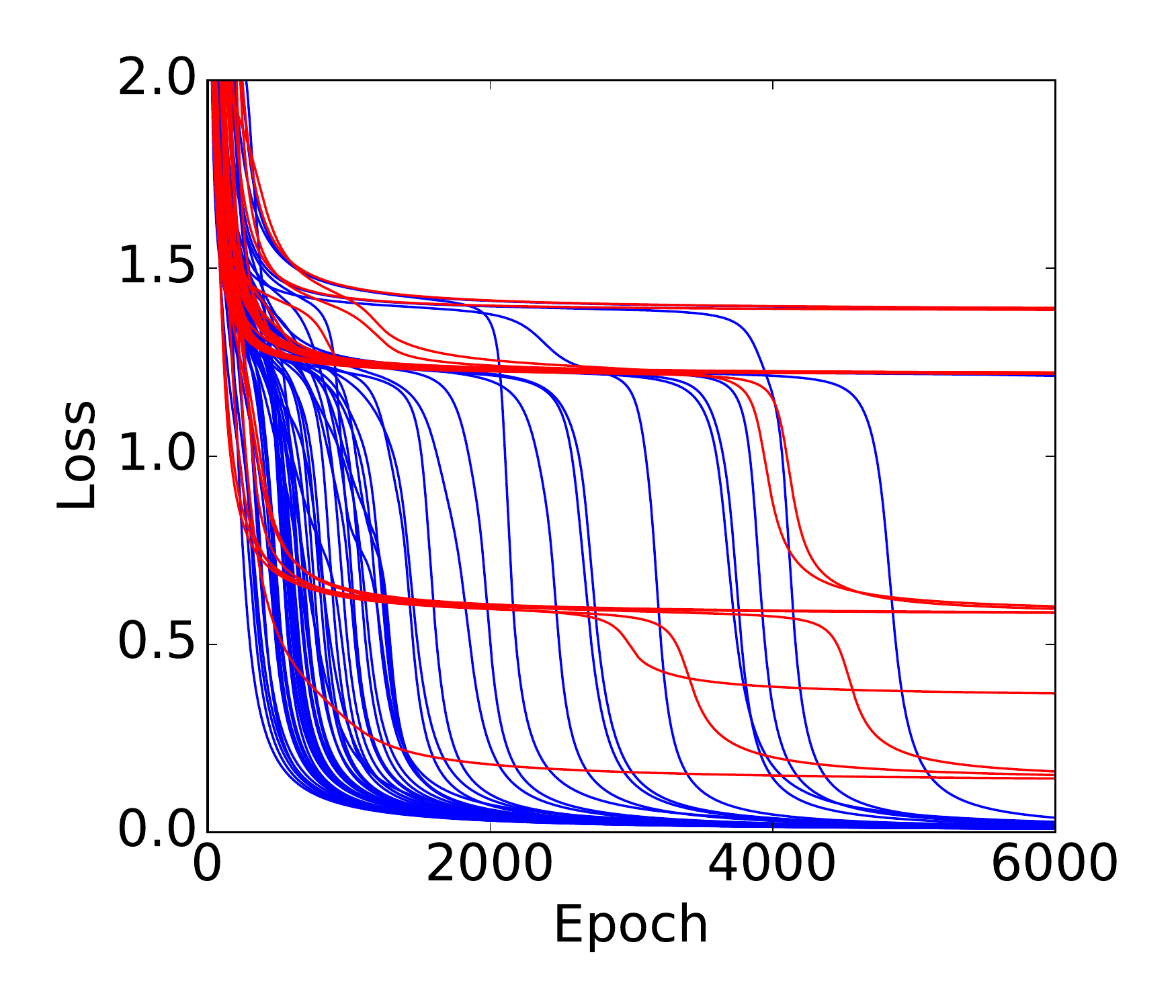}\raisebox{3.5cm}{\hspace*{-1.4cm}{\tool{SGD}}}&\hspace*{-0cm}\includegraphics[width=2in]{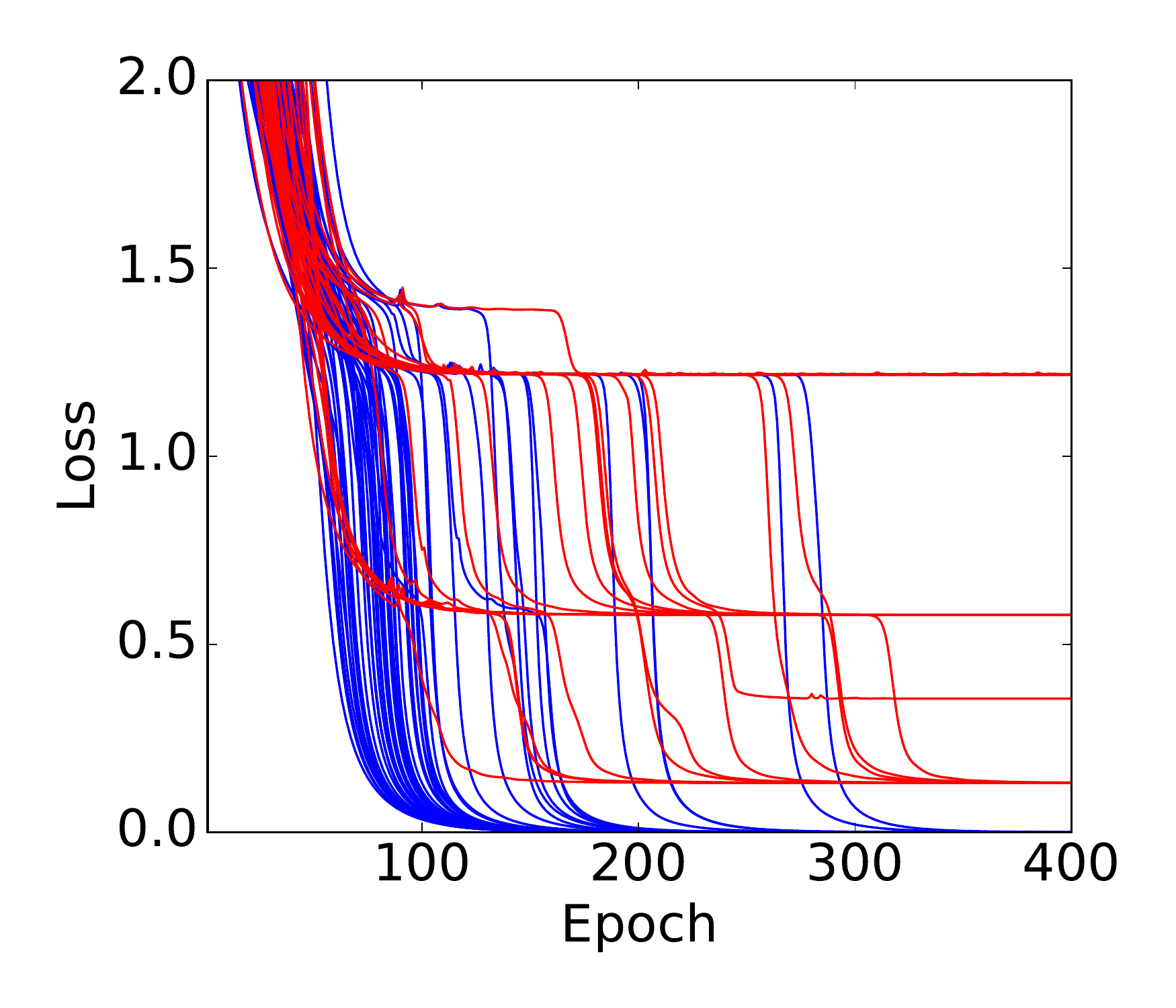}\raisebox{3.5cm}{\hspace*{-2.1cm}{\tool{RMSProp}}}\\
\end{tabular}
\end{center}
\caption{\label{fig:nand} Comparison of the learning trajectories when we increase the resources available to the synthesizer. (a) The minimal solution for producing a \code{NAND} gate, and a locally optimal configuration found by FMGD (the marginal for the input to the first gate puts 50\% weight on both wires). We plot the trajectory of 100 random initializations during learning, highlighting the successful runs in blue and the failures in red. (b) Adding an extra wire and gate changes the learning trajectories significantly, introducing plateaus indicative of saddle-points in the loss function. We can use \tool{RMSProp} rather than SGD (stochastic gradient descent) to navigate these features (note the change in the horizontal scale when using \tool{RMSProp}) }
\end{figure}

To assess how dimensionality affects FMGD, we first take a minimal example in the boolean circuit domain: the task of synthesizing a NAND gate. The minimum solution for this task is shown in \Figref{fig:nand}(a), along with an example configuration which resides at one local minimum of the FMGD loss surface. For a synthesis task involving two gates and two wires, there are a total of 14 independent degrees of freedom to be optimized, and there is only one global optimum. Increasing the available resources to three gates and three wires, gives an optimization problem over 30 dimensions and several global minima. The contrast between the learning trajectories in these two cases is shown in \Figref{fig:nand}. We attempt to infer the presence of saddle points by exploring the loss surface with vanilla gradient descent and a small learning rate. Temporary stagnation of the learning is an indication of a saddle-like feature. Such features are clearly more frequently encountered in the higher dimensional case where we also observe a greater overall success rate (36\% of 100 random initializations converge on a global optimum in the low dimensional case vs. 60\% in the high dimensional case).

\begin{figure}
\begin{center}
\begin{tabular}{ccc}
\includegraphics[height=1.3in]{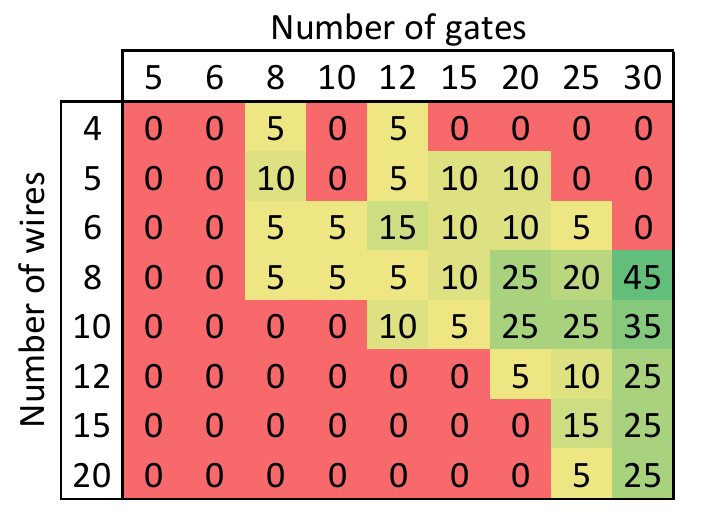} &
\includegraphics[height=1.3in]{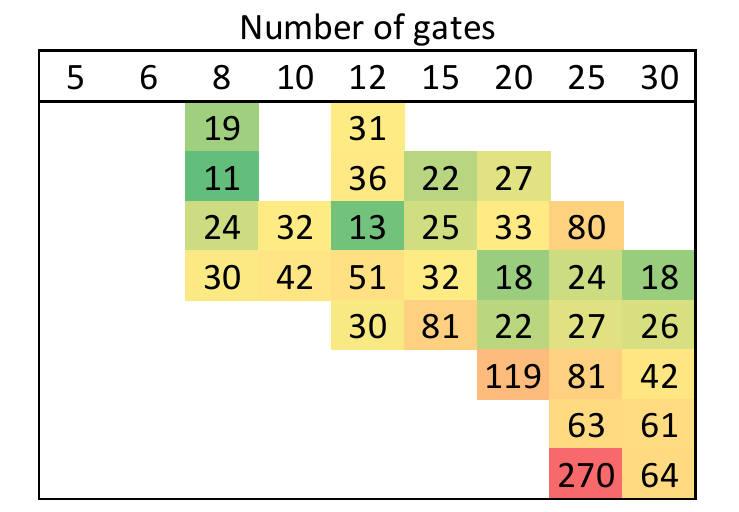}&
\includegraphics[height=1.3in]{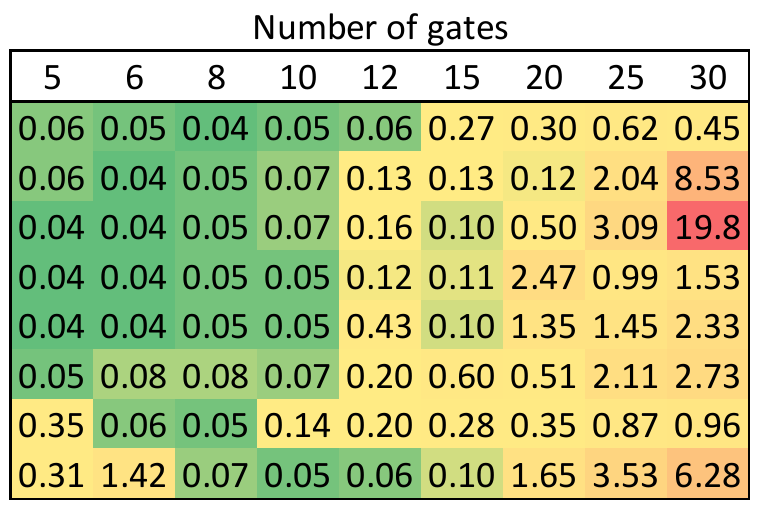}\\
(a) Success rate & (b) Expected solution time & (c) $\sketch$ solution time \\
& (minutes) & (minutes)
\end{tabular}
\end{center}
\caption{\label{fig:overcompleteness} The effect of increasing the dimension of the FMGD synthesis problem. We vary the number of wires and gates used in the synthesis of the full adder citcuit and present both (a) the percentage of 20 random initializations which converge to a global solution and (b) the expected time (in minutes) to solution (time for 1000 epochs / success rate). The solution time for the \sketch|backend is shown in (c) for comparison. }
\end{figure}

These observations are consistent with the intuition from \cite{dauphin2014identifying}, suggesting that we will have more success in the benchmark tasks if we provide more resources (i.e. a higher dimensional problem) than required to solve the task. We perform this experiment on the full adder benchmark by varying the number of wires and gates used in the synthesized solution. The results in \Figref{fig:overcompleteness} show the expected trend, with synthesis becoming more successful as the number of redundant resources increases above the minimal 4 wires and 5 gates. Furthermore, we see no clear increase in the expected time for FMGD to arrive at a solution as we increase the problem size (calculated as the time for 1000 epochs divided by the success rate). This is dramatically different to the trend seen when applying constraint solvers to the synthesis problem, where broadly speaking, increasing the number of constraints in the problem increases the time to solution (see \Figref{fig:overcompleteness}(c)).

\begin{figure}
\begin{center}
\includegraphics[height=2in]{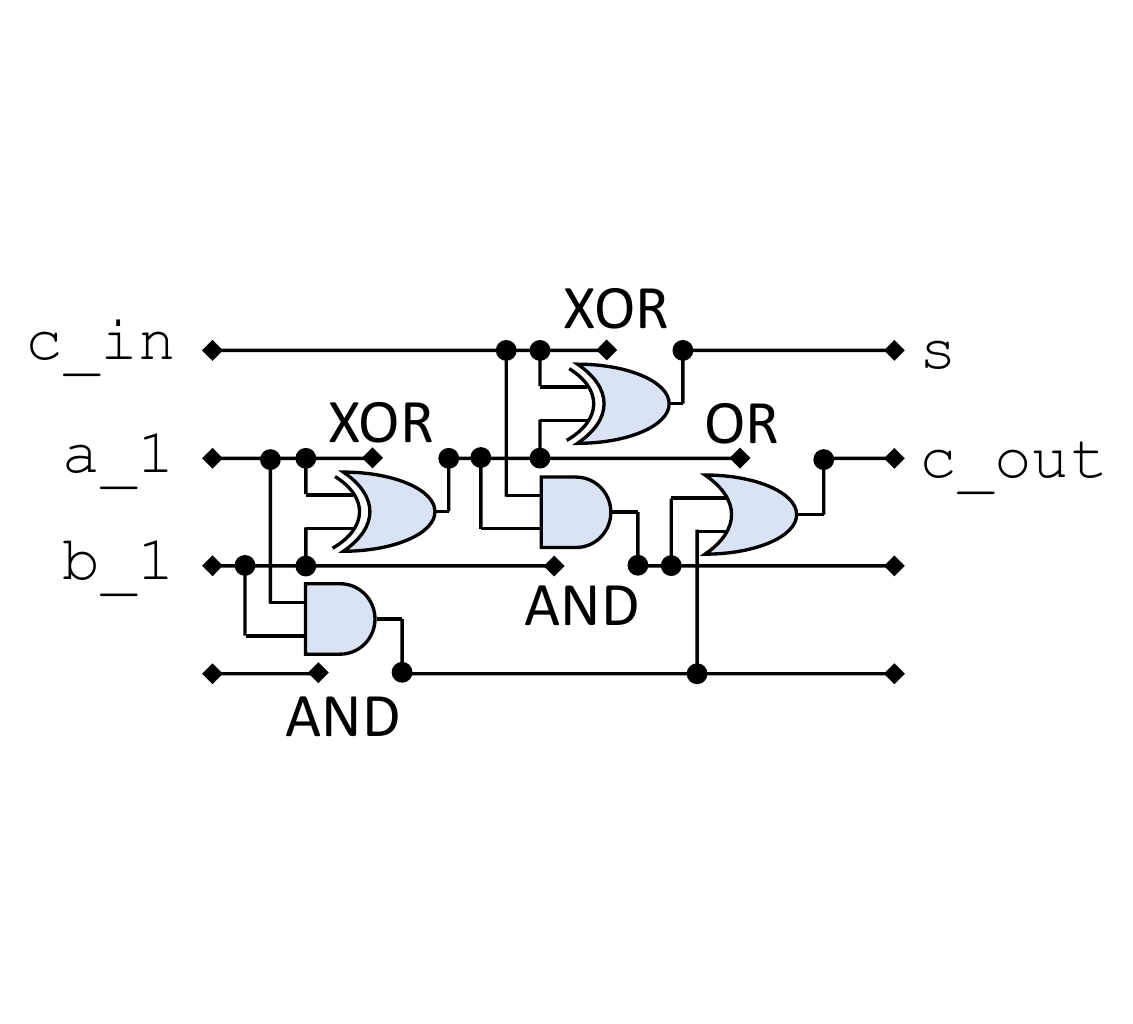}\includegraphics[height=2in]{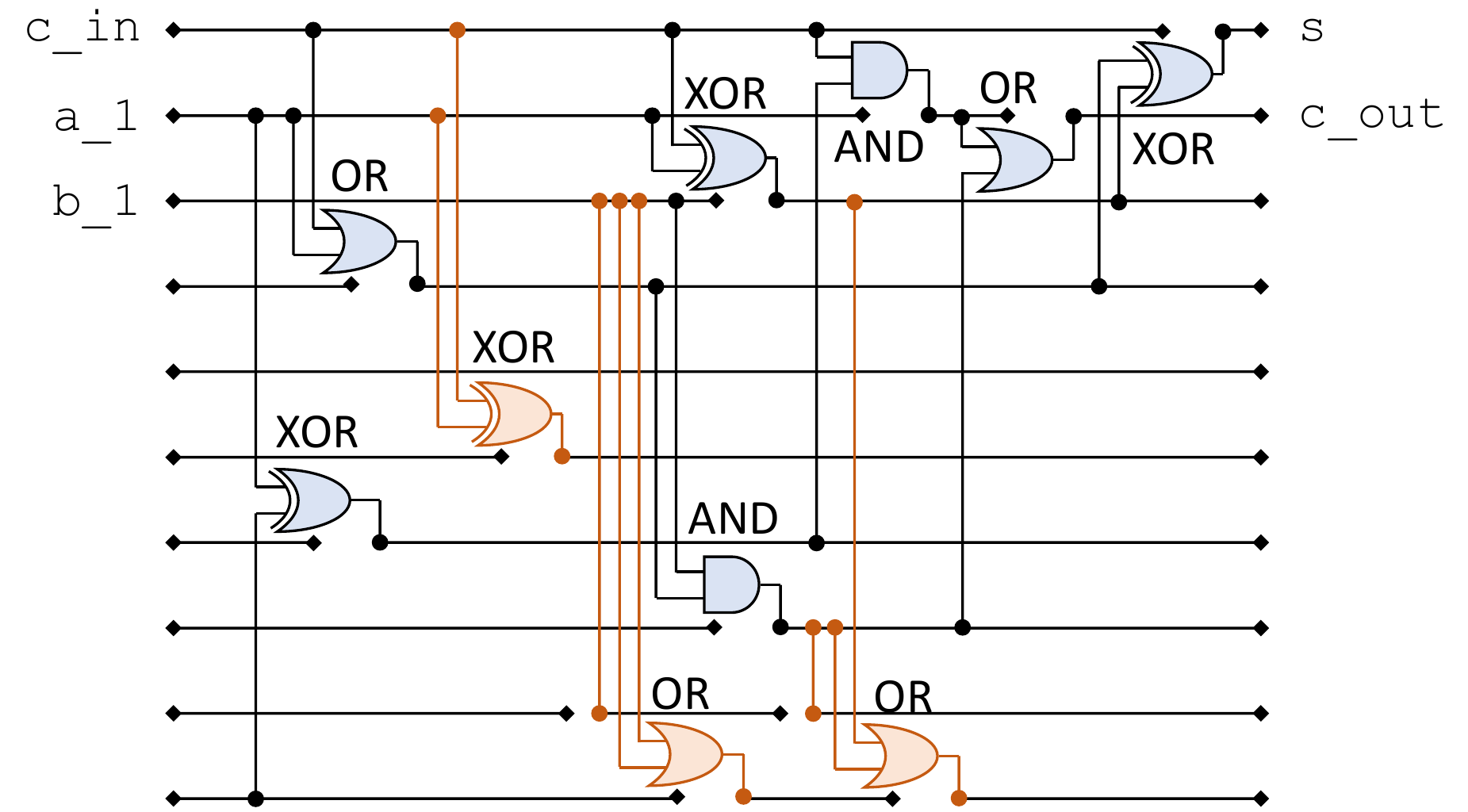}
\end{center}
\caption{\label{fig:overcompleteCircuit} Comparison of the minimal Full Adder solution and the synthesized solution when redundant resources are provided. The gates highlighted in orange can be removed in post-processing without affecting the output.}
\end{figure}

This feature of FMGD is particularly interesting when the minimal resources required to solve a problem is not known before synthesis. Whereas over-provisioning resources will usually harm the other back-ends we consider, it can help FMGD. The discovered programs can then be post-processed to recover some efficiency (see \Figref{fig:overcompleteCircuit})

\subsection{Challenge Benchmark}
\label{sec:sketchFailure}
Before leaving this section, we note that \sketch~has so far solved all of the benchmark tasks. To provide a goal for future work, we introduce a final benchmark which none of the back-ends are currently able to solve from 5 input-output examples even after 40 hours:

\vspace{5mm}
\begin{tabular}{p{1in}cccccp{3.1in}}
\toprule
ASSEMBLY & $M$ & $R$ & $B$ & $\log_{10}D$ & $T$ & Description\\
\midrule
Merge & 17 & 6 & 22 & 103 & 69 & Merge two contiguous sorted lists into one contiguous sorted list. The first entries in the initial heap are $\code{heap}_0\code{[0]} = p_1$, $\code{heap}_0\code{[1]} = p_2$, $\code{heap}_0\code{[2]} = p_{\rm out}$, where $p_1$ ($p_2$) is a pointer to the first (second) sorted sublist (terminated with a 0), and $p_{\rm out}$ is a pointer to the head of the desired output list. All elements of the sorted sublists are  larger than 0 and all unused cells in $\code{heap}_0$ are initialized to 0.\\
\bottomrule

\end{tabular}

\section{Related Work}
\label{sec:relatedWork}

\paragraph{Probabilistic Programming and Graphical Models}

There are now many probabilistic programming systems specialized to
different use-cases. One dominant axis of variability is in the
expressiveness of the language. Some probabilistic programming
languages, exemplified by Church \citep{goodman2008church}, allow
great freedom in the expressibility of the language, including
constructs like recursion and higher order functions. The cost of the
expressibility in the language is that the inference techniques cannot
be as specialized, and thus these systems tend to use general Markov
Chain Monte Carlo methods for inference. On the other end of the
spectrum are more specialized systems like Infer.NET
\citep{InferNET14} and Stan \citep{stan2015ppl,stan2015}.  These
systems restrict models to be constructed of predefined building
blocks and do not support arbitrary program constructs like recursion
and higher order functions. They do generally support basic loops and
branching structure, however.  In Infer.NET, for example, loops are
unrolled, and \code{if} statements are handled via special constructs
known as Gates \citep{minka2009gates}. The result is that the program
can be viewed as a finite gated factor graph, on which message passing
inference can be performed.

In these terms, \langname~is most similar to Infer.NET, and its
handling of loops and \code{if} statements are inspired by
Infer.NET. Compared to Infer.NET, \langname~is far more extreme in the
restrictions that it places upon modelling constructs.  The benefit is
that the restricted language allows us to support a broader range of
back-ends. Looking forward, Infer.NET provides inspiration for how
\langname~might be extended to handle richer data types like real numbers
and strings.

Another related line of work is in casting program synthesis as a
problem of inference in probabilistic models.
\cite{gulwani2007program} phrase program synthesis as inference in a
graphical model and use belief propagation inference. In future work,
we would like to create a belief propagation-based back-end for
\langname.  The problem of inducing samplers for probability
distributions has also been cast as a problem of inference in a
probabilistic program \citep{perov2016automatic}. \cite{lake2015human}
induce probabilistic programs by performing inference in a
probabilistic model describing how primitives are composed to form
types and instances of types.

\paragraph{Neural Networks with Memory}
In common neural network architectures handling sequences of inputs, ``memory''
only manifests itself as the highly compressed state of the network.
This is problematic when the task at hand requires to relate inputs that are far
apart from each other, which more recent models try to mitigate using tools
such as Long Short-Term Memory~\citep{Hochreiter97,Graves13} and Gated Recurrent
Units~\citep{Cho14}.
However, such recurrent units do not entirely solve the problems with long-range
interactions, and a range of additional techniques have been employed to improve
results (e.g., \citep{Mikolov14, Koutnik14}).

An alternative solution to this problem is to extend networks by providing
access to external storage.
Initial extensions provided a stack~\citep{Giles89} or a scratch pad simplified
to a stack~\citep{Mozer92}, and the controlling network learned when to push data
to and pop (load) data from that stack.\footnote{Interestingly enough,
  extracting an interpretable deterministic pushdown automaton from a trained
  stack-using recurrent network was already proposed in \citep{Das92}.}
Recently, similar ideas have been picked up again, leading to stack and queue-augmented
recurrent nets~\citep{Joulin15,grefenstette2015learning},
memory networks with freely addressable
storage~\citep{Weston14,Sukhbaatar15}, and extensions that additionally use
registers for intermediate results~\citep{Kurach15}.

\paragraph{Neural Networks Learning Algorithms}
Recurrent neural networks with access to memory are, essentially, learnable
implementations of the Von Neumann architecture.
A number of recent advances build on this observation to learn
algorithms from input-output data
\citep{Graves14,Joulin15,Neelakantan15,Reed15,Zaremba16}.
While these approaches differ in (a) the underlying execution models (e.g., Turing
Machines, Random Access Machines, Stack Automata), (b) learning methods (e.g., from
input/output samples or action sequences, supervised or by reinforcement
learning), and (c) program domains (arithmetic, simple data structure manipulation,
image manipulation), they share the overall idea of training a deep neural
network that learns to manipulate data by repeatedly calling deterministic
``modules'' (or ``actions'' or ``interfaces'') from a predefined set.
These models are able to learn and repeat simple algorithmic patterns, but are
all not interpretable; what they have learned only becomes evident through
actions on concrete inputs.

Very recent work has improved on this aspect and is closest to our approach.
To support adaptive neural compilation~\citep{Bunel16}, a machine model similar
to our assembly model (cf. \rSC{sec:assembly}) was introduced.
This allows a user to \emph{sketch} a (partial) program as input, and then use deep
learning methods to optimise it.
The result is again a program in the chosen assembly language, and can be
displayed easily.
Differentiable Forth~\citep{Riedel16} is a similar step in this direction, where
the learning task is to fill in holes in a partial Forth program.

\paragraph{Program Synthesis} The area of program synthesis has recently seen a renewed interest in the programming language community~\citep{sygus}. There have been many synthesis techniques developed for a wide range of problems including data wrangling~\citep{cacm12,flashmeta}, inference of efficient synchronization in concurrent programs,  synthesizing efficient low-level code from partial programs~\citep{solar2005bitstreaming}, compilers for low-power spatial architectures~\citep{chlorophyll}, efficient compilation of declarative specifications~\citep{comfusy}, statistical code completion~\citep{bigcode}, and automated feedback generation for programming assignments~\citep{autograder}. These techniques can be broadly categorized using three dimensions: i) specification mechanism, ii) complexity of hypothesis space, and iii) search algorithm. The different forms of specifications include input-output examples, partial programs, reference implementation, program traces etc. The hypothesis space of possible programs is typically defined using a domain-specific language, which is designed to be expressive enough to encode majority of desired tasks but at the same time concise enough for efficient learning. Finally, some of the common search algorithms include constraint-based symbolic synthesis algorithms~\citep{sketch,cvc4synthesis}, smart enumerative algorithms with pruning~\citep{transit}, version-space algebra based search algorithms~\citep{cacm12,gulwani2011automating}, and stochastic search~\citep{superoptimization}. There has also been some recent work on learning from inputs in addition to the input-output examples to guide the synthesis algorithm~\citep{blinkfill}, and synthesizing programs without any examples by performing a joint inference over the program and the inputs to recover compressed encodings of the observed data~\citep{ellisnips15}.

In this work, we are targeting specifications based on input-output examples as this form of specification is most natural to the work in the machine learning community. For defining the hypothesis space, we use our probabilistic programming language \langname, and we currently support compilation to intermediate representations for four inference (search) algorithms.
The key difference between our work and most of the previous work in the program synthesis community is that our language is built to allow compilation to different inference algorithms (from both the machine learning community and programming languages community) which enables like-to-like comparison. We note that another recent effort SyGuS~\citep{sygus} aims to unify different program synthesis approaches using a common intermediate format based on context-free grammars so that different inferences techniques can be compared, but the \langname language allows for encoding richer programming models than SyGuS, and also allows for compilation to gradient-descent based inference algorithms.


\section{Discussion \& Future Work}
\label{sec:discussion}

We presented \langname, a probabilistic programming language for specifying
IPS problems. \langname~can be used in combination with the FMGD back-end to
produce differentiable interpreters for a wide range of program representations
and languages. \langname~has several other back-ends including one based on
linear programming and two that are strong alternatives from the programming
languages community.

The biggest take-away from the experimental results is that the methods from
programming languages significantly outperform the machine learning approaches.
We believe this is an important take-away for machine learning researchers
studying program synthesis.
However, we remain optimistic about the future of machine learning-based approaches
to program synthesis, and we do not wish to discourage work in this area.
Quite the opposite; we hope that this work stimulates further research in the
area and helps to clarify how machine learning is likely to be useful.
The setting in this work is a minimal version of
the program synthesis problem, in which the main challenge is efficiently
searching over program space for programs that meet a given input-output
specification. The conclusion from our experiments is that gradient descent
is inferior to constraint-based discrete search algorithms for this task.

Our results also raise an interesting question when taken in
comparison to \citep{Kurach15}. The NRAM model is reported to solve
problems that our FMGD approach was not able to. We would like to
better understand what the source of this discrepancy is.  The two
main differences are in the execution model and in the program
parameterization.  In the NRAM execution model, all instructions are
executed in all timesteps (possibly multiple times), and it is up to
the controller to decide how to wire them up.  This creates additional
parallelism and redundancy relative to the Basic Block or Assembly
models. We speculate that this makes it possible in the NRAM model for
multiple hypotheses to be developed at once with little overlap in the
memory locations used. This property may make the optimization easier.
The other major difference is in the controller. The NRAM model uses a
neural network controller that maps from the state of registers to the
operations that are to be performed.  In the Basic Block and Assembly
models, this is done via the instruction pointer that is updated based
upon control flow decisions in the program. We speculate that the
neural network controller operates in a less constrained space (since
a different operation may be performed at each timestep if the model
pleases), and it offers some bias to the search. We suspect neural
networks may be biased towards repeating circuits in the earlier
stages of training due to their typical behavior of first predicting
averages before specializing to make strongly input-dependent
predictions. In future work we would like to explore these questions
more rigorously, in hopes of finding general principles that can be
used to develop more robust inference algorithms.

In future work, there are several extensions to \langname~that we would like to develop.
First, we would like to extend the \langname~language in several ways.
We would like to support non-uniform priors over program variables (which will require
converting the SMT and Sketch back-ends to use max-SMT solvers).
There are several data types that we would like to support, including
floating point numbers, strings, and richer data types.
Second, we would
like to continue to expand the number of back-ends. Natural next steps are
back-ends based on local search or Markov Chain Monte Carlo, and on
message passing inference in graphical models, perhaps taking inspiration from
\cite{sontag2008tightening}. Third, we would like to build
higher level languages on top of \langname, to support more compact specification
of common \langname~programming patterns.

More generally, we believe the opportunities for IPS come not from improving
discrete search in this setting, but in re-phrasing the program synthesis
problem to be more of a pattern-matching and big-data problem, and in
augmenting the specification beyond just input-output examples (for example,
incorporating natural language). In these cases, the importance of the
discrete search component decreases, and we believe there to be many
opportunities for machine learning. As we move forward in these directions,
we believe \langname~will continue to be valuable, as it makes it easy to
build a range of differentiable interpreters to be used in conjunction with
larger learning systems.

\section*{Acknowledgements}
We thank several people for discussions that helped improve this report:
Tom Minka for discussions related to the Gates LP relaxation;
John Winn for several discussions related to probabilistic programming and gates;
Ryota Tomioka for discussions related to the FMGD loss surface;
Andy Gordon for pushing us towards the probabilistic programming formulation of \langname;
Abdel-rahman Mohamed for discussions related to neural networks and program synthesis;
Jack Feser for being the first non-author \langname~user; 
Aditya Nori for helpful discussions about program synthesis; and 
Matej Balog for a critical reading of this manuscript.
\appendix
\section{Proof of Lemma 1}
\label{sec:fmgd-proof}

\textbf{Lemma 1.} \textit{All island structures have zero gradient.}

\begin{proof}


Notationally, let $\boldsymbol{s} = \{ s_i(a) \given i \in \{1, \ldots, K\}, a \in \{0, 1\}\}$
be the free parameters, where $s_i(a)$ is the unnormalized log
probability that $x_i$ is equal to $a$. The probability over $x_i$
is then given by a softmax; i.e.,
$p(x_i = a) = \mu_i(a) = \frac{\exp s_i(a)}{\exp s_i(0) + \exp s_i(1)}$.
Let $\boldsymbol{\mu}$ be the set
$\{ \mu_i(a) \given i \in \{1, \ldots, K\}, a \in \{0, 1\}\}$.
Let the objective $o(\boldsymbol{s})$ be the log probability of the observations,
i.e., $o(\boldsymbol{s})
= \sum_i \log p(y_i = 0 \given \boldsymbol{s})
= \sum_i \log p(y_i = 0 \given \boldsymbol{\mu})$.

The plan is to show that for each island structure described above,
the partial derivative
$\frac{\partial o(\boldsymbol{s})}{\partial s_i(a)}$
is 0 for every $i$ and $a$. This can be done by computing the partial
derivatives and showing that they are 0 for each possible local
configuration that arises in an island structure.

First, let us derive the gradient contribution from a single observation
$y_i = 0$. By the definition of parity and the FMGD model,
\begin{align}
  p(y_i = 0 \given \boldsymbol{\mu}) & = p(x_i = 0) p(x_{i+1} = 0) + p(x_i = 1) p(x_{i+1} = 1) \\
  & = \mu_{i}(0) \mu_{i+1}(0) + \mu_{i}(1) \mu_{i+1}(1).
\end{align}
The partial derivative $\frac{\partial \log p(y_i = 0)}{\partial s_j(a)}$
can be computed via the chain rule as
\begin{align}
  \frac{\partial \log p(y_i = 0)}{\partial s_j(a)}
  & =
  \frac{\partial \log p(y_i = 0)}{\partial p(y_i = 0)}
  \frac{\partial p(y_i = 0)}{\partial s_{j}(b)}.
\end{align}
Each of these can be computed straight-forwardly.
The partial derivative $\frac{\partial \log p(y_i = 0)}{\partial p(y_i = 0)}$ is
$\frac{1}{p(y_i = 0)}$.
The partial derivative $\frac{\partial p(y_i = 0)}{\partial s_{j}(b)}$ is
  as follows:
\begin{align}
  \frac{\partial p(y_i = 0)}{\partial s_{j}(b)} & =
  \left\{ \begin{array}{cl}
    \frac{\partial \mu_{i}(0)}{\partial s_{i}(b)} \mu_{i+1}(0) +
    \frac{\partial \mu_{i}(1)}{\partial s_{i}(b)} \mu_{i+1}(1)
    & \hbox{ if $j = i$} \\
    \mu_{i}(0) \frac{\partial \mu_{i+1}(0)}{\partial s_{i+1}(b)} +
    \mu_{i}(1) \frac{\partial \mu_{i+1}(1)}{\partial s_{i+1}(b)}
    & \hbox{ if $j = i + 1$} \\
    0 & \hbox{ otherwise.}
  \end{array} \right.
\end{align}
Finally, the partial derivative of the softmax is
\begin{align}
  \frac{\partial \mu_{i}(b)}{\partial s_{i}(a)}
  & =
  \mu_{i}(b) \frac{\partial \log \mu_{i}(b)}{\partial s_{i}(a)} \\
      & =
      \mu_{i}(b)
      \frac{\partial}{\partial s_{i}(a)}
      \left[
        s_i(b) - \log \left( \exp s_i(0) + \exp s_i(1) \right)
        \right] \\
      & =
      \mu_{i}(b)
      \left( 1\{a = b\} - \mu_i(a) \right).
\end{align}
Putting these together, we get the full partial derivatives:
{\footnotesize
\begin{align}
  \frac{\partial \log p(y_i = 0)}{\partial s_{j}(a)} & =
  \frac{1}{p(y_i = 0)} \cdot
  \left\{ \begin{array}{cl}
    \mu_{i}(0)
    \mu_{i+1}(0)
    \left( 1\{a = 0\} - \mu_i(a) \right)
    +
    \mu_{i}(1)
    \mu_{i+1}(1)
    \left( 1\{a = 1\} - \mu_i(a) \right)
    & \hbox{ if $j = i$} \\
    \mu_{i}(0)
    \mu_{i+1}(0)
    \left( 1\{a = 0\} - \mu_{i+1}(a) \right)
    +
    \mu_{i}(1)
    \mu_{i+1}(1)
    \left( 1\{a = 1\} - \mu_{i+1}(a) \right)
    & \hbox{ if $j = i + 1$} \\
    0 & \hbox{ otherwise}
  \end{array} \right.
\end{align}
}
Each $s_j(a)$ contributes to two terms in the objective:
$\log p(y_{j-1} = 0)$ and $\log p(y_{j} = 0)$. Thus the gradient
on $s_j(a)$ is the sum of gradient contributions from these two terms:
\begin{align}
  \frac{\partial o(\boldsymbol{s})}{\partial s_j(a)}
  & =
  \frac{\partial \log p(y_{j-1} = 0)}{\partial s_{j}(a)} +
  \frac{\partial \log p(y_j = 0)}{\partial s_{j}(a)}.
\end{align}

Restricting attention to $a=0$ (the case where $a=1$ follows
similarly and is omitted), we can simplify further:
\begin{align}
  \frac{\partial o(\boldsymbol{s})}{\partial s_j(0)}
  = &
  \frac{\partial \log p(y_{j-1} = 0)}{\partial s_{j}(0)} +
  \frac{\partial \log p(y_j = 0)}{\partial s_{j}(0)} \\
  = &
  \frac{1}{p(y_{j-1} = 0)}
  \left(
    \mu_{j-1}(0)
    \mu_{j}(0)
    ( 1 - \mu_{j}(0))
    +
    \mu_{j-1}(1)
    \mu_{j}(1)
    (-\mu_{j}(0))
  \right) \\
  & +
  \frac{1}{p(y_{j} = 0)}
  \left(
    \mu_{j}(0)
    \mu_{j+1}(0)
    \left( 1 - \mu_j(0) \right)
    +
    \mu_{j}(1)
    \mu_{j+1}(1)
    (-\mu_j(0))
  \right) \\
  = &
  \frac{1}{p(y_{j-1} = 0)}
  \left(
    \mu_{j-1}(0)
    \mu_{j}(0)
    \mu_{j}(1)
    -
    \mu_{j-1}(1)
    \mu_{j}(1)
    \mu_{j}(0)
  \right) \\
  & +
  \frac{1}{p(y_{j} = 0)}
  \left(
    \mu_{j}(0)
    \mu_{j+1}(0)
    \mu_j(1)
    -
    \mu_{j}(1)
    \mu_{j+1}(1)
    \mu_j(0)
    \right) \\
    = &
    \mu_j(0) \mu_j(1)
    \left(
      \frac{\mu_{j-1}(0) - \mu_{j-1}(1)}{p(y_{j-1} = 0)}
      +
      \frac{\mu_{j+1}(0) - \mu_{j+1}(1)}{p(y_{j} = 0)}
    \right)
\end{align}
The first note is that if $\mu_j(0) = 0$ or $\mu_j(1) = 0$, then
the gradient for $s_j(0)$ is 0. Thus, we only need to consider
triplets $\left(\mu_{j-1}(\cdot), \mu_{j}(\cdot), \mu_{j+1}(\cdot)\right)$
where $\mu_{j}(a) = .5$. The only two cases that arise in island
structures are $(0, .5, 1)$ and $(1, .5, 0)$. Consider the first
case, $(0, .5, 1)$:
\begin{align}
    & =
    \mu_j(0) \mu_j(1)
    \left(
      \frac{\mu_{j-1}(0) - \mu_{j-1}(1)}{p(y_{j-1} = 0)}
      +
      \frac{\mu_{j+1}(0) - \mu_{j+1}(1)}{p(y_{j} = 0)}
    \right) \\
    & =
    .5 \cdot .5 \cdot
    \left(
      \frac{1 - 0}{.5}
      +
      \frac{0 - 1}{.5}
    \right) \\
    & =
    .5 \cdot .5 \cdot
    \left(
    2 - 2
    \right) = 0.
\end{align}
The second case $(1, .5, 0)$ is similar:
\begin{align}
    & =
    \mu_j(0) \mu_j(1)
    \left(
      \frac{\mu_{j-1}(0) - \mu_{j-1}(1)}{p(y_{j-1} = 0)}
      +
      \frac{\mu_{j+1}(0) - \mu_{j+1}(1)}{p(y_{j} = 0)}
    \right) \\
    & =
    .5 \cdot .5 \cdot
    \left(
      \frac{0 - 1}{.5}
      +
      \frac{1 - 0}{.5}
    \right) \\
    & =
    .5 \cdot .5 \cdot
    \left(
    -2 + 2
    \right) = 0.
\end{align}
Thus, the gradients with respect to $s_j(a)$ are zero for all
$j$ and $a$ when an island-structured configuration is given.
The objective is clearly suboptimal (since some $p(y_i = 0)$
are less than 1 at the island boundaries).
Thus, each island structure is a suboptimal local optimum.

\comment{
We compute gradients arising for all triplets of $(x_{i-1}, x_{i}, x_{i+1})$
configurations (and the $y$ values that depend on them) that arise under
the island structure described above. If the gradient for all
overlapping triplets is $\boldsymbol{0}$, then the total gradient is $\boldsymbol{0}$
and we are at a local optimum.

First, let us derive the gradient contribution from a single observation
$y_i = 0$. By the definition of parity and the FMGD model,
\begin{align}
  p(y_i = 0) & = p(x_i = 0) p(x_{i+1} = 0) + p(x_i = 1) p(x_{i+1} = 1) \\
  & = \mu_{i}(0) \mu_{i+1}(0) + \mu_{i}(1) \mu_{i+1}(1),
\end{align}
where $\mu_i(a) = \frac{\exp s_{i}(a)}{\exp s_{i}(0) + \exp s_{i}(1)}$.

The partial derivative $\frac{\partial p(y_i = 0)}{\partial s_{j}(a)}$ is
  as follows:
\begin{align}
  \frac{\partial p(y_i = 0)}{\partial s_{j}(a)} & =
  \left\{ \begin{array}{cl}
    \frac{\partial \mu_{i}(0)}{\partial s_{i}(a)} \mu_{i+1}(0) +
    \frac{\partial \mu_{i}(1)}{\partial s_{i}(a)} \mu_{i+1}(1)
    & \hbox{ if $j = i$} \\
    \mu_{i}(0) \frac{\partial \mu_{i+1}(0)}{\partial s_{i+1}(a)} +
    \mu_{i}(1) \frac{\partial \mu_{i+1}(1)}{\partial s_{i+1}(a)}
    & \hbox{ if $j = i + 1$} \\
    0 & \hbox{ otherwise}
  \end{array} \right.
\end{align}

For completeness, the partial derivatives of the softmax are
\begin{align}
  \frac{\partial \mu_{i}(b)}{\partial s_{i}(a)}
  & =
  \mu_{i}(b) \frac{\partial \log \mu_{i}(b)}{\partial s_{i}(a)} \\
      & =
      \mu_{i}(b)
      \frac{\partial}{\partial s_{i}(a)}
      \left[
        s_i(b) - \log \left( \exp s_i(0) + \exp s_i(1) \right)
        \right] \\
      & =
      \mu_{i}(b)
      \left( 1\{a = b\} - \mu_i(a) \right)
\end{align}

Putting the two together:
\begin{align}
  \frac{\partial p(y_i = 0)}{\partial s_{j}(a)} & =
  \left\{ \begin{array}{cl}
    \mu_{i}(0)
    \mu_{i+1}(0)
    \left( 1\{a = 0\} - \mu_i(a) \right)
    +
    \mu_{i}(1)
    \mu_{i+1}(1)
    \left( 1\{a = 1\} - \mu_i(a) \right)
    & \hbox{ if $j = i$} \\
    \mu_{i}(0)
    \mu_{i+1}(0)
    \left( 1\{a = 0\} - \mu_{i+1}(a) \right)
    +
    \mu_{i}(1)
    \mu_{i+1}(1)
    \left( 1\{a = 1\} - \mu_{i+1}(a) \right)
    & \hbox{ if $j = i + 1$} \\
    0 & \hbox{ otherwise}
  \end{array} \right.
\end{align}

\paragraph{Case $(x_{i-1}, x_{i}, x_{i+1}) = (0, 0, 0)$. } 

Here, $\mu_j(0) = 1$ for $j=i-1, i, i+1$.
Consider the gradient coming from $p(y_i = 0)$.
\begin{align}
  \frac{\partial p(y_i = 0)}{\partial s_{i}(a)} & =
    \mu_{i}(0) \left( 1\{a = 0\} - \mu_i(a) \right)
    \mu_{i+1}(0) +
    \mu_{i}(1) \left( 1\{a = 1\} - \mu_i(a) \right)
    \mu_{i+1}(1) \\
    & = \left( 1\{a = 0\} - \mu_i(a) \right).
\end{align}

The other term comes from the effect of $\mu_{i+1}(a)$ on $p(y_i = 0)$:
\begin{align}
  \frac{\partial p(y_i = 0)}{\partial s_{i+1}(a)} & =
    \mu_{i}(0)
    \mu_{i+1}(0) \left( 1\{a = 0\} - \mu_{i+1}(a) \right)
    +
    \mu_{i}(1)
    \mu_{i+1}(1) \left( 1\{a = 1\} - \mu_{i+1}(a) \right) \\
    & = \left( 1\{a = 0\} - \mu_{i+1}(a) \right)
\end{align}

In both cases, if $a=0$, then this evaluates to $1 - 1$. If $a=1$, this evalutes
to $0 - 0$.
The case where $(x_{i-1}, x_{i}, x_{i+1}) = (1, 1, 1)$ follows similarly.

\paragraph{Case $(0, .5, 1)$. }
Consider the gradient on $s_i(a)$ coming from $p(y_{i-1} = 0)$ and $p(y_{i} = 0)$:
\begin{align}
  \frac{\partial p(y_{i-1} = 0)}{\partial s_{i}(a)} & =
    \mu_{i-1}(0)
    \mu_{i}(0)
    \left( 1\{a = 0\} - \mu_{i-1}(a) \right)
    +
    \mu_{i-1}(1)
    \mu_{i}(1)
    \left( 1\{a = 1\} - \mu_{i-1}(a) \right) \\
    & = .5 \left( 1\{a = 0\} - \mu_{i-1}(a) \right) \\
    & = \left\{ \begin{array}{cl}
      .5 \left( 1 - 1 \right) & \hbox{if $a = 0$} \\
      .5 \left( -\mu_{i-1}(a) \right) & \hbox{if $a = 1$} \\
    \end{array} \right. \\
  \frac{\partial p(y_{i} = 0)}{\partial s_{i}(a)} & =
    \mu_{i}(0)
    \mu_{i+1}(0)
    \left( 1\{a = 0\} - \mu_{i+1}(a) \right)
    +
    \mu_{i}(1)
    \mu_{i+1}(1)
    \left( 1\{a = 1\} - \mu_{i+1}(a) \right) \\
& = .5 \left( 1\{a = 1\} - \mu_{i+1}(a) \right)
\end{align}

The total gradient on $s_i(a)$ is then
\begin{align}
&  .5 \left( 1\{a = 1\} - \mu_i(a) \right)
  + .5 \left( 1\{a = 1\} - \mu_{i+1}(a) \right) \\
  = &
  .5 \left( 0 - .5 \right)
  + .5 \left( 0 \right) \\
\end{align}
}

\end{proof}

\pagebreak
\section{Benchmark models}
\label{app:models}

\subsection{Turing Machine}
\label{app:turingModel}
\begin{python}
const_nStateMem  = #__HYPERPARAM_const_nStateMem__
const_nStateHead = #__HYPERPARAM_const_nStateHead__
const_nTimesteps = #__HYPERPARAM_const_nTimesteps__
const_tapeLength = #__HYPERPARAM_const_tapeLength__
const_nDir = 3

const_nInstances = #__HYPERPARAM_const_nInstances__

@CompileMe([const_tapeLength,const_nDir],const_tapeLength)
def move(pos, direction):
    if   direction == 0: return pos
    elif direction == 1: return (pos + 1) % const_tapeLength
    elif direction == 2: return (pos - 1) % const_tapeLength
@CompileMe([const_tapeLength, const_tapeLength], 2)
def equalityTestPos(a,b): return 1 if a == b else 0
@CompileMe([const_nStateHead, const_nStateHead], 2)
def equalityTestState(a,b): return 1 if a == b else 0

#######################################################
#  Source code parametrisation                        #
#######################################################
newValue  = Param(const_nStateMem)[const_nStateHead, const_nStateMem]
direction = Param(const_nDir)[const_nStateHead, const_nStateMem]
newState  = Param(const_nStateHead)[const_nStateHead, const_nStateMem]

#######################################################
#  Interpreter model                                  #
#######################################################
# Memory tape
tape = Var(const_nStateMem)[const_nInstances, const_nTimesteps, const_tapeLength]
# Machine head
curPos   = Var(const_tapeLength)[const_nInstances, const_nTimesteps]
curState = Var(const_nStateHead)[const_nInstances, const_nTimesteps]
isHalted = Var(2)[const_nInstances, const_nTimesteps]
# Temporary values
tmpActiveCell = Var(2)[const_nInstances, const_nTimesteps - 1, const_tapeLength]
tmpMemState   = Var(const_nStateMem)[const_nInstances, const_nTimesteps - 1]

#__IMPORT_OBSERVED_INPUTS__

# Initialize machine head
for n in range(const_nInstances):
    curPos[n,0].set_to_constant(0)
    curState[n,0].set_to_constant(1)
    isHalted[n,0].set_to_constant(0)

# Run the Turing machine
for n in range(const_nInstances):          # loop over I/O examples
    for t in range(const_nTimesteps - 1):  # loop over program timesteps

        # Carry forward unmodified tape and head if halted
        if isHalted[n,t] == 1:
            for m in range(const_tapeLength):
                tape[n,t+1,m].set_to(tape[n,t,m])
            curState[n,t+1].set_to(curState[n,t])
            curPos[n,t+1].set_to(curPos[n,t])
            isHalted[n,t+1].set_to(isHalted[n,t])
        
        # Perform Turing update if not halted  
        elif isHalted[n,t] == 0:
            with curState[n,t] as s:
                with curPos[n,t] as x:
                    with tape[n,t,x] as Tx:
                        tmpMemState[n,t].set_to(newValue[s,Tx])
                        curPos[n,t+1].set_to(move(x, direction[s,Tx]))
                        curState[n,t+1].set_to(newState[s,Tx])
            
            # Machine halts if head enters state 0
            isHalted[n,t+1].set_to(equalityTestState(0,curState[n,t+1]))
                 
            # Write temporary value to tape                   
            for m in range(const_tapeLength):
                tmpActiveCell[n,t,m].set_to(equalityTestPos(m, curPos[n,t]))
                if tmpActiveCell[n,t,m] == 1:
                    tape[n,t+1,m].set_to(tmpMemState[n,t])
                elif tmpActiveCell[n,t,m] == 0:
                    tape[n,t+1,m].set_to(tape[n,t,m])

# Machine must be halted at end of execution
for n in range(const_nInstances):
    isHalted[n,const_nTimesteps - 1].observe_value(1)

#__IMPORT_OBSERVED_OUTPUTS__
\end{python}

\subsection{Boolean circuits}
\label{app:booleanModel}
\begin{python}
const_nGates = #__HYPERPARAM_const_nGates__
const_nWires = #__HYPERPARAM_const_nWires__
const_nGateTypes = 5

const_nInstances = #__HYPERPARAM_const_nInstances__

@CompileMe([const_two,const_two],const_two)
def AND(a,b): return int(a and b)
@CompileMe([const_two,const_two],const_two)
def OR(a,b): return int(a or b)
@CompileMe([const_two,const_two],const_two)
def XOR(a,b): return int(a ^ b)
@CompileMe([const_two],const_two)
def NOT(a): return int(not a)
@CompileMe([const_two],const_two)
def NOOP(a): return a
@CompileMe([const_nWires,const_nWires],const_two)
def equalityTest(a,b): return 1 if a == b else 0

#######################################################
#  Source code parametrisation                        #
#######################################################
gate = Param(const_nGateTypes)[const_nGates]
in1  = Param(const_nWires)[const_nGates]
in2  = Param(const_nWires)[const_nGates]
out  = Param(const_nWires)[const_nGates]

#######################################################
#  Interpreter model                                  #
#######################################################
wires = Var(2)[const_nInstances, const_nGates + 1, const_nWires]
tmpOutput  = Var(2)[const_nInstances, const_nGates]
tmpDoWrite = Var(2)[const_nInstances, const_nGates, const_nWires]
tmpArg1 = Var(2)[const_nInstances, const_nGates]
tmpArg2 = Var(2)[const_nInstances, const_nGates]

#__IMPORT_OBSERVED_INPUTS__

# Run the circuit
for n in range(const_nInstances):  # loop over I/O examples
    for g in range(const_nGates):  # loop over sequential gates

        # Load gate inputs
        with in1[g] as i1:
            with in2[g] as i2:
                tmpArg1[n,g].set_to(wires[n,g,i1])
                tmpArg2[n,g].set_to(wires[n,g,i2])
        
        # Compute gate output
        if gate[g] == 0:
            tmpOutput[n,g].set_to(  AND(tmpArg1[n,g], tmpArg2[n,g]) )
        elif gate[g] == 1:
            tmpOutput[n,g].set_to(   OR(tmpArg1[n,g], tmpArg2[n,g]) )
        elif gate[g] == 2:
            tmpOutput[n,g].set_to(  XOR(tmpArg1[n,g], tmpArg2[n,g]) )
        elif gate[g] == 3:
            tmpOutput[n,g].set_to(  NOT(tmpArg1[n,g]) )
        elif gate[g] == 4:
            tmpOutput[n,g].set_to( NOOP(tmpArg1[n,g]) )
            
        # Write gate output
        for w in range(const_nWires):
            tmpDoWrite[n,g,w].set_to(equalityTest(out[g], w))
            if tmpDoWrite[n,g,w] == 1:
                wires[n,g + 1,w].set_to(tmpOutput[n,g])
            elif tmpDoWrite[n,g,w] == 0:
                wires[n,g + 1,w].set_to(wires[n,g,w])

#__IMPORT_OBSERVED_OUTPUTS__
\end{python}

\subsection{Basic-block model}
\label{app:bbModel}
\begin{python}
const_nBlocks    = #__HYPERPARAM_const_nBlocks__
const_nRegisters = #__HYPERPARAM_const_nRegisters__
const_nTimesteps = #__HYPERPARAM_const_nTimesteps__
const_maxInt     = #__HYPERPARAM_const_maxInt__
const_nInstructions = 7
const_nActions = 2
const_nInstrPlusAct = const_nInstructions + const_nActions
const_noopIndex = const_nInstructions

const_nInstances = #__HYPERPARAM_const_nInstances__

@CompileMe([const_nInstrPlusAct], 2)
def isInstruction(a): return 1 if a < const_nInstructions else 0
@CompileMe([const_nInstrPlusAct], const_nInstructions)
def extractInstruction(a): return a
@CompileMe([const_nInstrPlusAct], const_nActions)
def extractAction(a): return a - const_nInstructions
@CompileMe([const_maxInt, const_maxInt], 2)
def equalityTestValue(a,b): return 1 if a == b else 0
@CompileMe([const_nRegisters, const_nRegisters], 2)
def equalityTestReg(a, b): return 1 if a == b else 0
@CompileMe([const_maxInt], 2)
def greaterThanZero(a): return 1 if a > 0 else 0

@CompileMe([], const_maxInt)
def ZERO(): return 0
@CompileMe([const_maxInt], const_maxInt)
def INC(a): return (a + 1) % const_maxInt
@CompileMe([const_maxInt], const_maxInt)
def DEC(a): return (a - 1) % const_maxInt
@CompileMe([const_maxInt, const_maxInt], const_maxInt)
def ADD(a, b): return (a + b) % const_maxInt
@CompileMe([const_maxInt, const_maxInt], const_maxInt)
def SUB(a, b): return (a - b) % const_maxInt
@CompileMe([const_maxInt, const_maxInt], const_maxInt)
def LESSTHAN(a, b): return 1 if a < b else 0

#######################################################
#  Source code parametrisation                        #
#######################################################
instructions = Param(const_nInstrPlusAct)[const_nBlocks]
arg1s = Param(const_nRegisters)[const_nBlocks]
arg2s = Param(const_nRegisters)[const_nBlocks]
rOuts = Param(const_nRegisters)[const_nBlocks]
thenBlocks = Param(const_nBlocks)[const_nBlocks]
elseBlocks = Param(const_nBlocks)[const_nBlocks]
rConds = Param(const_nRegisters)[const_nBlocks]

#######################################################
#  Interpreter model                                  #
#######################################################
# Program pointer
curBlocks = Var(const_nBlocks)[const_nInstances,const_nTimesteps]
# Memory
registers = Var(const_maxInt)[const_nInstances,const_nTimesteps,const_nRegisters]
heap = Var(const_maxInt)[const_nInstances,const_nTimesteps,const_maxInt]
# Temporary values
tmpIsInstr = Var(2)[const_nInstances,const_nTimesteps-1]
tmpInstr   = Var(const_nInstructions)[const_nInstances,const_nTimesteps-1]
tmpAction  = Var(const_nActions)[const_nInstances,const_nTimesteps-1]
tmpArg1Val = Var(const_maxInt)[const_nInstances,const_nTimesteps-1]
tmpArg2Val = Var(const_maxInt)[const_nInstances,const_nTimesteps-1]
tmpOutput  = Var(const_maxInt)[const_nInstances,const_nTimesteps-1]
tmpDoWrite = Var(2)[const_nInstances,const_nTimesteps-1, const_nRegisters]
tmpCondVal = Var(const_maxInt)[const_nInstances,const_nTimesteps-1]
tmpGotoThen  = Var(2)[const_nInstances,const_nTimesteps-1]
tmpWriteHeap = Var(2)[const_nInstances,const_nTimesteps-1,const_maxInt]

# Initialize block 0 to a spining STOP block
instructions[0].set_to_constant(const_noopIndex)
thenBlocks[0].set_to_constant(0)
elseBlocks[0].set_to_constant(0)

# Initialize the program pointer to block 1 and the registers to 0
for n in range(const_nInstances):
    curBlocks[n,0].set_to_constant(1)
    for r in range(const_nRegisters):
        registers[n,0,r].set_to_constant(0)

#__IMPORT_OBSERVED_INPUTS__

# Run the program
for n in range(const_nInstances):        # loop over I/O examples
    for t in range(const_nTimesteps-1):  # loop over program timesteps
        with curBlocks[n,t] as pc:
            
            # Load block inputs
            with arg1s[pc] as a1:
                tmpArg1Val[n,t].set_to(registers[n,t,a1])
            with arg2s[pc] as a2:
                tmpArg2Val[n,t].set_to(registers[n,t,a2])

            # Determine whether block performs a heap ACTION or register INSTRUCTION
            tmpIsInstr[n,t].set_to(isInstruction(instructions[pc]))
            
            # Handle heap ACTIONS
            if tmpIsInstr[n,t] == 0:
                tmpAction[n,t].set_to(extractAction(instructions[pc]))
                
                # Actions affect the heap ...
                if tmpAction[n,t] == 0:   # NOOP
                    for m in range(const_maxInt):
                        heap[n,t+1,m].set_to(heap[n,t,m])
                elif tmpAction[n,t] == 1: # WRITE
                    for m in range(const_maxInt):
                        tmpWriteHeap[n,t,m].set_to(equalityTestValue(tmpArg1Val[n,t], m))
                        if tmpWriteHeap[n,t,m] == 1:
                            heap[n,t+1,m].set_to(tmpArg2Val[n,t])
                        elif tmpWriteHeap[n,t,m] == 0:
                            heap[n,t+1,m].set_to(heap[n,t,m])
                
                # ... and do not affect registers
                for r in range(const_nRegisters):
                    registers[n,t+1,r].set_to(registers[n,t,r])
            
            # Handle register INSTRUCTIONS                                  
            elif tmpIsInstr[n,t] == 1:
                tmpInstr[n,t].set_to(extractInstruction(instructions[curBlocks[n,t]]))
             
                # Instructions affect registers ...
                if tmpInstr[n,t] == 0:
                    tmpOutput[n,t].set_to( ZERO() )
                elif tmpInstr[n,t] == 1:
                    tmpOutput[n,t].set_to( INC(tmpArg1Val[n,t]) )
                elif tmpInstr[n,t] == 2:
                    tmpOutput[n,t].set_to( DEC(tmpArg1Val[n,t]) )
                elif tmpInstr[n,t] == 3:
                    tmpOutput[n,t].set_to( ADD(tmpArg1Val[n,t],tmpArg2Val[n,t]) )
                elif tmpInstr[n,t] == 4:
                    tmpOutput[n,t].set_to( SUB(tmpArg1Val[n,t],tmpArg2Val[n,t]) )
                elif tmpInstr[n,t] == 5:
                    tmpOutput[n,t].set_to( LESSTHAN(tmpArg1Val[n,t],tmpArg2Val[n,t]) 
                elif tmpInstr[n,t] == 6: # READ         
                    with tmpArg1Val[n,t] as a1:
                        tmpOutput[n,t].set_to(heap[n,t,a1])     
                
                for r in range(const_nRegisters):
                    tmpDoWrite[n,t,r].set_to(equalityTestReg(rOuts[pc], r))
                    if tmpDoWrite[n,t,r] == 1:
                        registers[n,t+1,r].set_to(tmpOutput[n,t])
                    elif tmpDoWrite[n,t,r] == 0:
                        registers[n,t+1,r].set_to(registers[n,t,r])
                
                # ... and do not affect the heap
                for m in range(const_maxInt):
                    heap[n,t+1,m].set_to(heap[n,t,m])
               
            # Perform branching according to condition register
            with rConds[pc] as rc:
                tmpCondVal[n,t].set_to(registers[n,t+1,rc])
            
            tmpGotoThen[n,t].set_to(greaterThanZero(tmpCondVal[n,t]))
            if tmpGotoThen[n,t] == 1:
                curBlocks[n,t+1].set_to(thenBlocks[pc])
            elif tmpGotoThen[n,t] == 0:
                curBlocks[n,t+1].set_to(elseBlocks[pc])

# Program must terminate in the STOP block
for n in range(const_nInstances):
    curBlocks[n,const_nTimesteps - 1].observe_value(0)  
#__IMPORT_OBSERVED_OUTPUTS__
\end{python}

\subsection{Assembly Model}
\label{app:assemblyModel}
\begin{python}
const_nLines     = #__HYPERPARAM_const_nBlocks__
const_nRegisters = #__HYPERPARAM_const_nRegisters__
const_nTimesteps = #__HYPERPARAM_const_nTimesteps__
const_maxInt     = #__HYPERPARAM_const_maxInt__
const_nInstructions = 7
const_nActions = 1
const_nBranches = 2
const_nInstrActBranch = const_nInstructions + const_nActions + const_nBranches

const_nInstances = #__HYPERPARAM_const_nInstances__

@CompileMe([const_nInstrActBranch], 3)
def instructionType(a): 
    if a < const_nInstructions:
        return 0
    elif a < (const_nInstructions + const_nActions):
        return 1
    else:
        return 2
@CompileMe([const_nInstrActBranch], const_nInstructions)
def extractInstruction(a): return a
@CompileMe([const_nInstrActBranch], const_nBranches)
def extractBranch(a): return a - const_nInstructions - const_nActions
@CompileMe([const_nRegisters, const_nRegisters], 2)
def equalityTestReg(a,b): return 1 if a == b else 0        
@CompileMe([const_maxInt, const_maxInt], 2)
def equalityTestValue(a,b): return 1 if a == b else 0
@CompileMe([const_nLines, const_nLines], 2)
def equalityTestLine(a,b): return 1 if a == b else 0      
@CompileMe([const_maxInt], 2)
def valueEqualsZero(a): return 1 if a == 0 else 0 
@CompileMe([const_nLines], const_nLines)
def incLine(a): return (a + 1) % const_nLines

@CompileMe([], const_maxInt)
def ZERO(): return 0
@CompileMe([const_maxInt], const_maxInt)
def INC(a): return (a + 1) % const_maxInt
@CompileMe([const_maxInt, const_maxInt], const_maxInt)
def ADD(a, b): return (a + b) % const_maxInt
@CompileMe([const_maxInt, const_maxInt], const_maxInt)
def SUB(a, b): return (a - b) % const_maxInt
@CompileMe([const_maxInt], const_maxInt)
def DEC(a): return (a - 1) % const_maxInt
@CompileMe([const_maxInt, const_maxInt], const_maxInt)
def LESSTHAN(a,b): return 1 if a < b else 0

#######################################################
#  Source code parametrisation                        #
#######################################################
instructions = Param(const_nInstrActBranch)[const_nLines]
branchAddr   = Param(const_nLines)[const_nLines]
arg1s = Param(const_nRegisters)[const_nLines]
arg2s = Param(const_nRegisters)[const_nLines]
rOuts = Param(const_nRegisters)[const_nLines]

#######################################################
#  Interpreter model                                  #
#######################################################
# Program pointer
curLine = Var(const_nLines)[const_nInstances,const_nTimesteps]
# Memory
registers = Var(const_maxInt)[const_nInstances,const_nTimesteps,const_nRegisters]
heap = Var(const_maxInt)[const_nInstances,const_nTimesteps,const_maxInt]
# Temporary values
tmpInstrActBranch = Var(3)[const_nInstances, const_nTimesteps-1]
tmpInstr   = Var(const_nInstructions)[const_nInstances,const_nTimesteps-1]
tmpBranch  = Var(const_nBranches)[const_nInstances,const_nTimesteps-1]
tmpArg1Val = Var(const_maxInt)[const_nInstances,const_nTimesteps-1]
tmpArg2Val = Var(const_maxInt)[const_nInstances,const_nTimesteps-1]
tmpOutput  = Var(const_maxInt)[const_nInstances,const_nTimesteps-1]
tmpDoWrite = Var(2)[const_nInstances,const_nTimesteps-1, const_nRegisters]
tmpBranchIsZero = Var(2)[const_nInstances,const_nTimesteps-1]
tmpWriteHeap = Var(2)[const_nInstances,const_nTimesteps-1,const_maxInt]
isHalted = Var(2)[const_nInstances, const_nTimesteps - 1]

# Initialize the program pointer to block 1 and the registers to 0
for n in range(const_nInstances):
    curLine[n,0].set_to_constant(1)
    for r in range(const_nRegisters):
        registers[n,0,r].set_to_constant(0)

#__IMPORT_OBSERVED_INPUTS__

# Run the program
for n in range(const_nInstances):        # loop over I/O examples
    for t in range(const_nTimesteps-1):  # loop over program timesteps
        # Halt if we jump to line 0
        isHalted[n,t].set_to(equalityTestLine(curLine[n,t],0))
        
        # If not halted, execute current line
        if isHalted[n,t] == 0: 
            with curLine[n,t] as pc:
                # Load line inputs
                with arg1s[pc] as a1:
                    tmpArg1Val[n,t].set_to(registers[n,t,a1])
                with arg2s[pc] as a2:
                    tmpArg2Val[n,t].set_to(registers[n,t,a2])
                
                # Determine whether line performs a register INSTRUCTION, a heap ACTION or 
                # a control flow BRANCH
                tmpInstrActBranch[n,t].set_to(instructionType(instructions[pc]))

                # Handle register INSTRUCTIONS                    
                if tmpInstrActBranch[n,t] == 0:
                    tmpInstr[n,t].set_to(extractInstruction(instructions[pc]))

                    # Instructions affect registers ...            
                    if tmpInstr[n,t] == 0:
                        tmpOutput[n,t].set_to( ZERO() )
                    elif tmpInstr[n,t] == 1:
                        tmpOutput[n,t].set_to( INC(tmpArg1Val[n,t]) )
                    elif tmpInstr[n,t] == 2:
                        tmpOutput[n,t].set_to( ADD(tmpArg1Val[n,t],tmpArg2Val[n,t]) )
                    elif tmpInstr[n,t] == 3:
                        tmpOutput[n,t].set_to( SUB(tmpArg1Val[n,t],tmpArg2Val[n,t]) )
                    elif tmpInstr[n,t] == 4:
                        tmpOutput[n,t].set_to( DEC(tmpArg1Val[n,t]) )
                    elif tmpInstr[n,t] == 5:
                        tmpOutput[n,t].set_to( LESSTHAN(tmpArg1Val[n,t],tmpArg2Val[n,t]) )
                    elif tmpInstr[n,t] == 6:               
                        with tmpArg1Val[n,t]:
                            tmpOutput[n,t].set_to(heap[n,t,tmpArg1Val[n,t]])     
                    
                    for r in range(const_nRegisters):
                        tmpDoWrite[n,t,r].set_to(equalityTestReg(rOuts[pc], r))
                        if tmpDoWrite[n,t,r] == 1:
                            registers[n,t+1,r].set_to(tmpOutput[n,t])
                        elif tmpDoWrite[n,t,r] == 0:
                            registers[n,t+1,r].set_to(registers[n,t,r])
                    
                    # ... and do not affect the heap
                    for m in range(const_maxInt):
                        heap[n,t+1,m].set_to(heap[n,t,m])

                    # Progress to the next line    
                    curLine[n,t+1].set_to(incLine(pc))

                # Handle heap ACTIONS  
                elif tmpInstrActBranch[n,t] == 1:
                    # The only action is to write to the heap                
                    for m in range(const_maxInt):
                        tmpWriteHeap[n,t,m].set_to(equalityTestValue(tmpArg1Val[n,t],m))
                        if tmpWriteHeap[n,t,m] == 1:
                            heap[n,t+1,m].set_to(tmpArg2Val[n,t])
                        elif tmpWriteHeap[n,t,m] == 0:
                            heap[n,t+1,m].set_to(heap[n,t,m])
                    
                    # Actions do not affect the registers
                    for r in range(const_nRegisters):
                        registers[n,t+1,r].set_to(registers[n,t,r])

                    # Progress to the next line    
                    curLine[n,t+1].set_to(incLine(pc))
                
                # Handle control flow BRANCHES
                elif tmpInstrActBranch[n,t] == 2: # Branch
                    tmpBranch[n,t].set_to(extractBranch(instructions[pc]))
                    tmpBranchIsZero[n,t].set_to(valueEqualsZero(tmpArg1Val[n,t]))
                    
                    # BRANCHES affect the program counter ...
                    if tmpBranch[n,t] == 0:   # JZ
                        if tmpBranchIsZero[n,t] == 1:
                            curLine[n,t+1].set_to(branchAddr[pc])
                        elif tmpBranchIsZero[n,t] == 0:
                            curLine[n,t+1].set_to(incLine(pc))
                    elif tmpBranch[n,t] == 1: # JNZ
                        if tmpBranchIsZero[n,t] == 0:
                            curLine[n,t+1].set_to(branchAddr[pc])
                        elif tmpBranchIsZero[n,t] == 1:
                            curLine[n,t+1].set_to(incLine(pc))
                    
                    # ... and do not affect the registers and heap.
                    for r in range(const_nRegisters):
                        registers[n,t+1,r].set_to(registers[n,t,r])
                    for m in range(const_maxInt):
                        heap[n,t+1,m].set_to(heap[n,t,m])

        # Carry forward unmodified registers and heap if halted
        elif isHalted[n,t] == 1:
            for r in range(const_nRegisters):
                registers[n,t+1,r].set_to(registers[n,t,r])
            for m in range(const_maxInt):
                heap[n,t+1,m].set_to(heap[n,t,m])
            curLine[n,t+1].set_to(curLine[n,t])
            
# Program must terminate in the STOP line
for n in range(const_nInstances):
    curLine[n,const_nTimesteps-1].observe_value(0)
#__IMPORT_OBSERVED_OUTPUTS__
\end{python}

\bibliography{references}
\bibliographystyle{plainnat}

\end{document}